%% file: main.tex
\documentclass{article}

\PassOptionsToPackage{compress, round,authoryear}{natbib}
\usepackage[preprint]{neurips_2026}

\input{notations}

\usepackage[utf8]{inputenc} 
\usepackage[T1]{fontenc}    
\usepackage{hyperref}       
\usepackage{url}            
\usepackage{booktabs}       
\usepackage{amsfonts}       
\usepackage{nicefrac}       
\usepackage{microtype}      
\usepackage{xcolor}         

\title{Evolutionary Feature Engineering for Structured Data}

\author{%
\begingroup
\normalfont
\begin{tabular}{c}
\textbf{Ege Onur Taga}\textsuperscript{1} \quad
\textbf{Yilin Zhuang}\textsuperscript{1} \quad
\textbf{M.\ Emrullah Ildiz}\textsuperscript{1} \quad
\textbf{Petros Mol}\textsuperscript{2} \vspace{3pt}\\ 
\textbf{Abhimanyu Das}\textsuperscript{2} \quad
\textbf{Karthik Duraisamy}\textsuperscript{1} \quad
\textbf{Samet Oymak}\textsuperscript{1,2} \\[2mm]
\textsuperscript{1}\,University of Michigan \qquad
\textsuperscript{2}\,Google Research 
\end{tabular}
\endgroup
}

\begin{document}

\maketitle

\begin{abstract}
Large language models are increasingly used as open-ended search operators in evolutionary optimization. We introduce Evolutionary Feature Engineering (EFE), a framework for using LLM-based evolution to discover preprocessing transformations for structured data. EFE represents transformations as Python programs with a standardized \texttt{fit}/\texttt{transform} interface, allowing them to be inserted directly into existing machine learning pipelines. During evolution, candidate programs are refined using dataset context, summary statistics, and downstream performance feedback on validation set. We instantiate EFE in two settings. For time-series forecasting, EFE-Time learns invertible, dataset-specific normalizations that improve off-the-shelf time-series foundation models. It reduces forecasting errors (MASE, WQL, MAE) 3\% or more when averaged across datasets and improvements are as much as 19\% on the COVID-Deaths dataset. Notably, these improvements occur with recent TSFMs such as Chronos-2. For tabular prediction, EFE-Tab evolves compact feature programs that add useful interpretable features and remove redundant ones, improving or matching existing LLM-based feature-engineering methods. We found EFE-Tab to be particularly effective on classical decision trees, where small sets of evolved features yield competitive accuracy while preserving interpretability. Overall, EFE demonstrates that LLM-based evolution can improve both accuracy and interpretability when automatically tackling structured data.
\end{abstract}

\input{sections/intro}
\input{sections/method}

\input{sections/experiments_edited}
\input{sections/conclusion}

\input{sections/acknowledgements}

\bibliographystyle{plainnat}
\bibliography{references}
\clearpage

\appendix
\input{sections/related_work_edit}
\input{sections/appendix}


\end{document}

%% file: notations.tex
\usepackage{times}
\usepackage{subcaption}
\usepackage{caption}

\usepackage{booktabs}
\usepackage{multirow}

\usepackage[utf8]{inputenc} 
\usepackage[T1]{fontenc}    
\usepackage{hyperref}       
\usepackage{url}            
\usepackage{booktabs}       
\usepackage{amsfonts,amsmath,amssymb}       
\usepackage{nicefrac}       
\usepackage{microtype}      
\usepackage{xcolor}         
\usepackage{txfonts}
\usepackage{graphicx}
\usepackage{wrapfig,bm,comment, color}
\usepackage{breakurl,epsfig,epsf,fmtcount,semtrans,multirow,boldline}
\usepackage{array}
\usepackage{tcolorbox}
\tcbuselibrary{skins,breakable}
\usepackage{tikz}
\usepackage{fancyvrb}
\usepackage[frozencache,cachedir=minted-cache]{minted}

\definecolor{boxblue}{RGB}{70,130,180}     
\definecolor{boxgreen}{RGB}{110,165,120}   
\definecolor{boxorange}{RGB}{230,150,80}   

\newtcolorbox{promptbox}[1][]{%
  enhanced, colback=boxblue!4, colframe=boxblue,
  boxrule=0.5pt, arc=1.5pt, breakable,
  coltitle=white, colbacktitle=boxblue,
  fonttitle=\bfseries\small, title=#1,
  left=4pt, right=4pt, top=4pt, bottom=4pt}

\newtcolorbox{responsebox}[1][]{%
  enhanced, colback=boxgreen!4, colframe=boxgreen,
  boxrule=0.5pt, arc=1.5pt, breakable,
  coltitle=white, colbacktitle=boxgreen,
  fonttitle=\bfseries\small, title=#1,
  left=4pt, right=4pt, top=4pt, bottom=4pt}

\newtcolorbox{codebox}[1][]{%
  enhanced, colback=boxorange!4, colframe=boxorange,
  boxrule=0.5pt, arc=1.5pt, breakable,
  coltitle=white, colbacktitle=boxorange,
  fonttitle=\bfseries\small, title=#1,
  left=4pt, right=4pt, top=4pt, bottom=4pt}

\setminted{fontsize=\footnotesize, breaklines=true, breakanywhere=true,
           tabsize=2, frame=none, autogobble}
\newcommand{\best}[1]{\(\bm{#1}\)}
\definecolor{darkred}{RGB}{150,0,0}
\definecolor{darkredred}{RGB}{0,0,0}
\definecolor{darkgreen}{RGB}{0,150,0}
\definecolor{darkblue}{RGB}{0,0,200}
\hypersetup{colorlinks=true, linkcolor=darkred, citecolor=darkgreen, urlcolor=darkblue}

\newcommand{\beq}{\begin{equation}}
\newcommand{\ba}{\begin{align}}
\newcommand{\ea}{\end{align}}

\newcommand{\eeq}{\end{equation}}
\newcommand{\na}{--}



%% file: sections/intro.tex
\section{Introduction}
\label{sec:introduction}

Feature engineering has a simple premise: a model need not change if the data can
be presented in a better form. In forecasting, normalization or detrending can make a
sequence easier to extrapolate; in tabular prediction, a ratio, interaction, threshold,
or aggregate can make a simple classifier substantially more expressive. Such transformations are often obvious in hindsight to human experts but difficult to specify in advance, since they depend on
the dataset, the downstream model, and structure not captured by fixed preprocessing
libraries.

Large language models make it possible to search over a richer open-ended space of executable
feature-engineering programs. Recent systems such as AlphaEvolve
\citep{novikov2025alphaevolve} show that LLMs can guide evolutionary code discovery,
while LLM-based feature-engineering methods such as CAAFE and LLM-FE show that dataset
context can help construct useful tabular features \citep{caafe,llmfe}.  These results suggest a broader question: can LLM-based evolution discover useful preprocessing programs for structured data, specifically time series and tables, where temporal order and column structure constrain and guide LLM's action space?

We propose \emph{Evolutionary Feature Engineering} (EFE) as a framework for evolving
feature transformations as Python programs. Each candidate follows a standard
\texttt{fit}/\texttt{transform} interface and is evaluated by inserting it before a
fixed downstream model. An LLM proposes transformations using dataset metadata, summary
statistics, and outcomes of previous trials; validation performance is then returned
to the evolutionary loop as feedback. Thus, search is driven directly by downstream task
performance and dataset-specific context which captures domain expertise.

\begin{figure}[t]
  \centering
  \includegraphics[width=0.95\linewidth]{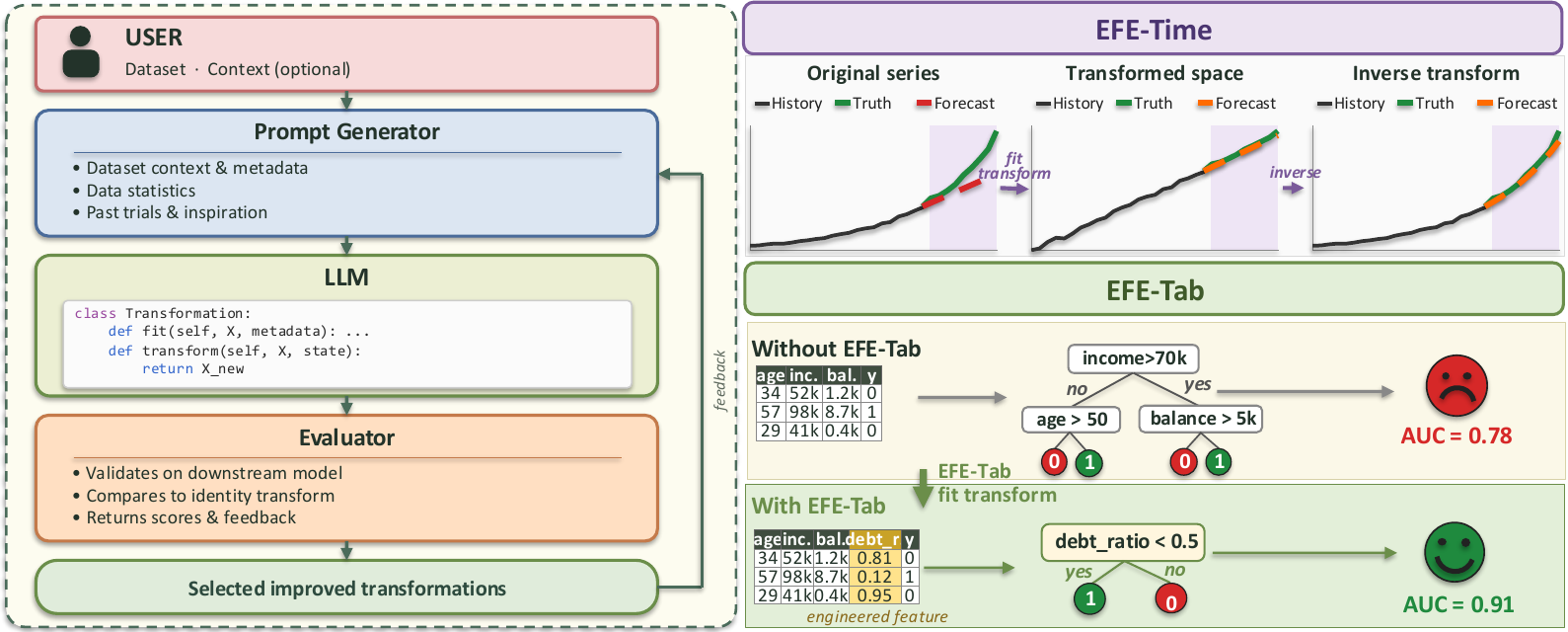}
  \caption{\small Given a dataset and optional context, EFE uses metadata, statistics,
  and past feedback to prompt an LLM to propose \texttt{fit}/\texttt{transform}
  programs. Candidates are evaluated against the identity baseline, and scores are fed
  back into an evolutionary loop.}
  \vspace{-14pt}
  \label{fig:main_figure}
\end{figure}

We instantiate EFE in two settings as depicted in Figure \ref{fig:main_figure}. For time-series forecasting, \emph{EFE-Time}
evolves dataset-specific, invertible normalization programs with \texttt{fit},
\texttt{transform}, and \texttt{inverse\_transform} methods. This is motivated by the
importance of normalization for time-series foundation models: widely used transforms
such as RevIN \citep{kim2022reversible} or arcsinh-style scaling
\citep{ansari2025chronos} are useful but necessarily limited, since no single rule can
handle all trends, scales, outliers, and seasonal patterns. EFE-Time instead searches
for transformations tailored to each dataset. The forecaster operates in the transformed
space, and predictions are mapped back before evaluation (Figure \ref{fig:main_figure}, top right).

For tabular prediction, EFE-Tab searches for small, high-value feature programs as illustrated in Figure \ref{fig:main_figure}.
Rather than generating large feature sets, it optimizes validation AUC improvement over
the raw-feature baseline while penalizing added or dropped features. This encourages
transformations that improve prediction enough to justify their complexity, making the
approach especially suitable when we intend the downstream model to remain simple and
interpretable. 

Experiments show that EFE improves performance in both domains. On synthetic
exponential-trend forecasting tasks, EFE-Time discovers transformations that make
series easier for foundation models to extrapolate. On real time-series datasets from
GIFT-Eval \citep{aksu2024giftevalbenchmarkgeneraltime}, EFE-Time improves Chronos-2
\citep{ansari2025chronos} on several datasets and transfers to other forecasters,
including TimesFM~2.5 \citep{das2024decoder}, Moirai~2.0
\citep{liu2026moirai20timeseries}, and Reverso \citep{reverso}. It is also
complementary to model adaptation: applying the evolved transform before fine-tuning yields larger average gains than either component alone (Figure \ref{fig:finetune-summary}). On tabular
datasets, EFE-Tab achieves the best mean rank among compared
feature-engineering methods across TabPFN, LightGBM, and decision trees, with
particularly strong gains for decision trees (see Table \ref{tab:efe_tab_results}, Figure \ref{fig:efetab_rank}). 

Overall, our contributions are threefold: \textbf{First}, we formulate feature engineering for structured
data as evolutionary search over stateful preprocessing programs. \textbf{Second}, we introduce
EFE-Time, which evolves invertible dataset-specific transformations for time-series
foundation models. \textbf{Finally}, we introduce EFE-Tab, a parsimonious feature-engineering
method that balances predictive improvement against feature complexity. 

\textbf{Related Work.}
Recent work has used LLMs to automate feature engineering for tabular data, including CAAFE, which proposes semantic feature interactions from dataset descriptions~\citep{caafe}, and OCTree, which combines LLM reasoning with feedback from shallow decision trees~\citep{octree}. Closest to our work, LLM-FE applies FunSearch-style evolutionary search to optimize tabular feature-transformation programs using dataset context~\citep{llmfe,romera-paredes2024funsearch}. In contrast, we also evolve full invertible time-series normalization programs with train-fitted state and explicit inverse transformations, following the broader full-program optimization paradigm of AlphaEvolve~\citep{novikov2025alphaevolve}. While ELATE also uses evolutionary LLM search for time-series covariate generation~\citep{murray2025elateevolutionarylanguagemodel}, it does not target invertible normalizations. Since no codebase is available for ELATE, we omit comparisons against its covariate-generation method. We provide further discussion of related work in Appendix~\ref{app:relatedWork}.

  \begin{figure}[t]                                                             
      \centering
      \begin{subfigure}[t]{0.45\linewidth}              
          \centering                                                                 
          \includegraphics[width=\linewidth]{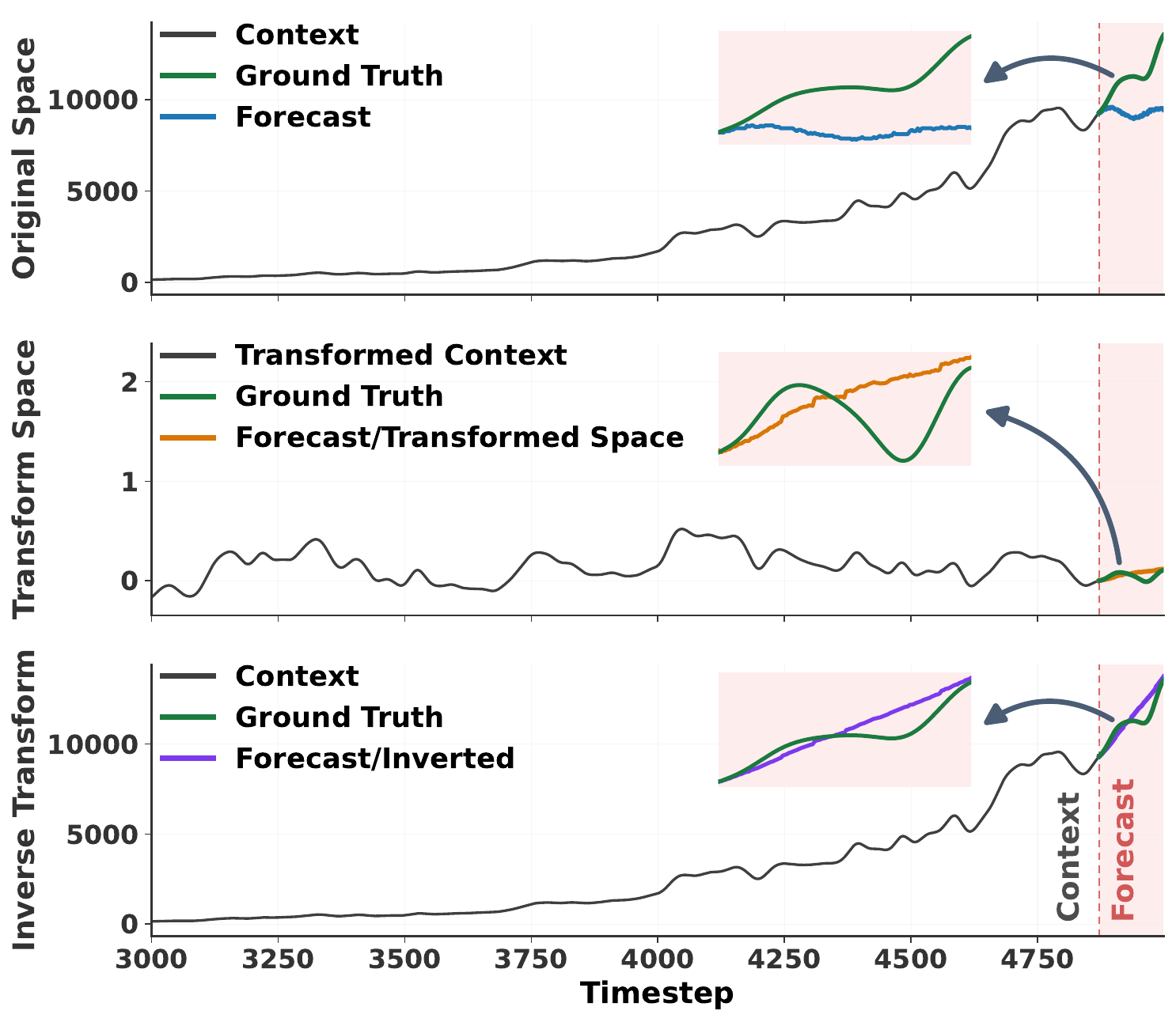}
          \caption{}
          \label{fig:synthetic-3panels}                                                                                       
      \end{subfigure}                                    
      \hfill                                                
      \begin{subfigure}[t]{0.49\linewidth}
          \centering                                                                                                          
          \includegraphics[width=\linewidth]{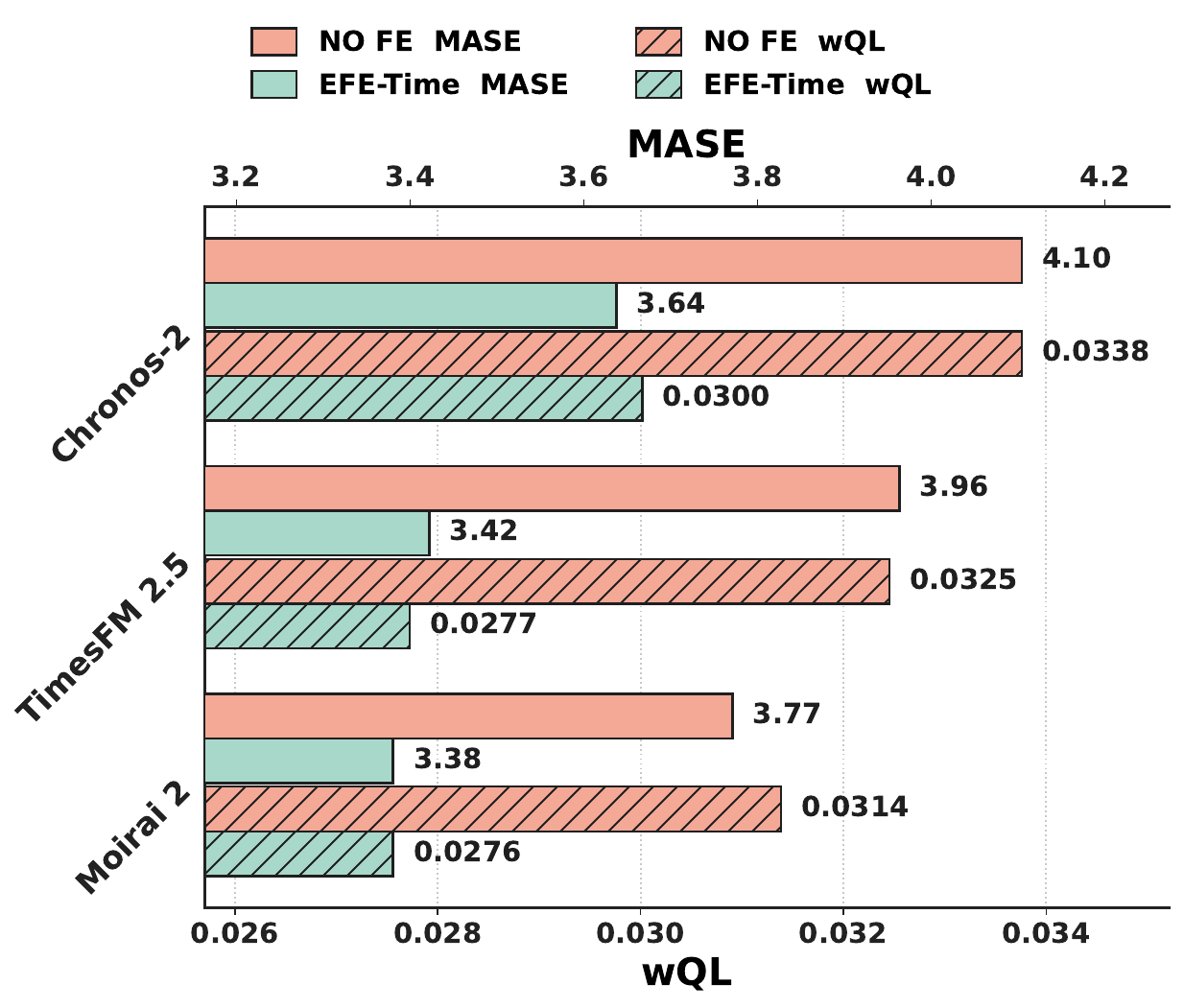}
          \caption{                 
          }                                        
          \label{fig:synthetic-benchmark-bars}                                                            
      \end{subfigure}                                
      \caption{\small (a) Effect of the evolved transform on forecasting: Top: Chronos-2 on a steep exponential series under-extrapolates vs. ground truth; Middle: after the transform the series is near-stationary and forecasts better match the transformed truth; Bottom: inverse-transforming the forecast yields a much closer fit. (b) Aggregate forecast metrics on a synthetic exponential-growth benchmark (100 series, length 5000, 128-step horizon): NO FE uses raw inputs, EFE-Time forecasts in transformed space and inverts back.
}                                                                 
      \label{fig:transform-overview}                 
  \end{figure}

%% file: sections/method.tex
\input{sections/setup}

%% file: sections/setup.tex
\section{Evolving Feature Programs}
\label{sec:method}
Evolutionary Feature Engineering (EFE) searches over executable preprocessing programs rather
than over model architectures. Each candidate program follows a standard
\texttt{fit}/\texttt{transform} interface, so the selected transformation can be inserted directly
before an existing forecasting or prediction pipeline. The downstream predictive routine is fixed: EFE does not change model architecture, hyperparameters, or training protocol, but only the representation passed into that routine.

Starting from the identity transformation, EFE runs an evolutionary loop. At each iteration, a prompt
generator provides an LLM with dataset context, metadata, summary statistics, previous evaluation
feedback, and examples of high-performing programs. The LLM proposes a modified transformation
program. The evaluator checks that the program is executable and leakage-free, inserts it before the
fixed downstream model, and scores it against the identity baseline. The resulting score and feedback
are returned to the prompt generator, allowing later candidates to build on previous successes and
failures. Thus, the LLM acts as a variation operator, while selection is driven by downstream
validation performance.

\subsection{General Setup}

Let $\mathcal{D}$ be a structured dataset with optional context $c$. A validation
protocol produces $K$ evaluation instances \(
    \Pi(\mathcal{D})
    =
    \{(D^{\mathrm{fit}}_j,D^{\mathrm{in}}_j,Y_j)\}_{j=1}^K,
\) where $D^{\mathrm{fit}}_j$ is the data available to a preprocessing program when fitting
its state, $D^{\mathrm{in}}_j$ is the input passed to the downstream routine, and $Y_j$
is the held-out target used only for evaluation. A candidate program $p$ consists of three operations:
$\mathrm{fit}_p$, $\mathrm{transform}_p$, and $\mathrm{post}_p$. On validation instance
$j$, it is applied as
\[
    \sigma_{j,p}=\mathrm{fit}_p(D^{\mathrm{fit}}_j;c),
    \quad
    \widetilde D_{j,p}=\mathrm{transform}_p(D^{\mathrm{in}}_j;\sigma_{j,p}) \quad     \widehat Y_{j,p}
    =
    \mathrm{post}_p\!\left(
        \mathcal{B}_j(\widetilde D_{j,p});
        \sigma_{j,p}
    \right),
\]

where $\mathcal{B}_j$ is a fixed downstream predictive routine. EFE does not modify
$\mathcal{B}_j$; it searches only over preprocessing programs $p$. For a lower-is-better loss $\ell$, we define
\(\widehat L(p)=
    \frac{1}{K}\sum_{j=1}^K
    \ell(Y_j,\widehat Y_{j,p}).\) Let $p_{\mathrm{id}}$ denote the identity preprocessing program. For loss-based tasks,
we measure relative improvement as
\(    \Delta(p)
    =
    1-
    \widehat L(p)/\widehat L(p_{\mathrm{id}}).\) For utility-based metrics such as AUC, we instead use additive improvement over
$p_{\mathrm{id}}$. The final score $s(p)$ combines validation improvement with
reliability, runtime, and complexity penalties. Exact final scores for EFE-Time and EFE-Tab are provided in Sections~\ref{subsec:specialization_efe_time} and \ref{subsec:specialization_efe_tab} The valid program class
$\mathcal{P}_{\mathrm{valid}}$ enforces executability, deterministic behavior, preservation
of the required input structure, and leakage prevention: $\mathrm{fit}_p$ may use only
$D^{\mathrm{fit}}_j$ and context $c$, while $Y_j$ is accessible only to the evaluator.

\subsection{LLM-Driven Evolutionary Optimization}

Following \citet{liu2026evox}, we formalize the evolutionary optimization where EFE performs sequential search over $\mathcal{P}_{\mathrm{valid}}$. After $t$ evaluations,
the optimizer maintains the history \(\mathcal{H}_t=\{(p_i,s_i,a_i)\}_{i=1}^{t},\)
where $s_i=S(p_i)$ and $a_i$ contains auxiliary feedback such as logs, diagnostics, or
evaluator feedback. At step $t$, a search strategy $S_t$ selects parent candidates, a prompt-level
variation operator, and optional inspiration examples: \(
    (pc_t,\pi_t,I_t)\sim C_{S_t}(\mathcal{H}_t).
\)
An LLM-based generator proposes a new candidate, \(p_{t+1}\sim G_{\mathrm{sol}}(\cdot\mid pc_t,\pi_t,I_t),\)
which is evaluated and appended to the history:
\[
    (s_{t+1},a_{t+1})=E(p_{t+1}),
    \qquad
    \mathcal{H}_{t+1}
    =
    \mathcal{H}_t\cup\{(p_{t+1},s_{t+1},a_{t+1})\}.
\]
Under budget $T$, EFE returns the best program in the final population:
\[
    p^\star \in \arg\max_{(p,s,a)\in\mathcal{H}_T} s .
\]
Thus, the search strategy determines which candidates are reused, how they are varied, and
which prior examples are shown to the generator.

\subsection{Specialization to EFE-Time}
\label{subsec:specialization_efe_time}

For time-series forecasting, \(\mathcal{D}=\{y^{(i)}_{1:T_i}\}_{i=1}^M\) and each validation instance is a rolling forecast window $j=(i,t)$ with history
$y^{(i)}_{1:t}$ and future target $y^{(i)}_{t+1:t+H}$:
\[
    D^{\mathrm{fit}}_j=y^{(i)}_{1:t},
    \qquad
    D^{\mathrm{in}}_j=(y^{(i)}_{1:t},H),
    \qquad
    Y_j=y^{(i)}_{t+1:t+H}.
\]
The fixed downstream routine is a forecaster $F_\theta$. A candidate program transforms
the history, the forecaster predicts in the transformed space, and the program maps the
forecast back:
\[
    z^{(i)}_{1:t}
    =
    \mathrm{transform}_p(y^{(i)}_{1:t};\sigma_{j,p}),
    \quad
    \widehat z^{(i)}_{t+1:t+H}
    =
    F_\theta(z^{(i)}_{1:t},H,c^{(i)}) \quad \widehat y^{(i)}_{t+1:t+H}
    =
    \mathrm{inverse\_transform}_p
    \left(
        \widehat z^{(i)}_{t+1:t+H};
        \sigma_{j,p}
    \right),
\]

Thus, $\mathrm{post}_p\equiv \mathrm{inverse\_transform}_p$. Because predictions are made
in the transformed space, EFE-Time restricts search to approximately invertible programs. For example, a candidate may fit a recent median $m$ and robust scale $q$, transform $y$ to $(y-m)/q$, let the forecaster predict in that normalized space, and invert predictions as $qz+m$. Using MASE as the selection loss,
\[
    s_{\mathrm{time}}(p)
    =
    \left(
        1-
        \frac{\mathrm{MASE}(p)}
             {\mathrm{MASE}(p_{\mathrm{id}})}
        -
        \lambda^{\mathrm{time}}_{\tau}\tau(p)
    \right)
    \mathbf{1}\{p\in\mathcal{P}_{\mathrm{inv}}\}.
\]
Typical EFE-Time normalizations include robust scaling, variance-stabilizations, history-only detrending, seasonal adjustment, and changepoint-aware rescaling computed from the observed history.

\subsection{Specialization to EFE-Tab}\label{subsec:specialization_efe_tab}

For tabular prediction, \(\mathcal{D}=\{(x_i,y_i)\}_{i=1}^n\), and
each validation instance is a fold
$(I^{\mathrm{tr}}_j,I^{\mathrm{val}}_j)$:
\[
    D^{\mathrm{fit}}_j
    =
    (X_{I^{\mathrm{tr}}_j},y_{I^{\mathrm{tr}}_j}),
    \qquad
    D^{\mathrm{in}}_j
    =
    (X_{I^{\mathrm{tr}}_j},y_{I^{\mathrm{tr}}_j},X_{I^{\mathrm{val}}_j}),
    \qquad
    Y_j=y_{I^{\mathrm{val}}_j}.
\]
A candidate program fits its state on the training fold and applies the same state to both
training and validation features:
\[
    \sigma_{j,p}
    =
    \mathrm{fit}_p(X_{I^{\mathrm{tr}}_j},y_{I^{\mathrm{tr}}_j};c),
    \qquad
    \widetilde X_{{I_j^{\mathrm{tr}}},p}
    =
    \mathrm{transform}_p(X_{I_j^{\mathrm{tr}}};\sigma_{j,p}), \qquad
    \widetilde X_{{I_j^{\mathrm{val}}},p}
    =\mathrm{transform}_p(X_{I_j^{\mathrm{val}}};\sigma_{j,p}).
\]
The fixed learner $\mathcal{A}$ is trained on the transformed training fold and evaluated on
the transformed validation fold:
\[
    f_{j,p}
    =
    \mathcal{A}
    \left(
        \widetilde X_{I^{\mathrm{tr}}_j,p},
        y_{I^{\mathrm{tr}}_j}
    \right),
    \qquad
    \widehat y_{I^{\mathrm{val}}_j,p}
    =
    f_{j,p}(\widetilde X_{I^{\mathrm{val}}_j,p}).
\]
Here $\mathrm{post}_p\equiv \mathrm{id}$ because predictions are already in the original
target space.

For binary classification, EFE-Tab uses mean AUC improvement over the identity program:
\[
    \bar{\delta}(p)
    =
    \frac{1}{K}\sum_{j=1}^K
    \left[
        \mathrm{AUC}_j(p)-\mathrm{AUC}_j(p_{\mathrm{id}})
    \right].
\]
The final score is
\[
    s_{\mathrm{tab}}(p)
    =
    \bar{\delta}(p)
    -
    \lambda_f\sqrt{\frac{k(p)}{n}}
    -
    \lambda^{\mathrm{tab}}_{\tau}\tau(p),
\]
where $k(p)$ counts generated and dropped features, and $\tau(p)$ is runtime. This favors
feature-engineering programs whose performance gains justify their added complexity.

%% file: sections/experiments_edited.tex
\section{Experiments}
\label{sec:experiments}
We evaluate the selected programs $p^\star$ returned by EFE in two settings:
time-series forecasting and tabular prediction. EFE-Time evolves approximately
invertible programs $p$ for fixed time-series foundation models, while EFE-Tab
evolves compact feature programs for tabular classifiers. Since LLM-based
tabular feature generation has already been studied extensively
\citep{caafe,octree,llmfe}, our tabular experiments focus on whether EFE can find
small, useful programs that improve validation performance enough to justify
their complexity.\footnote{Code is available at \url{https://github.com/egetaga/EFE}.}

\subsection{Experimental Setup}
\label{sec:experimental_setup}

\textbf{Evolutionary Optimization Interface. }
Both EFE-Time and EFE-Tab are implemented on top of OpenEvolve~\citep{openevolve},
an open-source evolutionary program-search system inspired by
AlphaEvolve~\citep{novikov2025alphaevolve}. We use its code-diversity coordinate
as the MAP-Elites feature and use \texttt{Claude-Opus-4.6} as the LLM backbone in
all main experiments. All runs start from $p_{\mathrm{id}}$. At each iteration, the LLM receives
parent code, selected inspiration programs, score metadata, dataset context, and
aggregate evaluator feedback. The proposed program $p_{t+1}$ is executed, checked
for validity and leakage, inserted before a fixed downstream model, and scored
against $p_{\mathrm{id}}$. Thus, evolution changes only the preprocessing program,
not the downstream model.
\input{tables/efetime_table}

\textbf{EFE-Time implementation details. }
For EFE-Time, each dataset has a fixed dataset context computed from the
training split. It summarizes metadata such as frequency, prediction length,
seasonal period, number of evaluation series, and aggregate series statistics. At
each iteration, valid candidates $p$ receive an \texttt{evaluation\_summary} containing
aggregate validation metrics for the identity baseline and candidate, metric
ratios, error counts, timing information, and the fraction of evaluated series
helped or harmed. Failed candidates receive capped transformation or inverse
transformation errors. The prompt never includes raw time-series values, forecasts,
per-series errors, or sampled evaluation subsets.

Each candidate $p$ is a \texttt{TransformProgram} with \texttt{fit},
\texttt{transform}, and \texttt{inverse\_transform}. At execution time,
\texttt{fit} receives only one historical series $D^{\mathrm{fit}}_j$ and compact
summary metadata, returning $\sigma_{j,p}$ for the forward and inverse
transformations. The evaluator rejects candidates that change the series length,
introduce invalid values, or collapse nonconstant series. Candidates receive no
future targets $Y_j$, no data from other series, no baseline forecasts, no scoring
feedback, no evolution step, and no dataset identity at execution time.

\textbf{EFE-Tab implementation details. }
\label{subsubsec:efetab-impl-details}
For EFE-Tab, each iteration forms a validation instance $j$ by splitting the
training data into a fitting pool (Pool A) and held-out scoring pool (Pool B).
The dataset context is computed from Pool A and includes the dataset name, shape,
target column, problem type,
class balance, task description, column statistics, correlations, and a small
number of sample rows. Because Pool A rotates across iterations, these summaries
may vary slightly during evolution.

EFE-Tab evolves the \texttt{fit}/\texttt{transform} logic of a
\texttt{FeatureProgram}. Programs learn $\sigma_{j,p}$ only from Pool A and apply
the same fitted state to both Pool A and Pool B.
Candidates may create interactions, ratios, nonlinear transforms, binned features,
target encodings, group aggregates, or threshold indicators, and may also drop weak
or redundant original columns.

Candidates are first checked for required entry points, valid outputs, schema
consistency, and absence of in-place dataframe mutation. Valid programs are then
scored on Pool B with $\mathcal{A}$ set to TabPFN: the model is fit on raw Pool A
features to obtain the $p_{\mathrm{id}}$ ROC-AUC on Pool B, and then refit on the
candidate-engineered Pool A features to obtain the candidate ROC-AUC on
transformed Pool B. The evaluator reports to the LLM the baseline and candidate
AUCs, $s_{\mathrm{tab}}(p)$, generated columns, dropped columns, output-column
counts, and validation status. The evaluator also returns interpretable feedback to the LLM, computed only from Pool A:
permutation-importance labels for surviving features, correlations between features in the table, a readable decision tree
trained on the candidate representation, and an original-column importance report.

\textbf{Benchmarks and evaluation. }
We evaluate EFE-Time on ten GIFT-Eval datasets~\citep{aksu2024giftevalbenchmarkgeneraltime}
spanning healthcare, energy, finance and business, weather and nature,
transportation, and web/cloud operations. Programs
are evolved against Chronos-2~\citep{ansari2025chronos} using validation MASE
improvement over $p_{\mathrm{id}}$ as the selection signal. For each dataset and
seed, we freeze $p^\star$ and evaluate it on the held-out test split. We also
test cross-model transfer by applying the same frozen $p^\star$, without
re-evolution, to TimesFM~2.5, Moirai~2-Small, and
Reverso-Nano~\citep{das2024decoder,liu2026moirai20timeseries,reverso}. We run
three independent seeds for 100 iterations and report mean MASE, wQL, and MAE with standard
deviations. 

\begin{figure}[t]
  \centering

  \begin{subfigure}[t]{0.49\linewidth}
    \centering
    \includegraphics[width=\linewidth]{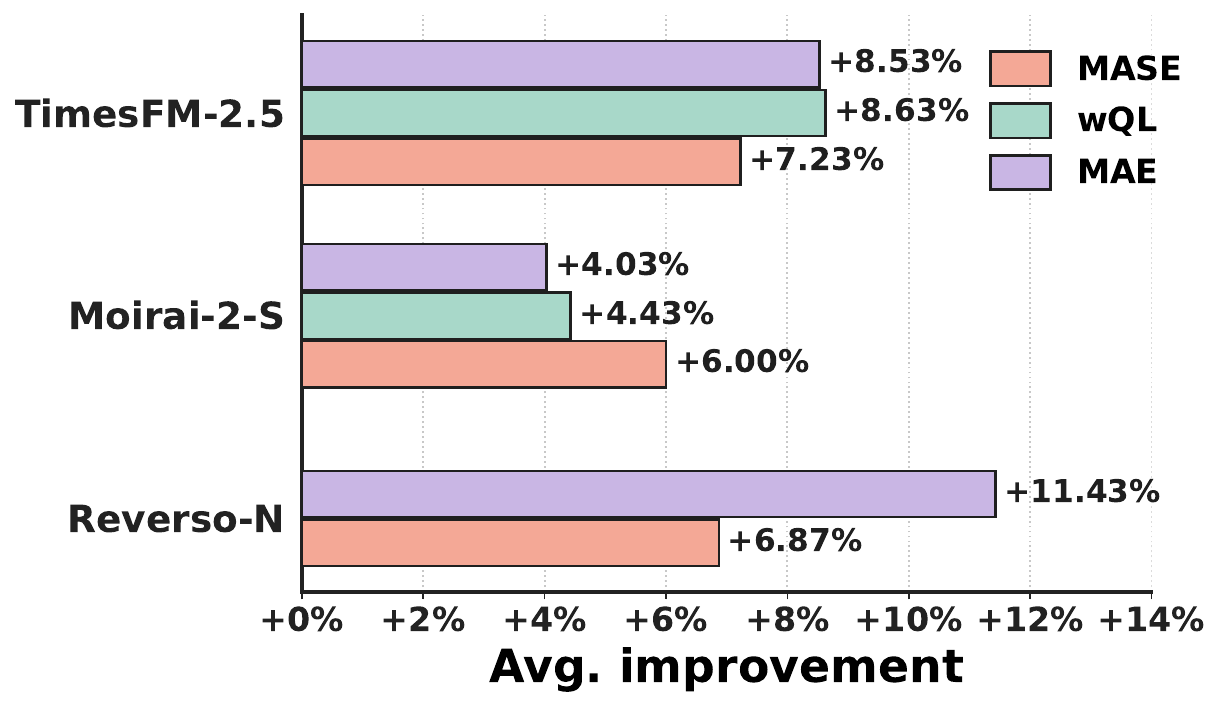}
    \caption{Over three datasets}
    \label{fig:efe-time-transfer-avg3}
  \end{subfigure}\hfill
  \begin{subfigure}[t]{0.49\linewidth}
    \centering
    \includegraphics[width=\linewidth]{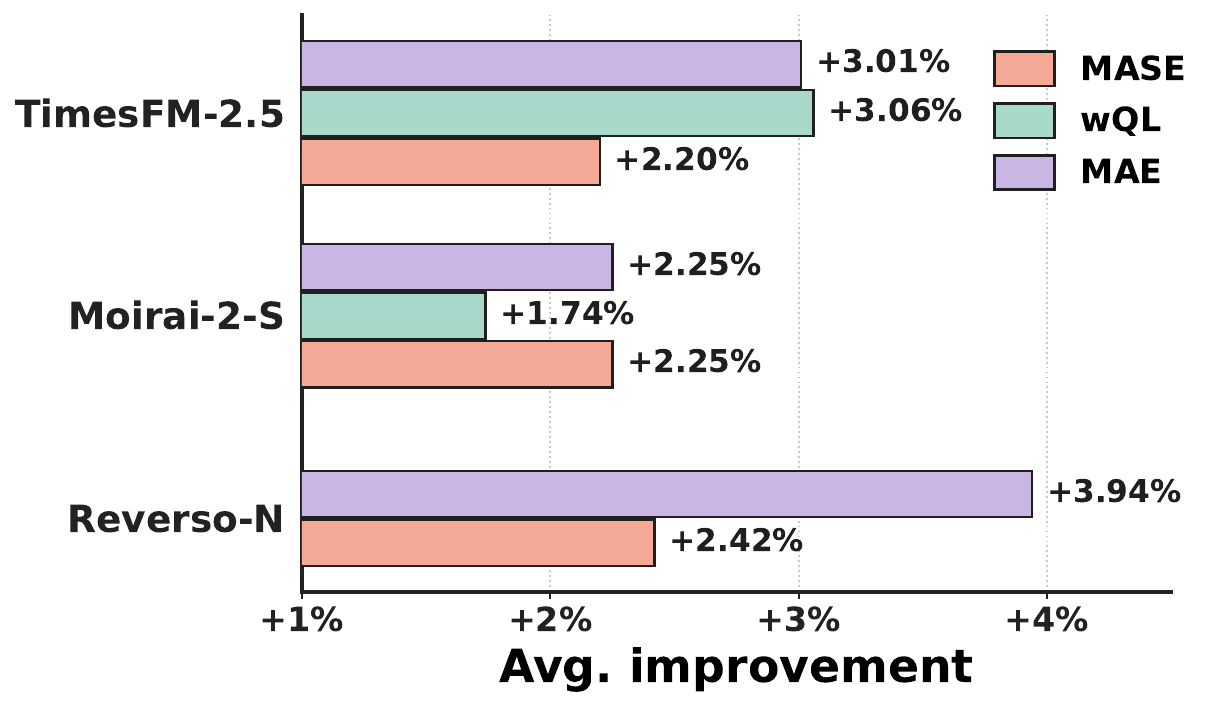}
    \caption{Over entire benchmark set}
    \label{fig:efe-time-transfer-avg10}
  \end{subfigure}

  \caption{\small Cross-model transfer of EFE-Time. The programs were evolved against Chronos-2 only and reused without modification on TimesFM-2.5, Moirai-2-Small, and Reverso-Nano. 
  (a) Average percent improvement across Covid Deaths,  M4-yearly, and solar hourly.  (b) Same setup averaged across all 10 GIFT-Eval datasets in Table~\ref{tab:efe_time_chronos_results}. We report average scores across three replicates.}
  \vspace{-7pt}
  \label{fig:efe-time-transfer}
\end{figure}

We evaluate EFE-Tab on nine binary-classification datasets from
TabArena~\citep{tabarena}, spanning telecommunication churn, employment,
e-commerce, banking, healthcare, and sports. During evolution, TabPFN-v2 \citep{grinsztajn2025tabpfn} is the
scoring model. After evolution, the selected programs are frozen and
evaluated with TabPFN-v2 \citep{grinsztajn2025tabpfn}, LightGBM, and single decision trees. To control LLM API
cost, we use the three folds from the first official repetition, run EFE-Tab for each fold for 100 iterations, and report averages and standard deviation over folds. We ran CAAFE and LLM-FE in the same way for 100 iterations for each fold. 

As API costs prevented us from conducting experiments on the entirety of Gift-Eval and TabArena, we used selected datasets from these benchmarks, chosen to ensure domain diversity.

\subsection{EFE-Time Experiments}

We begin by demonstrating that time-series foundation models struggle with synthetic data containing exponential trends, and that the learned EFE-Time programs help mitigate this issue. We then show that EFE-Time improves Chronos-2 across a range of benchmarks. Next, we demonstrate that the same learned normalizations, optimized using Chronos-2, transfer to other time-series foundation models. Finally, we show that EFE-Time provides benefits comparable to fine-tuning Chronos-2, and in some datasets even exceeds them. More importantly, EFE-Time and fine-tuning provide additive gains: combining the two improves performance beyond either approach alone. Overall, these results show that learning data-specific normalizations can significantly improve TSFM performance.

\subsubsection{TSFMs Struggle with Multiplicative Growth Patterns}

We generate synthetic time series that exhibit smooth exponential growth without spikes or abrupt events, the generation process detailed in Appendix~\ref{app:exp-synth}. Each sequence is sampled by choosing a random initial value and total growth factor, then constructing a smoothly varying monotone growth trajectory with smooth multiplicative observation noise. This produces a clean toy exponential-growth dataset in which the latent trend is monotone, while the observed sequence contains only smooth multiplicative deviations around that trend. We sample 100 examples of length 5{,}000, reserving the final 128 time steps of each series for testing.

As shown in Figure \ref{fig:synthetic-benchmark-bars}, Chronos-2, TimesFM-2.5, and Moirai-2.0 struggle with this type of multiplicative-growth data. Despite the smoothness of the observed sequences and the monotonicity of the latent trend, all three models obtain poor MASE scores. By contrast, Figure \ref{fig:synthetic-benchmark-bars} shows that the EFE-Time program $p^\star$, optimized using Chronos-2, substantially improves evaluation performance across all TSFMs considered. Figure~\ref{fig:synthetic-3panels} illustrates why this happens: the program first transforms the original series into an approximately stationary space, where Chronos-2 can produce a more accurate forecast. Applying the inverse transformation then maps this forecast back to the original scale, closely matching the ground-truth trajectory.

\begin{figure}[t]
  \centering

  \begin{subfigure}[t]{0.49\linewidth}
    \centering
    \includegraphics[width=\linewidth]{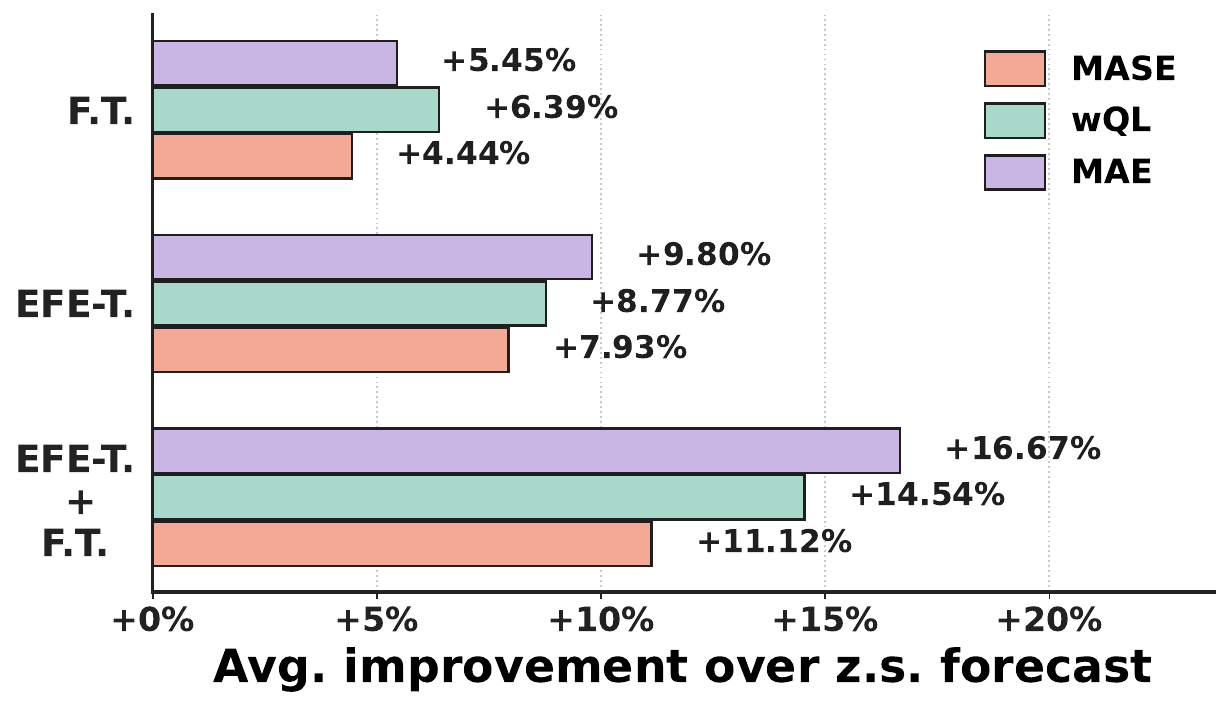}
    \caption{Over three datasets}
    \label{fig:avg3-finetune-bars}
  \end{subfigure}\hfill
  \begin{subfigure}[t]{0.49\linewidth}
    \centering
    \includegraphics[width=\linewidth]{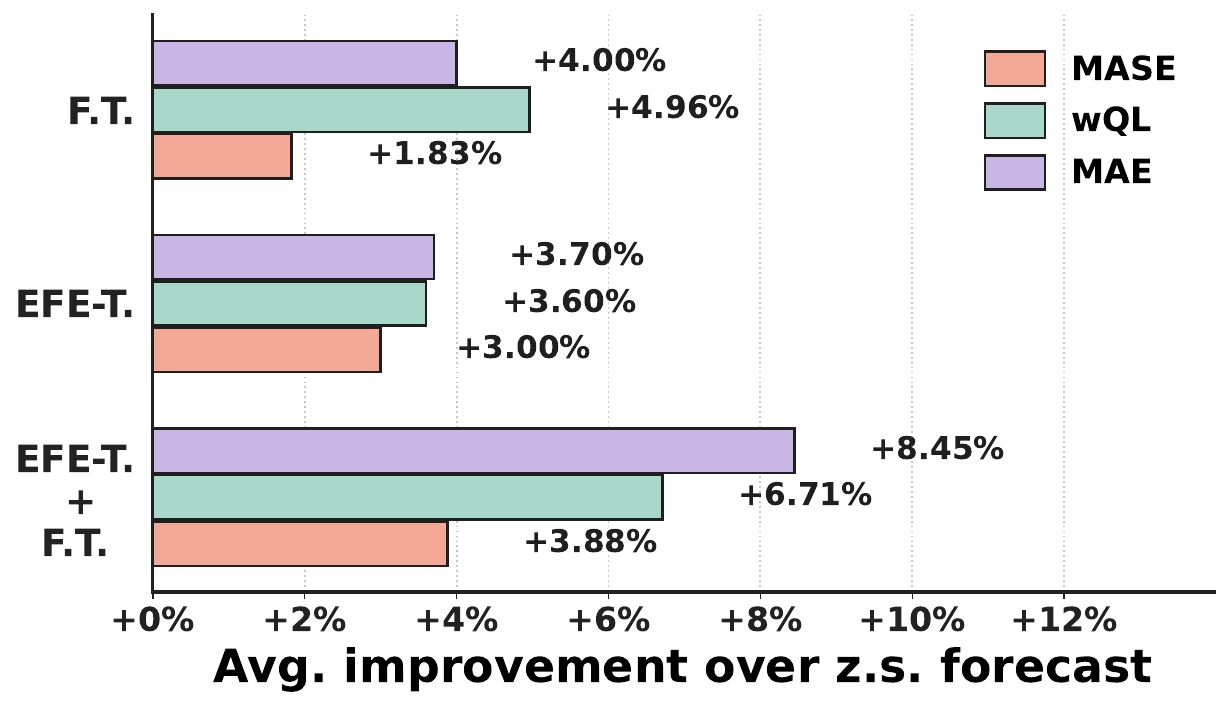}
    \caption{Over entire benchmark set}
    \label{fig:avg10-finetune-bars}
  \end{subfigure}

  \caption{\small EFE-Time, Fine-tuning, and EFE-Time+Fine-tuning gains over the Chronos-2 zero-shot forecast. (a) Average percent improvement across Covid Deaths, M4-yearly, and solar hourly. (b) Same metrics averaged across 10 GIFT-Eval datasets. In both cases, combining the evolved time-series transform with fine-tuning (EFE-T.\,+\,F.T.) yields the largest and  additive gains on every metric, outperforming either component alone.}
  \label{fig:finetune-summary}
\end{figure}
\subsubsection{Dataset Specific Normalizations Improve TSFMs}

\textbf{Learning Normalization Programs for Chronos 2.} We now evaluate EFE-Time on real datasets from the Gift-Eval benchmark \citep{aksu2024giftevalbenchmarkgeneraltime}. We begin by comparing the selected programs with Chronos-2 as the downstream forecaster. As shown in Table~\ref{tab:efe_time_chronos_results}, EFE-Time yields average improvements of $3.0\%$, $3.6\%$, and $3.7\%$ in MASE, WQL, and MAE, respectively, compared to the raw Chronos-2 forecasts, i.e., $p_{\mathrm{id}}$. Another observation is that the magnitude of improvement varies substantially across datasets: for Covid Deaths, Solar (hourly), and M4-Yearly, the improvement is as high as $19\%$, whereas for \texttt{kdd\_cup\_2018} (hourly), we observe an improvement of approximately $3\%$. For several other datasets, the gains are more modest, close to $0\%$, and statistically insignificant. This is expected, as not all datasets benefit from normalization, especially when they are already relatively regular or when normalization does not substantially reduce their forecasting difficulty. We visualize the raw Chronos-2 forecasts, the forecasts in the transformed space, and their inversion back to the original space in Figures~\ref{fig:perseries_covid_2x2}, \ref{fig:perseries_m4_2x2}, \ref{fig:perseries_solar_2x2}, \ref{fig:perseries_restaurant_2x2}, and \ref{fig:perseries_bitbrains_2x2}.

\textbf{Cross-Model Transfer of Learned Normalization Programs.} We investigate whether the learned dataset-specific normalizations transfer to other time-series foundation models. To this end, we evaluate TimesFM-2.5, Moirai-2-small (the other Moirai-2 variants are not open-sourced), and Reverso-Nano \citep{reverso}. We chose Reverso-Nano because it is extremely lightweight compared to the other TSFMs considered, allowing us to investigate whether the improvements persist for smaller TSFMs as well. We use the same programs learned using Chronos-2, without modification.

As shown in Figure~\ref{fig:efe-time-transfer}, the benefits of learned normalization programs transfer to other TSFMs. Specifically, on Covid Deaths, M4-Yearly, and Solar (hourly), shown in Figure~\ref{fig:efe-time-transfer-avg3}, we observe an average MASE improvement of more than $6\%$ across all TSFMs, reaching $7\%$ for TimesFM. Over the entire benchmark set, shown in Figure~\ref{fig:efe-time-transfer-avg10}, we again observe consistent MASE improvements of $3.06\%$, $2.25\%$, and $2.42\%$ for TimesFM, Moirai-2-small, and Reverso-Nano, respectively. Overall, the consistent improvements achieved by these TSFMs on the same datasets, using programs learned with Chronos-2, suggest that the forecasting difficulty of these datasets is not specific to a particular model architecture. Rather, it arises from the underlying time-series dynamics, which EFE-Time makes more amenable to forecasting.

\textbf{Fine-tuning and Normalizations are Additive.} Both fine-tuning and learning EFE-Time programs use the training portion of the data. In Figure~\ref{fig:finetune-summary}, we investigate how they compare and whether their benefits are additive; that is, whether first learning normalizations and then fine-tuning the model in the normalized space yields substantial gains. On datasets where normalization is highly effective, Figure~\ref{fig:avg3-finetune-bars} demonstrates that EFE-Time outperforms fine-tuning on every metric. On the entire benchmark set, shown in Figure~\ref{fig:avg10-finetune-bars}, we observe that EFE-Time provides larger gains than fine-tuning for MASE, but slightly lower gains for wQL and MAE. Importantly, as shown in Figure~\ref{fig:finetune-summary}, EFE-Time+fine-tuning yields additive benefits in both settings, outperforming either approach. 
\input{tables/efetab_table}

\textbf{Learning EFE-Time Programs Requires Strong LLMs.} Appendix~\ref{app:efetime_local_model} evaluates EFE-Time with the Qwen3.5 family~\citep{qwen3} on Covid-Deaths and Solar Hourly. Since these models are substantially weaker than Opus-4.6, we ran each experiment for 300 iterations. Qwen models perform reasonably well on Solar Hourly but are much less stable on Covid-Deaths. Given the difficulty of generating full transform and inverse-transform pipelines, this performance drop is expected. We believe that adapting EFE-Time to smaller, non-frontier LLMs remains a promising direction.

\subsection{EFE-Tab Experiments}
\textbf{Learning Parsimonious Feature Programs.} Learning feature-engineering programs for tabular data with LLMs has been widely studied \citep{caafe,llmfe,octree}. Thus, our main goal here is to show that the same EFE  paradigm can learn parsimonious programs $p^\star$, where each added or removed feature must yield a benefit greater than its associated penalty. We show that this approach improves decision trees while providing a significant interpretability advantage in real-world use.

\begin{wrapfigure}{r}{0.35\linewidth}
    \centering
    \includegraphics[width=\linewidth]{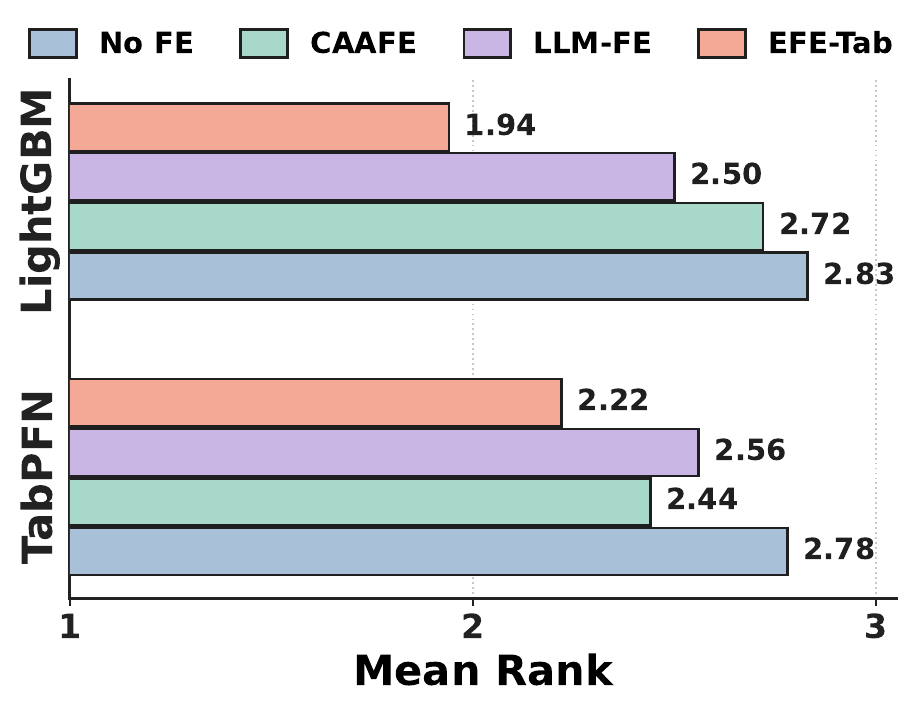}
    \caption{\small Mean rank of methods across our tabular benchmark.}
    \label{fig:efetab_rank}
    \vspace{-1.75em}
\end{wrapfigure}

We evaluate the EFE-Tab programs $p^\star$ in Table \ref{tab:efe_tab_results} and compare it with no feature engineering, CAAFE, and LLM-FE, on a single decision tree. We see that EFE-Tab yields mean rank of $1.39$ among 4 approaches when used with a single decision tree with the closest method after it achieving a mean rank of $2.44$. We see in Figure \ref{fig:efetab_rank} that EFE-Tab on our tabular benchmark also outperforms LLM-FE, and CAAFE on LightGBM and TabPFN downstream models. The separation on LightGBM is high, whereas on TabPFN, each method achieves a similar mean rank. During experiments, we have observed that due to its explicit parsimony penalty, EFE-Tab yields compact programs, whereas LLM-FE and CAAFE usually generate larger feature sets. We hypothesize that TabPFN appears more robust to redundant features, yet a single decision tree, and also LightGBM has less so. Thus, we see a lower separation between parsimonious feature engineering programs of EFE-Tab and competing approaches in TabPFN.

\begin{figure}[t]
\centering
\begin{subfigure}{0.32\linewidth}\includegraphics[width=\linewidth]{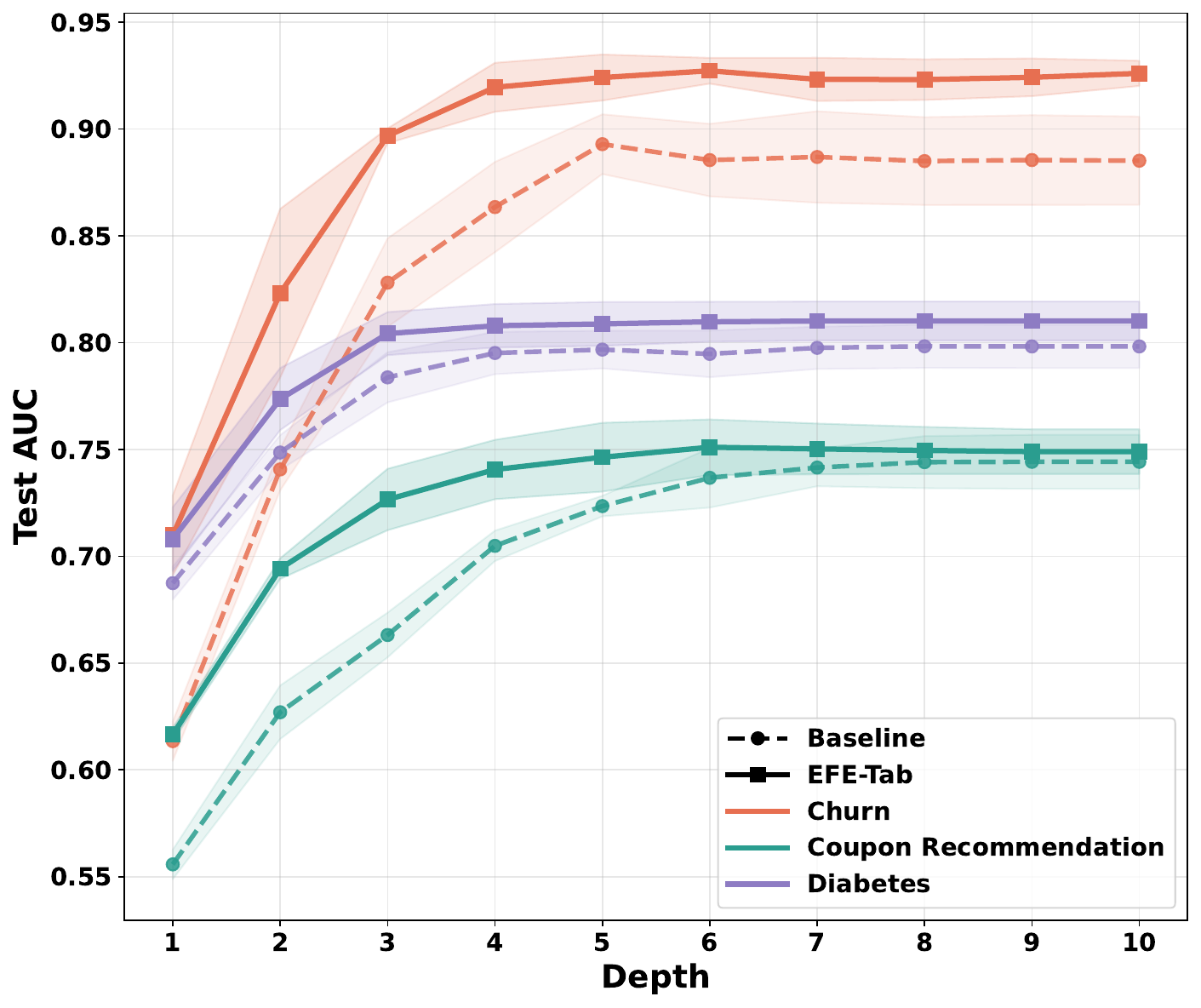}\caption{Test AUC vs.\ tree                                             
  depth.}\label{fig:efe-tab-depth-auc}
  \end{subfigure}
  \hfill
  \begin{subfigure}{0.32\linewidth}\includegraphics[width=\linewidth]{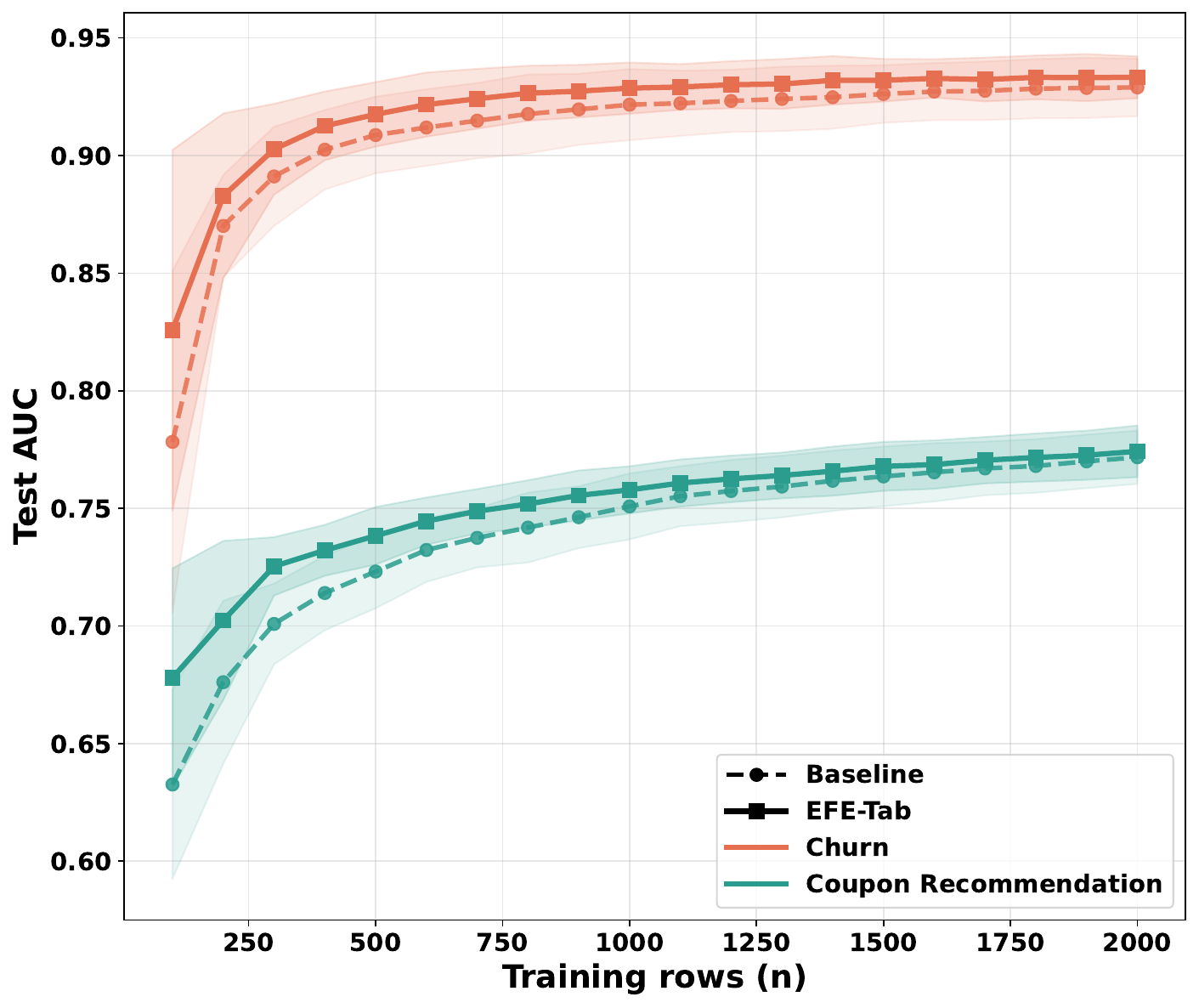}\caption{Test AUC vs.\ training
   rows.}
   \label{fig:efe-tab-data-regime-auc}
   \end{subfigure}
   \hfill
   \begin{subfigure}{0.32\linewidth}\includegraphics[width=\linewidth]{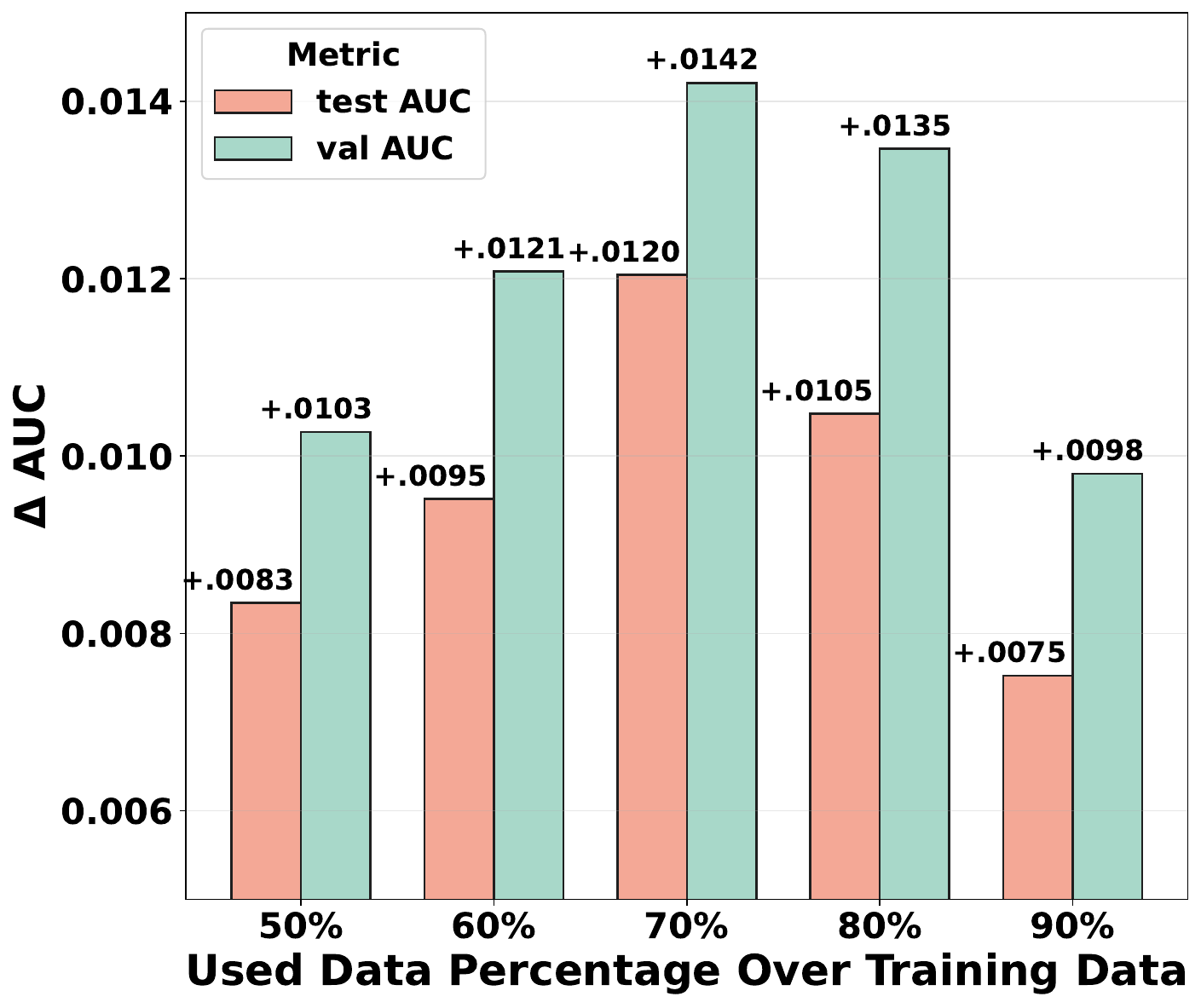}
  \caption
  {$\Delta$ AUC vs.\ used training fraction.}
  \label{fig:efe-tab-holdout-sweep}
  \end{subfigure}
  \caption{\small (a) Single decision tree over 3
  folds $\times$ 5 seeds (15 runs per cell). (b) Test AUC vs.\ training-set
  size for TabPFN. EFE-Tab features deliver a larger lift in the
  low-data regime, with the gap shrinking as more training data
  becomes available. (c) Average $\Delta$ AUC (EFE-Tab - baseline) over the held-out test and validation sets vs.\
  the fraction of training data exposed to the evolution loop,
  averaged over TabPFN/LightGBM/decision-tree and 3 seeds on \textit{churn} and
  \textit{in\_vehicle\_coupon\_recommendation}. \vspace{-7pt}}
  \label{fig:efe-tab-decision-tree}\end{figure}

\textbf{The Regimes where EFE-Tab is Useful.} In Figure \ref{fig:efe-tab-depth-auc}, we vary the decision tree depth and observe that EFE-Tab programs have particularly strong benefit on shallower decision trees. Yet, the separation exists in higher depths too, and more importantly only with depth 3, EFE-Tab achieves better performance than the no feature engineered baseline with higher depths. This signals that EFE-Tab is particularly good for learning simple but effective decision rules, which are particularly important for real-world applications requiring an explanation of the final decision. 

In Figure~\ref{fig:efe-tab-data-regime-auc}, we quantify the effect of EFE-Tab programs in low-data regimes. We use TabPFN here because it performs well in low-data regimes and does not require hyperparameter tuning, which we observed to be unstable for decision trees and LightGBM when only small amounts of training data are available. We observe that, with less training data, feature engineering significantly boosts the performance of TabPFN. As the amount of data increases, this separation diminishes, demonstrating that generating meaningful features is particularly helpful for models in low-data scenarios.

\textbf{Characterizing the Fraction of Training Data Used.} As described in Section~\ref{subsubsec:efetab-impl-details}, at each iteration we instantiate $j$ by randomly splitting the training data into Pool A and Pool B. Pool A summary is provided to the LLM and used as input to the downstream model, while Pool B is used for scoring. This split naturally introduces diversity across iterations. Figure~\ref{fig:efe-tab-holdout-sweep} shows that EFE-Tab improvements follow an inverted-U trend as the fraction assigned to pool A increases. Large pool A fractions reduce diversity across iterations, whereas very small fractions make pool A less representative of the underlying dataset.

%% file: tables/efetime_table.tex
\begin{table}[t]
\caption{\small Comparison of EFE-Time against identity transformation across time-series datasets for Chronos-2.}
\scriptsize
\label{tab:efe_time_chronos_results}
\centering
\setlength{\tabcolsep}{1.5pt}
\renewcommand{\arraystretch}{1.05}
\resizebox{\columnwidth}{!}{%
\begin{tabular}{@{}lccccccccc@{}}
\toprule
\multirow{2}{*}{\textbf{Dataset}}
& \multicolumn{3}{c}{\textbf{Baseline}}
& \multicolumn{3}{c}{\textbf{EFE-Time}}
& \multicolumn{3}{c}{\textbf{\% Improvement}} \\
\cmidrule(lr){2-4} \cmidrule(lr){5-7} \cmidrule(lr){8-10}
& \textbf{MASE} & \textbf{WQL} & \textbf{MAE}
& \textbf{MASE} & \textbf{WQL} & \textbf{MAE}
& \textbf{MASE} & \textbf{WQL} & \textbf{MAE} \\
\midrule

$\mathtt{CovidDeaths}$
& $35.444$ & $0.039$ & $124.241$
& \best{31.853 \pm 0.835} & \best{0.032 \pm 0.003} & \best{102.830 \pm 13.968}
& $10.1 \pm 2.3$ & $19.5 \pm 8.7$ & $17.2 \pm 11.2$ \\

$\mathtt{SolarHourly}$
& $0.996$ & $0.352$ & $12.934$
& \best{0.915 \pm 0.049} & \best{0.335 \pm 0.002} & \best{11.883 \pm 0.648}
& $8.1 \pm 4.9$ & $4.7 \pm 0.7$ & $8.1 \pm 5.0$ \\

$\mathtt{M4Yearly}$
& $3.370$ & $0.117$ & $905.276$
& \best{3.183 \pm 0.135} & \best{0.115 \pm 0.002} & \best{868.517 \pm 26.300}
& $5.6 \pm 3.9$ & $2.1 \pm 1.6$ & $4.1 \pm 2.9$ \\

$\mathtt{JenaWeather\ (H)}$
& $0.547$ & $0.042$ & $8.745$
& \best{0.538 \pm 0.004} & \best{0.041 \pm 0.001} & \best{8.436 \pm 0.199}
& $1.6 \pm 0.6$ & $2.1 \pm 1.3$ & $3.5 \pm 2.3$ \\

$\mathtt{SZ\_TAXI\_15T}$
& $0.543$ & $0.200$ & $2.722$
& $0.543 \pm 0.000$ & $0.200 \pm 0.000$ & $2.722 \pm 0.001$
& $0.0 \pm 0.0$ & $0.0 \pm 0.0$ & $0.0 \pm 0.0$ \\

$\mathtt{bitbrains\_storage/H}$
& $0.993$ & $0.815$ & $384.949$
& $0.993 \pm 0.000$ & $0.815 \pm 0.000$ & $384.949 \pm 0.000$
& $0.0 \pm 0.0$ & $0.0 \pm 0.0$ & $0.0 \pm 0.0$ \\

$\mathtt{bitbrains\_rnd/H}$
& $5.821$ & $1.032$ & $175.961$
& \best{5.818 \pm 0.002} & \best{1.016 \pm 0.022} & \best{168.439 \pm 10.611}
& $0.1 \pm 0.0$ & $4.3 \pm 6.0$ & $1.5 \pm 2.1$ \\

$\mathtt{us\_births\_D}$
& $0.345$ & $0.017$ & $234.437$
& \best{0.342 \pm 0.002} & \best{0.017 \pm 0.000} & \best{232.163 \pm 1.191}
& $1.0 \pm 0.5$ & $1.0 \pm 0.5$ & $0.4 \pm 0.2$ \\

$\mathtt{kdd\_cup\_2018\_H}$
& $0.999$ & $0.396$ & $24.712$
& \best{0.971 \pm 0.004} & \best{0.388 \pm 0.002} & \best{24.281 \pm 0.213}
& $2.8 \pm 0.4$ & $2.0 \pm 0.6$ & $1.7 \pm 0.9$ \\

$\mathtt{restaurant}$
& $0.684$ & $0.256$ & $7.055$
& \best{0.682 \pm 0.002} & $0.256 \pm 0.001$ & \best{7.041 \pm 0.016}
& $0.3 \pm 0.2$ & $-0.2 \pm 0.2$ & $0.2 \pm 0.2$ \\

\midrule
\textbf{Mean \% Improvement}
& \multicolumn{3}{c}{\na}
& \multicolumn{3}{c}{\na}
& $\mathbf{3.0}$ & $\mathbf{3.6}$ & $\mathbf{3.7}$ \\

\bottomrule
\end{tabular}%
}
\end{table}

%% file: tables/efetab_table.tex
\begin{table}[t]
\caption{Comparison of feature engineering methods for single decision trees across tabular datasets over ROC-AUC. The reported results are 3 outer-fold averages from the TabArena benchmark.}
\label{tab:efe_tab_results}
\small
\centering
\setlength{\tabcolsep}{4pt}
\begin{tabular}{@{}lcccc@{}}
\toprule
\multirow{2}{*}{\textbf{Dataset}}
& \multicolumn{4}{c}{\textbf{Feature Engineering Method}} \\
\cmidrule(lr){2-5}
& \textbf{No FE} & \textbf{CAAFE} & \textbf{LLM-FE} & \textbf{EFE-Tab} \\
\midrule

$\mathtt{Churn}$ 
& $.8944 \pm .0178$ & $.9239 \pm .0181$ & $.9080 \pm .0213$ & \best{.9266 \pm .0108} \\

$\mathtt{HRAnalyticsJobChange}$ 
& $.7898 \pm .0069$ & $.7864 \pm .0093$ & $.7914 \pm .0065$ & \best{.7923 \pm .0041} \\

$\mathtt{E-CommerceShippingData}$ 
& $.7417 \pm .0037$ & $.7441 \pm .0034$ & $.7433 \pm .0032$ & \best{.7448 \pm .0035} \\

$\mathtt{in\_vehicle\_coupon\_recommendation}$ 
& $.7296 \pm .0140$ & $.7285 \pm .0124$ & $.6720 \pm .1042$ & \best{.7473 \pm .0134} \\

$\mathtt{online\_shoppers\_intention}$ 
& $.9235 \pm .0068$ & $.9242 \pm .0041$ & \best{.9382 \pm .0133} & $.9235 \pm .0040$ \\

$\mathtt{Bank\_Customer\_Churn}$ 
& $.8340 \pm .0035$ & $.8409 \pm .0077$ & \best{.8502 \pm .0096} & $.8434 \pm .0040$ \\

$\mathtt{BankMarketing}$ 
& $.7329 \pm .0056$ & $.7339 \pm .0055$ & $.7345 \pm .0048$ & \best{.7348 \pm .0064} \\

$\mathtt{Diabetes}$ 
& $.7974 \pm .0185$ & $.7852 \pm .0473$ & $.7909 \pm .0112$ & \best{.8136 \pm .0082} \\

$\mathtt{Fitness\_Club}$ 
& $.8042 \pm .0118$ & $.7943 \pm .0071$ & $.7971 \pm .0105$ & \best{.8048 \pm .0128} \\

\midrule
\textbf{Mean Rank}
& $3.17$ & $3.00$ & $2.44$ & $\mathbf{1.39}$ \\

\bottomrule
\end{tabular}
\vspace{-5pt}
\end{table}

%% file: sections/conclusion.tex
\section{Conclusion and Limitations}
\label{sec:conclusion}
We introduced EFE, an evolutionary framework for feature engineering that iteratively refines feature programs. EFE-Time improves the forecasting performance of TSFMs across diverse downstream applications and achieves performance competitive with fine-tuning. More importantly, combining EFE-Time with fine-tuning yields additive gains, suggesting that learned normalizations and model adaptation provide complementary benefits. EFE-Tab, in turn, learns compact and parsimonious feature-engineering programs that substantially improve simple decision trees while preserving interpretability. Our paper also points to several promising directions for future work. First, the evolutionary process currently uses fixed hyperparameters, such as the exploration--exploitation ratio or the number of evolutionary islands. This design keeps the experimental protocol simple and controlled, but future work could study how these choices affect performance. Second, the quality and structure of feedback play an important role in guiding program evolution. A detailed ablation of which feedback components are most useful, and when they help or hinder refinement, would provide a clearer understanding of the mechanisms driving EFE’s improvements.

%% file: sections/acknowledgements.tex
\section*{Acknowledgments}
\label{sec:ack}

We thank Halil Alperen Gozeten of the University of Michigan for discussions on LLM-based evolutionary discovery. This work was supported by the National Science Foundation under grants CCF-2046816, CCF-2403075, and CCF-2212426; by the Office of Naval Research under grant N000142412289; and by Los Alamos National Laboratory under grant AWD026741 at the University of Michigan. The computational aspects of this research were generously supported by resources provided through the Amazon Research Award on Foundation Model Development.

%% file: sections/related_work_edit.tex
\section{Extended Related Work}
\label{app:relatedWork}

\textbf{Classical Automated Feature Engineering Methods.}
Early work on automated feature engineering for tabular data has largely relied on predefined transformation operators combined with search or selection strategies. Deep Feature Synthesis generates features from relational data by applying aggregation and transformation primitives~\citep{kanter2015deep}, whereas Cognito uses a greedy search procedure to explore feature transformations that improve supervised learning performance~\citep{khurana2016cognito}. AutoFeat constructs and selects nonlinear feature combinations~\citep{horn2019autofeat}, while learning-based methods, including LFE and reinforcement-learning approaches, aim to automatically identify useful transformations~\citep{nargesian2017learning,zhang2019automatic}. Despite their effectiveness, these methods typically make limited use of rich dataset-level context, such as column semantics or task descriptions, and are instead primarily guided by fixed operators and heuristic search procedures.

\textbf{Evolutionary Optimization With LLMs.}
Recent studies have investigated the use of LLMs to guide evolutionary optimization for discovering programs, prompts, and algorithms. FunSearch evolves candidate functions by combining LLM-generated proposals with evaluator feedback~\citep{romera-paredes2024funsearch}, while AlphaEvolve extends this paradigm to full-program optimization through iterative code modification and evaluation~\citep{novikov2025alphaevolve}. OpenEvolve offers an open-source framework for this code search~\citep{openevolve}. Other approaches further adapt the search process itself: EvoX evolves both candidate solutions and search strategies~\citep{liu2026evox}, and AdaEvolve leverages LLM feedback for adaptive zeroth-order optimization~\citep{cemri2026adaevolve}. LLM-based search has also been applied to scientific software generation~\citep{aygun2025ai} and reflective prompt evolution~\citep{agrawal2025gepa}. However, these works primarily focus on general-purpose program, prompt, or algorithm discovery, rather than feature engineering for structured data.

\textbf{Evolutionary Feature Engineering for Structured Data with LLMs.}
More recent work has explored the use of LLMs for automated feature engineering on structured data. CAAFE iteratively prompts an LLM with dataset descriptions to propose semantically meaningful feature interactions and selects features one at a time based on changes in cross-validation performance~\citep{caafe}. OCTree combines LLM reasoning with feedback from shallow decision trees fitted to the data, a direction that also informs our use of tree-based dataset reasoning~\citep{octree}. Most closely related to our work, LLM-FE adopts a FunSearch-style evolutionary search framework that incorporates dataset context to optimize tabular feature-transformation programs~\citep{llmfe,romera-paredes2024funsearch}. However, LLM-FE evolves a single transformation function, rather than an AlphaEvolve-style full program that includes state fitted on the training data and an explicit inverse transformation~\citep{novikov2025alphaevolve}. In the time-series setting, ELATE uses evolutionary LLM search to generate predictive covariates, but it does not focus on invertible normalizations, and no open-source implementation is available for direct comparison~\citep{murray2025elateevolutionarylanguagemodel}. To the best of our knowledge, our work is the first to evolve invertible time-series normalization modules.

\section{Broader Impact}
We develop a methodology for LLM-based evolutionary feature engineering. A potential risk of using LLMs for automated feature engineering is that they may introduce bias into the preprocessed data representations. This risk is relatively limited for EFE-Time, which learns invertible time-series normalizations, but is more pronounced for EFE-Tab, which operates directly on tabular features. In particular, semantic cues from column names may lead the LLM to generate spurious or discriminatory features. Therefore, EFE-Tab should be supervised carefully, and generated preprocessing programs should only be used after human review. 

%% file: sections/appendix.tex
\section{Additional Experimental Details}
\label{app:experimental_details}

We give reproducibility details that are not explicit in the main text in
Section~\ref{sec:experimental_setup}. 

\subsection{Evolutionary optimization interface}
\label{app:evolution_interface}

The OpenEvolve-specific configuration uses its island population and MAP-Elites archive, with
code-diversity coordinates as the archive descriptor. Candidate proposals are complete program
rewrites rather than parameter updates. The optimizer records both valid and invalid attempts so
that later prompts can use failures as debugging feedback, but only valid programs are eligible for
final selection. After the evolution budget is exhausted, the selected source code is frozen and
reloaded by a separate held-out evaluator.

Prompts contain the parent program, a small set of previous programs or summaries selected by the
optimizer, dataset-level context, and aggregate evaluator feedback. They do not contain evaluator
source code, private scoring logic, held-out examples, or test-set results. 

\subsection{EFE-Time Evolution}
\label{app:efetime_details}

\textbf{Evolution Design.}  Each time-series candidate implements \texttt{fit}, \texttt{transform}, and
\texttt{inverse\_transform}. At a forecast origin, \texttt{fit(y\_hist, meta)} receives a single
observed history and a compact metadata dictionary. The metadata keys are: prediction length,
seasonal period, history length, number of valid observations, mean, standard deviation, median,
minimum, maximum, range, positive fraction, zero fraction, skewness, coefficient of variation,
trend strength, recent mean, and recent standard deviation. The returned state dictionary must
include \texttt{hist\_len}. This is required because \texttt{inverse\_transform} is called on
forecast-horizon arrays rather than on the original history; any detrending, time-varying scaling,
or seasonal-phase operation must therefore align forecast positions using the stored history length.

The validity gate checks the forward and inverse transforms before forecasting. A valid transform
must preserve length, order, and NaN locations; avoid Inf values or new NaNs; remain numerically
stable on forecast medians and quantiles; and avoid mapping nonconstant histories to constants. The
inverse must be well defined on arrays of arbitrary forecast length, not only on arrays of length
\texttt{hist\_len}.

The prompt-visible \texttt{dataset\_context} is fixed once per dataset from the training split. It
contains only aggregate information such as frequency, horizon, seasonal period, number of
evaluation series, and distributional summaries of histories. The per-iteration
\texttt{evaluation\_summary} is also aggregate: valid candidates receive baseline and candidate
metrics, metric ratios, valid-series counts, error counts, timing, and help/harm rates; invalid
candidates receive capped transformation or inverse-transformation errors. The LLM never sees raw
series values, future values, sampled evaluation windows, model forecasts, per-series errors,
baseline forecasts, scoring code, or test-set information.

\textbf{Hyperparameters.} All experiments use fixed Openevolve hyperparameters. We used Openevolve evolution with 3 islands, with exploitation ratio of 0.70, exploration ratio of 0.10, and elite selection ratio of 0.20. We give the system prompt we have used at Section \ref{app:prompt_example_covid}. We have a code diversity axis with 8 bins. We have used Claude Opus-4.6 through AWS API access with temperature of 0.7. While calculating the combined score, to let invalid programs live so that they could be fixed, we added an offset of 0.5 to the combined score so that bad performing programs won't get the same score as failing ones through validity checks. Therefore, an identity program corresponds to a combined score of 0.5. For each dataset, we extracted 3 validation portions from the training portion of the data with non-overlapping windows, using the given forecast length of the data. At each Openevolve iteration, we use 1/3 of total extracted samples, matching the original dataset size. For fine-tuning experiments with Chronos-2, we have used the official Chronos script from their github. We used Chronos in its univariate mode, forecasting each time series independently. 

\subsection{EFE-Tab Evolution}
\label{app:efetab_details}

\textbf{Evolution Design.} Beyond the Pool A/Pool B protocol described in Section~\ref{sec:experimental_setup}, the tabular
evaluator exposes a structured \texttt{dataset\_context} computed only from Pool A. It includes the
dataset name, shape, target column, problem type, class balance, task description, column-level
statistics, target correlations, top inter-feature correlations, and a small number of Pool A rows.
For numerical columns, the context reports ranges, quartiles, skewness, cardinality, and missingness, while
for non-numerical columns, it reports cardinality, top values, and missingness. Since Pool A rotates
across iterations, these summaries vary slightly during evolution.

The stage-one validity gate checks that the program has valid function signatures, returns
DataFrames for both pools, preserves row alignment, avoids in-place mutation of the input frames,
and produces a schema that can be applied consistently to Pool A and Pool B. Target-dependent operations such as target encodings or group aggregates are permitted only through state fitted on Pool A and then reused unchanged on Pool B.

For valid candidates, \texttt{stage2\_summary} reports the baseline AUC, candidate AUC, AUC
difference, combined fitness score, pool sizes, iteration seed, newly created columns, dropped
original columns, output-column counts, and validation status. The evaluator may also return
Pool-A-only interpretability feedback, including permutation-importance labels, an original-column
importance report, and a readable decision tree trained on the candidate representation. Pool B is
visible to the LLM only through aggregate AUC values and their difference. Raw or transformed Pool B
rows, per-row predictions, per-row losses are not exposed.

The complexity term counts both added columns and dropped original columns as feature
modifications. This makes the objective prefer feature programs whose held-out AUC gain is large
enough to justify their additional representation cost and runtime.

\textbf{Hyperparameters.} Just like EFE-Time, all experiments use fixed Openevolve hyperparameters. We used Openevolve evolution with 3 islands, with exploitation ratio of 0.70, exploration ratio of 0.10, and elite selection ratio of 0.20. Thus, all the hyperparameters with respect to Openevolve are the same with EFE-Time. During evolution, we used 70\% of the training data for Pool A, and the rest 30\% of the data for pool B. At each Openevolve iteration, pool A and pool B are resampled again. 

\subsection{Benchmarks and Reporting Conventions}
\label{app:benchmarks_reporting}

For EFE-Time, each dataset is evolved under three random seeds; reported MASE, wQL, and MAE values are means and standard deviations across those runs. Transfer experiments reuse the frozen programs learned with Chronos-2 without re-evolving or tuning them for the target forecaster.

For EFE-Tab, TabArena provides three repetitions of three-fold splits for the medium-sized datasets.
Using all repetitions with three EFE seeds would require $27$ evolution/evaluation runs per dataset,
so the main benchmark uses the three folds from the first official repetition. Baseline comparisons
against CAAFE and LLM-FE use the same 100-iteration LLM-query budget as EFE-Tab. Frozen feature
programs are then evaluated with TabPFN, LightGBM, and single decision trees.

\subsection{The List of Datasets Used}

\begin{table}[htbp]
\centering
\caption{EFE-Time datasets used in the evaluation.}
\label{tab:gift_eval_datasets}
\resizebox{\textwidth}{!}{%
\begin{tabular}{llccrcr}
\toprule
\textbf{User label} & \textbf{Official GIFT-Eval name} & \textbf{Freq.} & \textbf{Pred. len.} & \textbf{\# series (train)} & \textbf{Series length (min / median / max)} & \textbf{\# test instances} \\
\midrule
CovidDeaths & \texttt{covid\_deaths} & D & 30 & 266 & 152 / 152 / 152 & 266 \\
SolarHourly & \texttt{solar/H} & H & 48 & 137 & 7,800 / 7,800 / 7,800 & 2,603 \\
M4Yearly & \texttt{m4\_yearly} & A & 6 & 22,974 & 7 / 23 / 272 & 22,974 \\
JenaWeather (H) & \texttt{jena\_weather/H} & H & 48 & 21 (univariate explode) & 7,824 / 7,824 / 7,824 & 399 \\
SZ\_TAXI\_15T & \texttt{SZ\_TAXI/15T} & 15min & 48 & 156 & 2,592 / 2,592 / 2,592 & 1,092 \\
bitbrains\_storage/H & \texttt{bitbrains\_fast\_storage/H} & H & 48 & 2,500 (1,250 $\times$ \{read, write\}) & 577 / 577 / 577 & 5,000 \\
bitbrains\_rnd/H & \texttt{bitbrains\_rnd/H} & H & 48 & 1,000 (500 $\times$ \{read, write\}) & 576 / 576 / 576 & 2,000 \\
us\_births\_D & \texttt{us\_births/D} & D & 30 & 1 & 6,675 / 6,675 / 6,675 & 20 \\
kdd\_cup\_2018\_H & \texttt{kdd\_cup\_2018\_with\_missing/H} & H & 48 & 270 & 8,496 / 9,890 / 9,912 & 5,400 \\
restaurant & \texttt{restaurant} & D & 30 & 807 & 7 / 236 / 418 & 807 \\
\bottomrule
\end{tabular}%
}
\end{table}

\begin{table}[htbp]
\centering
\caption{The TabArena datasets used in the EFE-Tab evaluation.}
\label{tab:tabarena_datasets}
\resizebox{\textwidth}{!}{%
\begin{tabular}{llrrr}
\toprule
\textbf{User label} & \textbf{TabArena Name} & \textbf{OpenEvolve ID} & \textbf{Rows} & \textbf{Feat.} \\
\midrule
Churn & \texttt{churn} & 363623 & 5,000 & 19 \\
HRAnalyticsJobChange & \texttt{HR\_Analytics\_Job\_Change\_of\_Data\_Scientists} & 363679 & 19,158 & 12 \\
E-CommereShippingData & \texttt{E-CommereShippingData} & 363632 & 10,999 & 10 \\
in\_vehicle\_coupon\_recommendation & \texttt{in\_vehicle\_coupon\_recommendation} & 363681 & 12,684 & 24 \\
online\_shoppers\_intention & \texttt{online\_shoppers\_intention} & 363691 & 12,330 & 17 \\
Bank\_Customer\_Churn & \texttt{Bank\_Customer\_Churn} & 363619 & 10,000 & 10 \\
BankMarketing & \texttt{bank-marketing} & 363618 & 45,211 & 13 \\
Diabetes & \texttt{diabetes} & 363629 & 768 & 8 \\
Fitness\_Club & \texttt{Fitness\_Club} & 363671 & 1,500 & 6 \\
\bottomrule
\end{tabular}%
}
\end{table}

\subsection{Synthetic Smooth Exponential Time-Series Generator}
\label{app:exp-synth}

We generate synthetic exponential-growth time series designed to isolate smooth multiplicative growth from transient spike effects. 

For each series, let the sequence length be $L$ and define $T=L-1$, with time index
\[
t \in \{0,1,\ldots,T\}.
\]
We first sample a starting value $s$ and an overall growth ratio $r$ log-uniformly:
\[
s \sim \mathrm{LogUniform}(s_{\min}, s_{\max}),
\]
\[
r \sim \mathrm{LogUniform}(r_{\min}, r_{\max}).
\]
Equivalently,
\[
\log s \sim \mathrm{Uniform}(\log s_{\min}, \log s_{\max}),
\]
\[
\log r \sim \mathrm{Uniform}(\log r_{\min}, \log r_{\max}).
\]
The total log-growth over the full sequence is therefore
\[
G = \log r.
\]

The generator constructs a monotone trajectory in log space. Let
\[
\eta_i = \mathrm{SmoothNoise}(T,\sigma),
\qquad i=1,\ldots,T,
\]
where $\mathrm{SmoothNoise}$ denotes Gaussian white noise smoothed by convolution with a one-dimensional Gaussian kernel of width controlled by $\sigma$, then normalized to have approximately zero mean and unit root-mean-square magnitude. We convert this smooth noise into positive growth weights:
\[
w_i = \exp(\gamma \eta_i),
\]
where $\gamma$ controls the volatility of the growth rate. These weights are normalized so that the total log-growth is exactly $G$:
\[
\Delta \ell_i
=
\frac{w_i}{\sum_{j=1}^{T} w_j}G.
\]
The latent log-trend is then
\[
\ell_0 = \log s,
\]
\[
\ell_t
=
\log s
+
\sum_{i=1}^{t} \Delta \ell_i,
\qquad t=1,\ldots,T.
\]
The corresponding latent exponential trend is
\[
m_t = \exp(\ell_t).
\]
Because each $\Delta \ell_i$ is positive, the latent trend is monotone increasing. In addition, the normalization enforces exact endpoint control:
\[
m_0 = s,
\]
\[
m_T = sr.
\]

For reference, we also define the clean log-linear exponential baseline
\[
c_t = s r^{t/T},
\]
which is the trajectory obtained by distributing the total log-growth uniformly over time.

Finally, the observable time series is obtained by applying smooth multiplicative observation noise. Let
\[
\varepsilon_t
=
\alpha \, \mathrm{SmoothNoise}(L,\sigma/2),
\]
where $\alpha$ controls the observation noise scale. The final observed sequence is
\[
x_t
=
m_t \exp(\varepsilon_t).
\]
Equivalently,
\[
x_t
=
\exp(\ell_t + \varepsilon_t).
\]

Combining the above definitions, the generated series can be written as
\[
x_t
=
\exp\left(
\log s
+
\sum_{i=1}^{t}
\frac{\exp(\gamma \eta_i)}
{\sum_{j=1}^{T} \exp(\gamma \eta_j)}
\log r
+
\varepsilon_t
\right)
\]
for $t=0,\ldots,T$, with the convention that the summation term is zero when $t=0$.

In the experiments, the default parameter ranges are
\[
s \sim \mathrm{LogUniform}(0.5,2.0),
\]
\[
r \sim \mathrm{LogUniform}(5000, 8000),
\]
\[
\gamma \sim \mathrm{Uniform}(0.2,1.2),
\]
\[
\alpha \sim \mathrm{Uniform}(0.0,0.08),
\]
\[
\sigma \sim \mathrm{Uniform}(8.0,32.0).
\]
Each time series draws its own independent values of $s$, $r$, $\gamma$, $\alpha$, and $\sigma$. 

\subsection{Compute}

Except for running local LLMs detailed in Section \ref{app:efetime_local_model}, we have only used 4 L40S GPUs throughout experiments, with 8 CPU cores and 128 GB of memory. 

\subsection{Local-model evolution}
\label{app:efetime_local_model}

The main paper used a hosted frontier LLM to drive evolution. To test whether the same loop also works with smaller open-weight models, we repeated the experiment on $\mathtt{covid\_deaths}$ and $\mathtt{solar\_H}$ using four Qwen3.5 sizes: 4B, 9B, 27B, and 35B-A3B. All non-model settings were held fixed. The 4B and 9B models were served locally with vLLM on one RTX A6000, while 27B and 35B-A3B used two A6000s or a hosted Qwen endpoint. Each setting was run for 300 iterations with three random seeds. 

Table~\ref{tab:appendix_local_model} reports the final optimizer-selected program for each setting. On $\mathtt{covid\_deaths}$, every model size found useful MASE improvements, but larger models were not consistently better. The 9B model had the largest mean MASE gain ($+5.3 \pm 3.3\%$), while 4B also improved and matched the larger models on its best seed. However, some configurations produced poorly calibrated quantiles: 9B and 35B-A3B both showed large WQL regressions, even when their median forecasts improved. The 27B model was more stable in WQL but had the smallest MASE gain. Overall, performance did not scale monotonically with model size.

The $\mathtt{solar\_H}$ results were cleaner. All four sizes improved MASE, WQL, and MAE on average, with no calibration failures like those seen on $\mathtt{covid\_deaths}$. The 9B and 35B-A3B models performed best, reaching about $+11\%$ mean MASE improvement, while 4B still achieved a strong mean gain of $+7.8 \pm 6.9\%$. The 27B model was less strong but more consistent. Much of the variance came from whether a seed discovered a high-performing program: successful seeds reached a $\sim$$15\%$ improvement regime, while weaker seeds stayed close to baseline. As before, gains did not follow a smooth model-size scaling trend.

Yet, in the end,  our results demonstrate that end-to-end program evolution, with our existing framework and large context sizes provided to LLMs, is not stable for non-frontier, small-scale models. 

\input{tables/appendix_local_model}

\begin{figure}[!t]
\centering
\includegraphics[width=\linewidth]{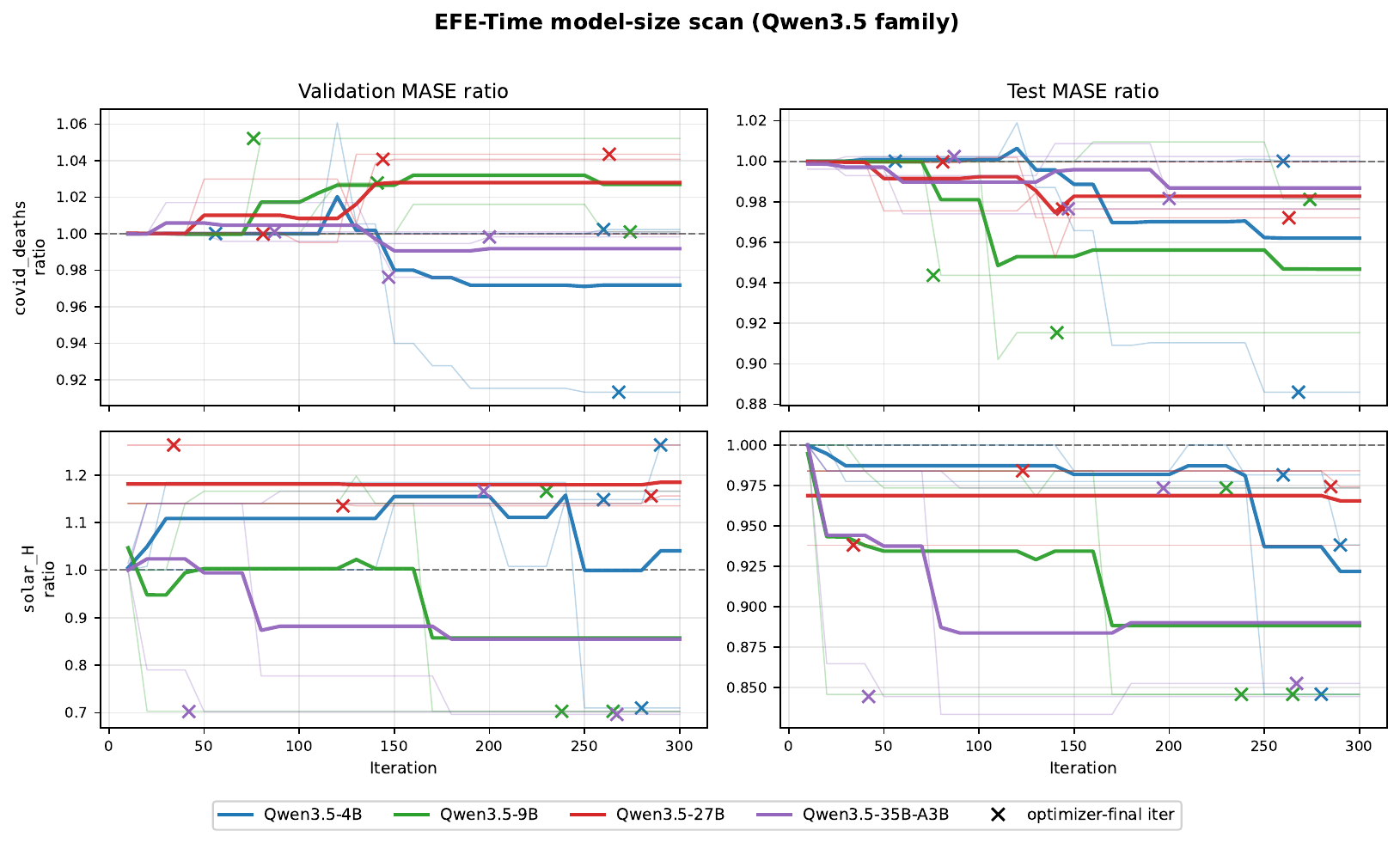}
\caption{\small{Per-checkpoint validation and test MASE ratios for the $\mathtt{covid\_deaths}$ and $\mathtt{solar\_H}$ model-size scans. Faint traces are individual seeds; solid lines are per-size means; $\times$-marks denote the iteration at which each run's optimizer-final program was first found. Values below the dashed line ($\mathrm{ratio}{=}1$) improve over the Chronos-2 identity-transform baseline.}}
\label{fig:appendix_local_model_overlay}
\end{figure}

\section{Extended Results}
\begin{figure}[t!]
    \centering
    \includegraphics[width=1\linewidth]{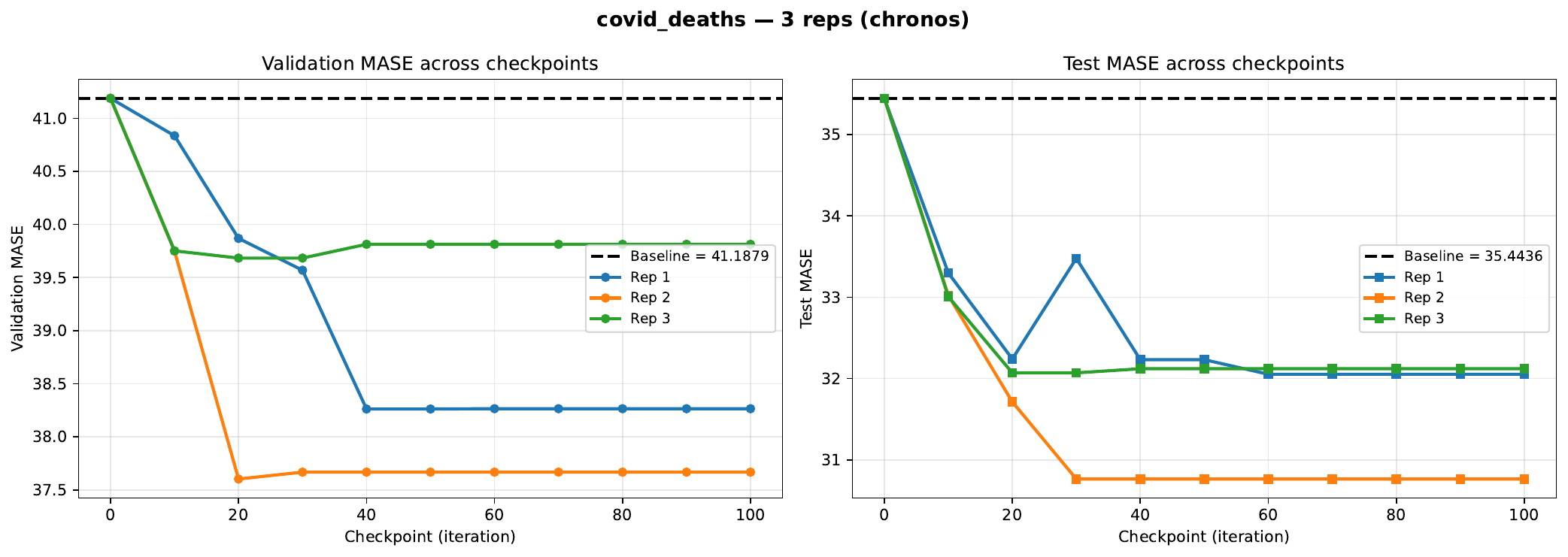}
    \caption{The change of validation and test MASE during EFE-Time evolution for Covid-Deaths.}
    \label{fig:combined-curves-covid-deaths}
\end{figure}

\begin{figure}[t!]
    \centering
    \includegraphics[width=1\linewidth]{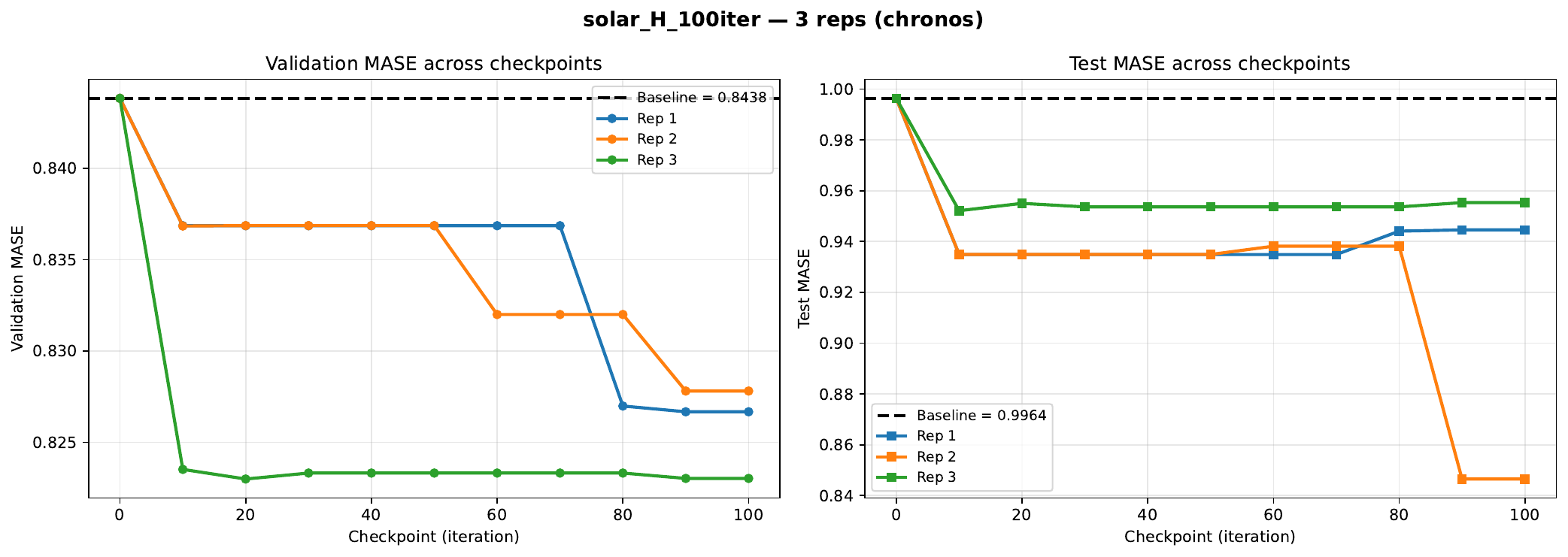}
    \caption{The change of validation and test MASE during EFE-Time evolution for Solar-hourly.}
    \label{fig:combined-curves-solar-hourly}
\end{figure}

We present here our raw results in Table \ref{app:efe_time_results} and in Table \ref{app:efe_tab_results}. We also demonstrate EFE-Time evolutions in Figure \ref{fig:combined-curves-covid-deaths} and Figure \ref{fig:combined-curves-solar-hourly} across 3 repetitions.

\clearpage
\label{app:detailed-results}

\input{tables/efetime_table_appendix}

\input{tables/efetab_table_appendix}

\clearpage

\clearpage

\section{Per-series forecast trajectories under the evolved transform}
\label{app:efetime_perseries}

This section visualizes the per-series effect of the evolved transform on a small selection of test instances drawn from each dataset in the EFE-Time evaluation suite. Each three-panel column shows, top to bottom, Chronos-2 on the raw input window, Chronos-2 in the transformed space (the forecaster sees $z = \texttt{transform}(y)$ and predicts the transformed continuation), and the inverse-transformed forecast plotted back on the original scale. The four picks per dataset include both series where the transform helps and series where it has little or no effect, so the panels should be read as qualitative inspection rather than as a ranking. On $\mathtt{covid\_deaths}$ and $\mathtt{solar\_H}$ the transformed input window often resembles a near-periodic shape that Chronos-2 extrapolates more cleanly. We point this out in the per-dataset paragraphs where it applies. A recurring theme across the harder cases is the role of exogenous, perhaps unpredictable drivers with existing information: when the dynamics that govern the held-out window are not visible in the input history, no input-only transform can recover what is missing.

\subsection{$\mathtt{covid\_deaths}$}
\label{app:perseries_covid_deaths}

\begin{figure}[!t]
  \centering
  \begin{subfigure}[t]{0.48\linewidth}\centering
    \includegraphics[width=\linewidth]{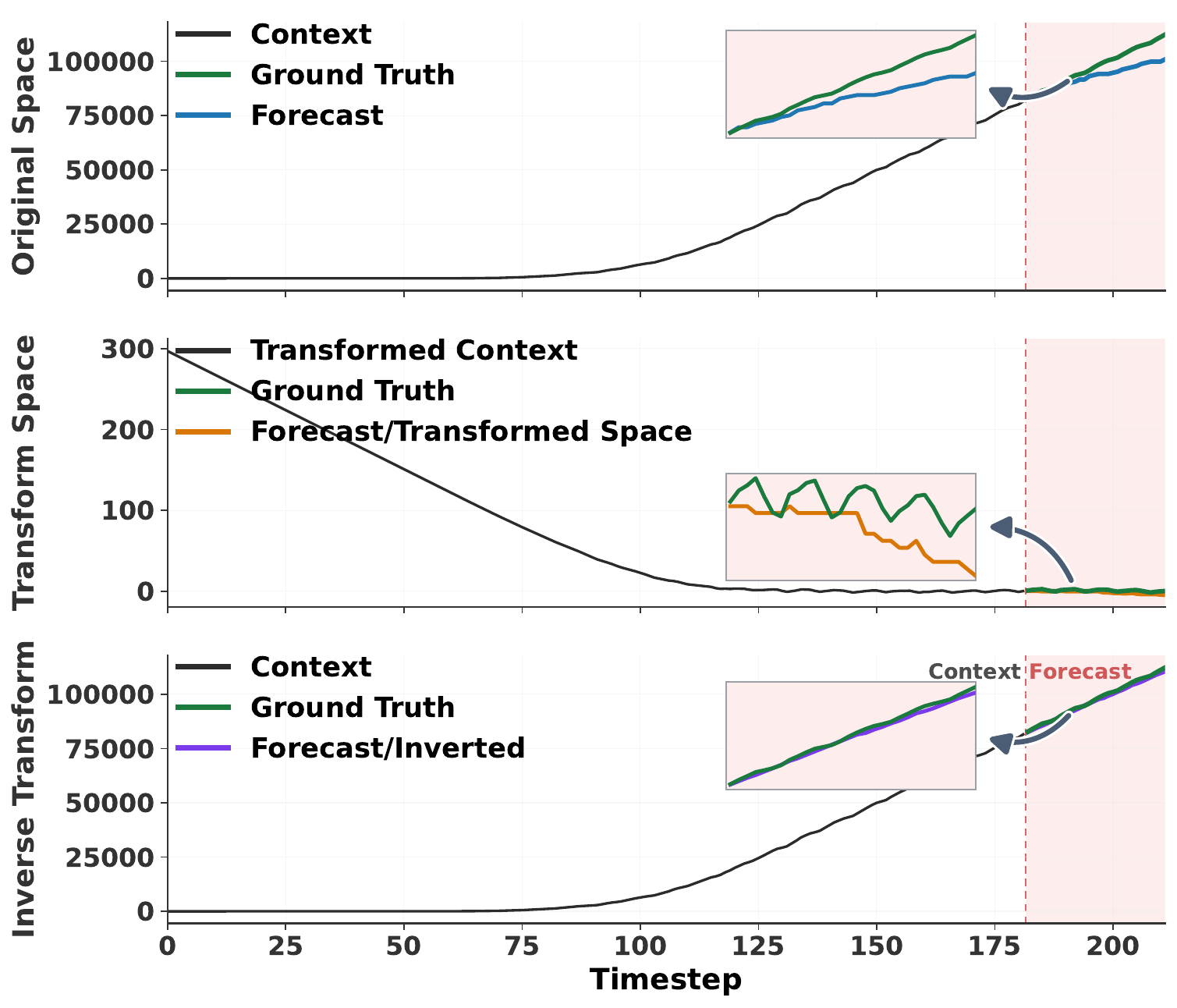}
    \caption{MASE $10.68 \rightarrow 1.63$, wQL $0.038 \rightarrow 0.006$, MAE $4810 \rightarrow 735$.}
    \label{fig:perseries_covid_30}
  \end{subfigure}\hfill
  \begin{subfigure}[t]{0.48\linewidth}\centering
    \includegraphics[width=\linewidth]{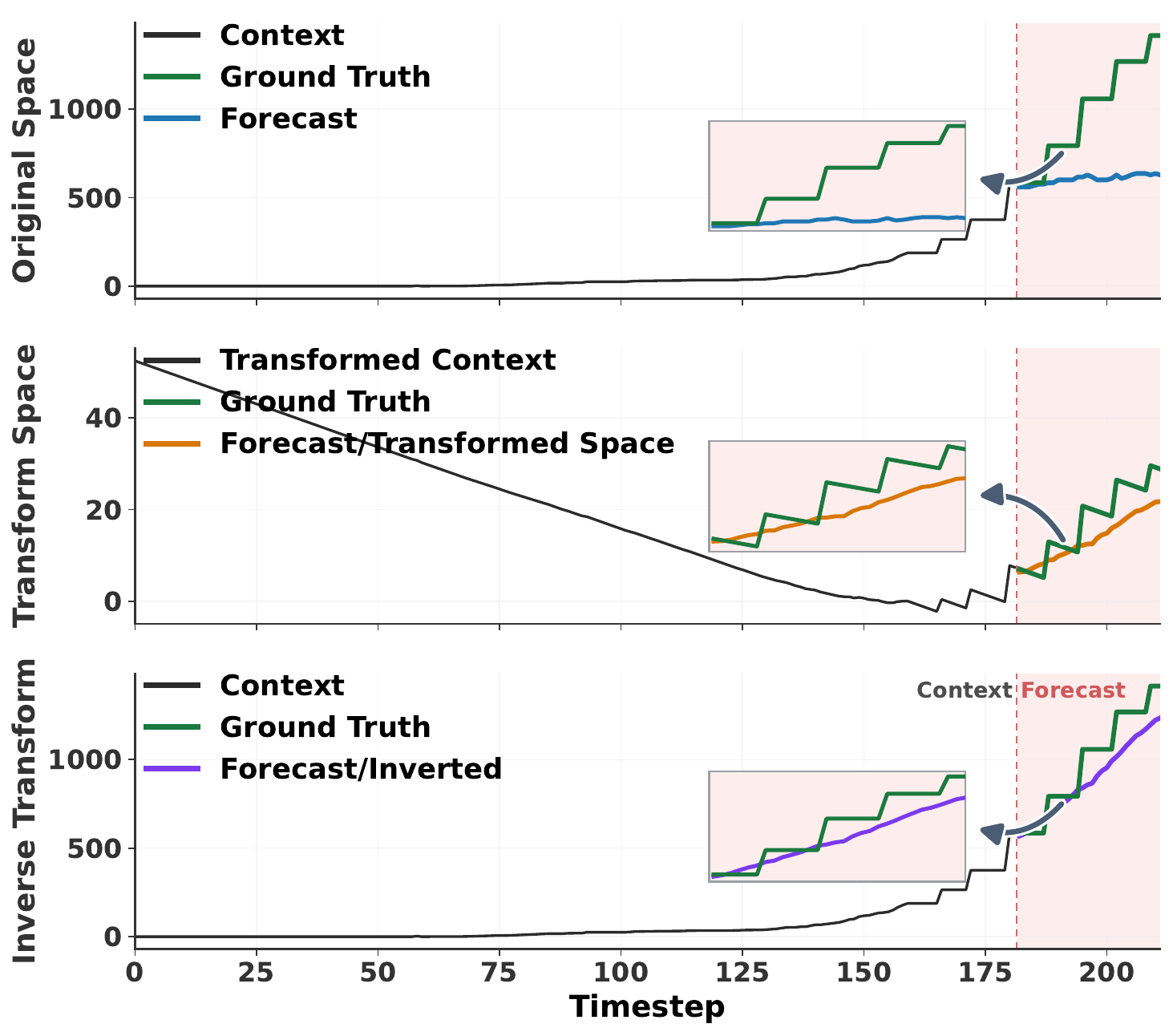}
    \caption{MASE $116.63 \rightarrow 34.50$, wQL $0.279 \rightarrow 0.107$, MAE $382.1 \rightarrow 113.0$.}
    \label{fig:perseries_covid_153}
  \end{subfigure}\\[4pt]
  \begin{subfigure}[t]{0.48\linewidth}\centering
    \includegraphics[width=\linewidth]{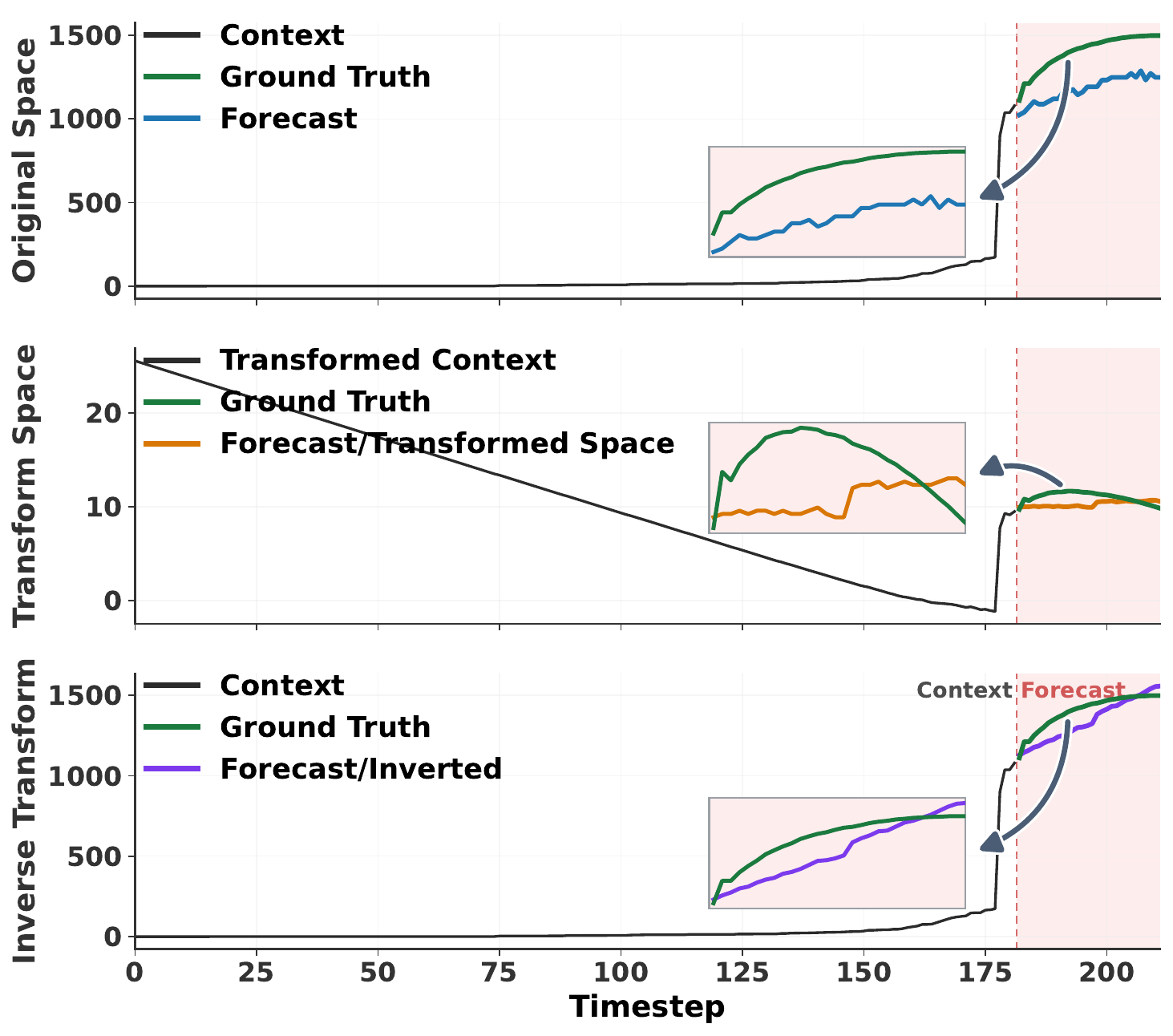}
    \caption{MASE $37.39 \rightarrow 12.09$, wQL $0.117 \rightarrow 0.057$, MAE $222.9 \rightarrow 72.1$.}
    \label{fig:perseries_covid_158}
  \end{subfigure}\hfill
  \begin{subfigure}[t]{0.48\linewidth}\centering
    \includegraphics[width=\linewidth]{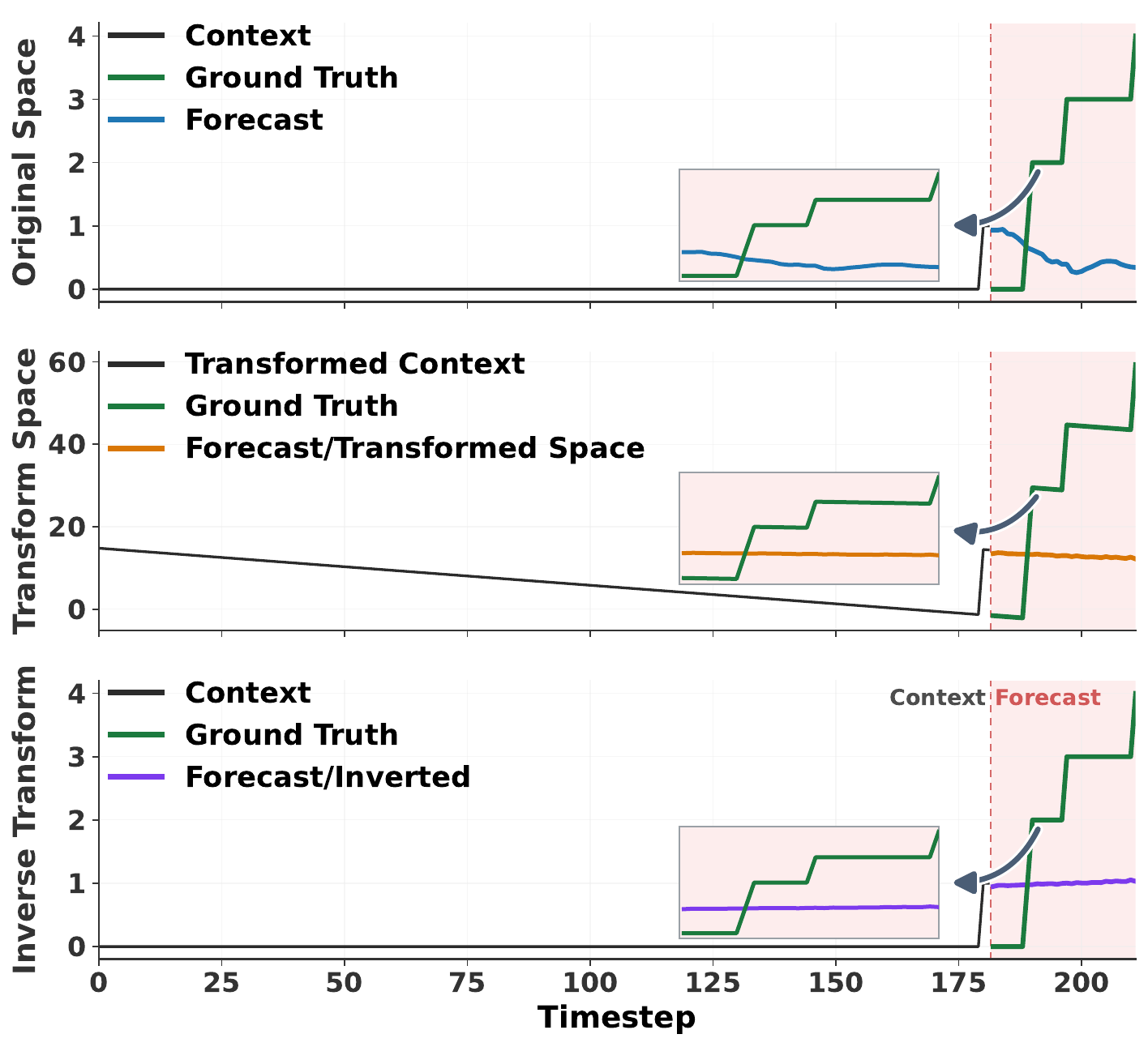}
    \caption{MASE $346.8 \rightarrow 268.9$, wQL $0.539 \rightarrow 0.697$, MAE $1.92 \rightarrow 1.49$.}
    \label{fig:perseries_covid_200}
  \end{subfigure}
  \caption{\small{Forecast trajectories on four $\mathtt{covid\_deaths}$ test instances. Each three-panel column shows, top to bottom, Chronos-2 on the raw input, Chronos-2 in the transformed space, and the inverse-transformed forecast back on the original scale. Panels (a) and (b) show series where the transform substantially reduces the inverted-forecast error. Panels (c) and (d) show series whose held-out windows contain sudden spikes; the transform partially recovers the level on (c) and improves the median MAE but not the quantile loss on (d).}}
  \label{fig:perseries_covid_2x2}
\end{figure}

The transformed input on Fig.~\ref{fig:perseries_covid_30} and Fig.~\ref{fig:perseries_covid_153} is closer to a smooth periodic signal than the raw counter is, and Chronos-2's median forecast in the transformed space follows the gentle wave the model is now seeing. The inverse step then restores the original level and brings the forecast median into the neighbourhood of the held-out trajectory. Fig.~\ref{fig:perseries_covid_158} and Fig.~\ref{fig:perseries_covid_200} display sudden spikes in the held-out window that have no precedent in the input history: a death surge of this kind is typically triggered by an exogenous event (a regional outbreak, a hospital admission wave, a reporting catch-up). The dynamics governing such jumps are non-Markovian with respect to the model's input, which places a fundamental limit on what an input-only transform can recover. The transform absorbs some of the level shift and reduces the inverted-forecast MAE on both panels, but the residual on Fig.~\ref{fig:perseries_covid_200} widens the quantile cone enough that wQL regresses even as the median improves.

\subsection{$\mathtt{m4\_yearly}$}
\label{app:perseries_m4_yearly}

\begin{figure}[!t]
  \centering
  \begin{subfigure}[t]{0.48\linewidth}\centering
    \includegraphics[width=\linewidth]{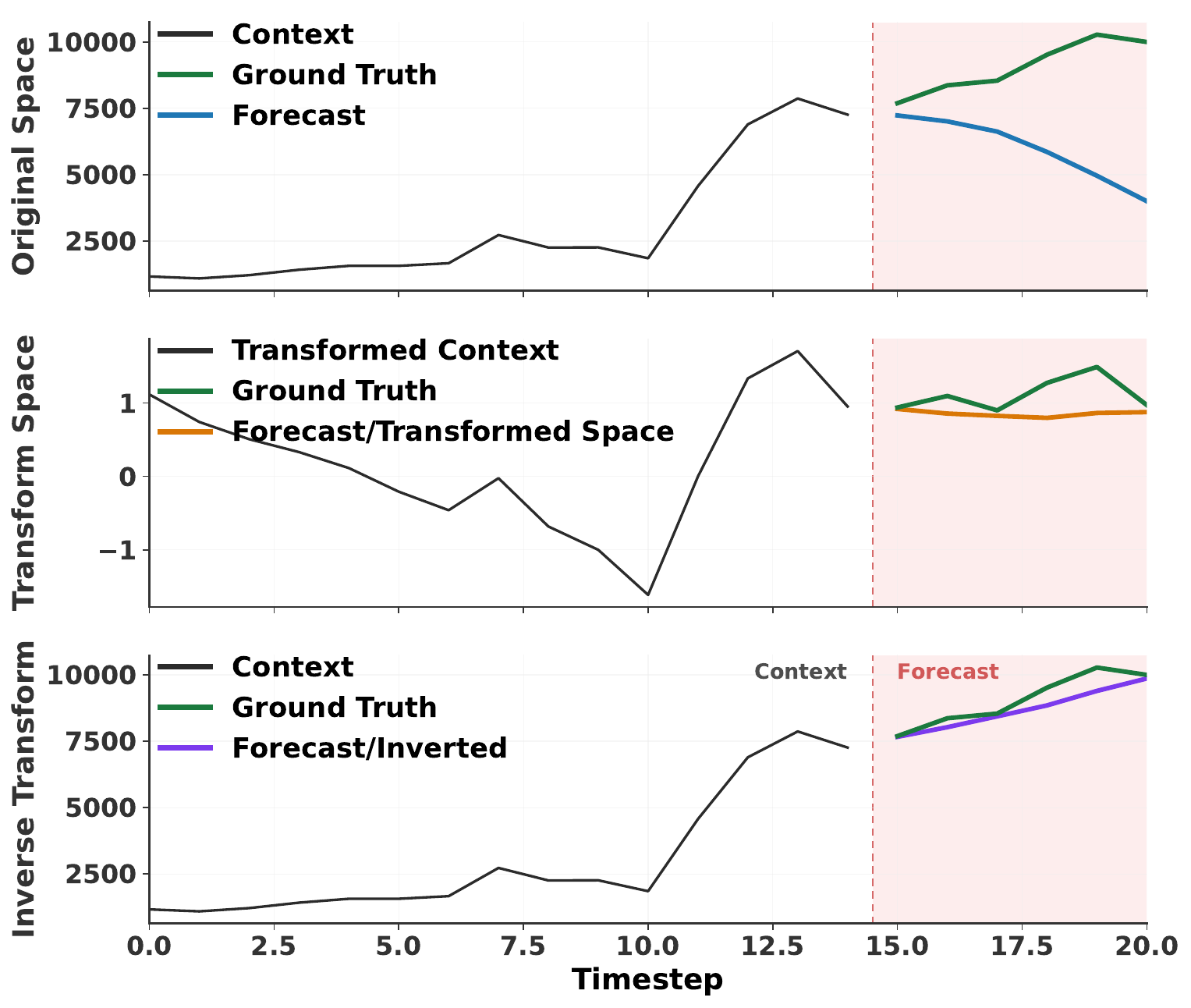}
    \caption{MASE $4.74 \rightarrow 0.55$, wQL $0.242 \rightarrow 0.052$, MAE $3120 \rightarrow 359$.}
    \label{fig:perseries_m4_12691}
  \end{subfigure}\hfill
  \begin{subfigure}[t]{0.48\linewidth}\centering
    \includegraphics[width=\linewidth]{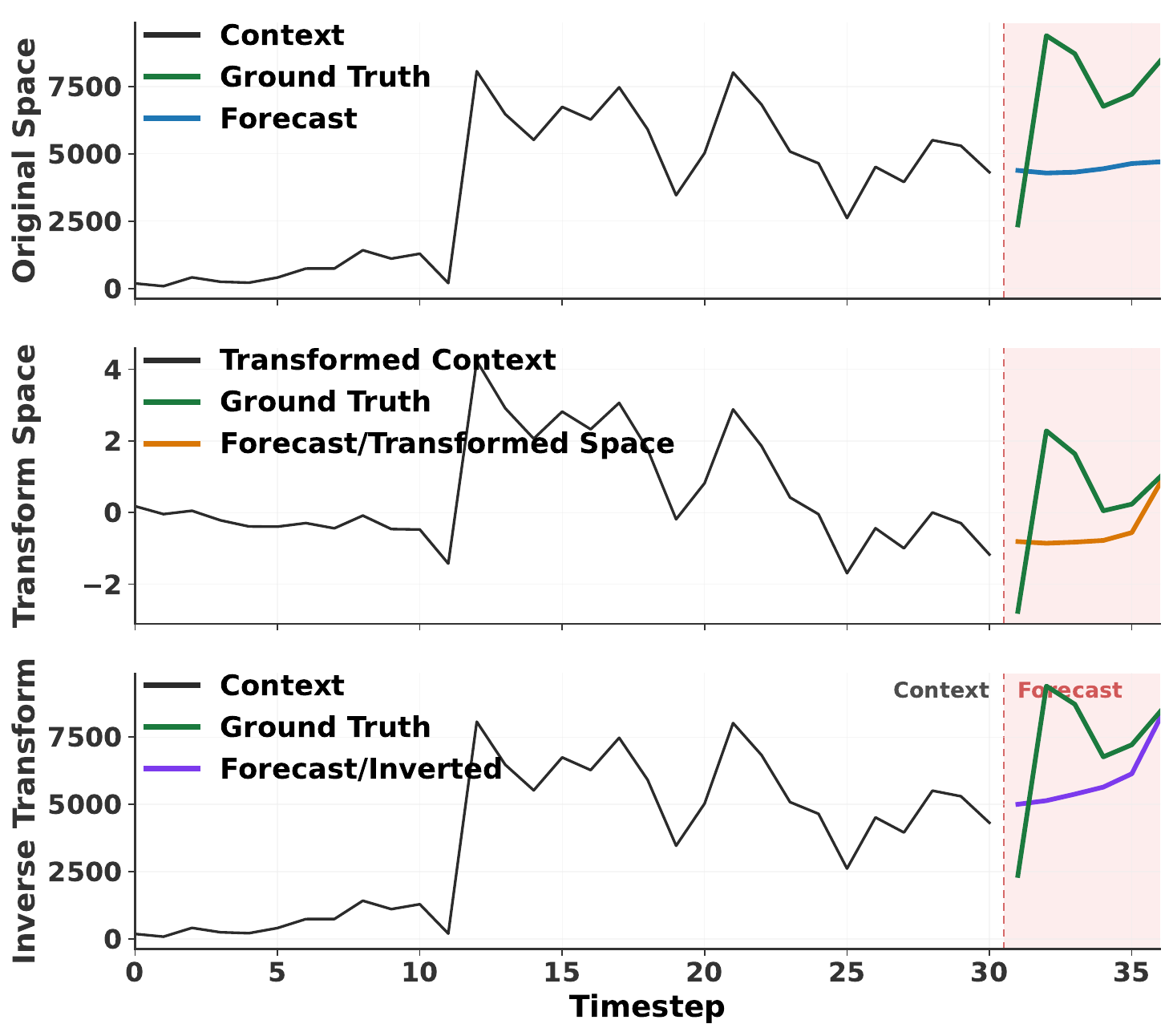}
    \caption{MASE $2.81 \rightarrow 1.77$, wQL $0.352 \rightarrow 0.256$, MAE $3359 \rightarrow 2112$.}
    \label{fig:perseries_m4_21318}
  \end{subfigure}\\[4pt]
  \begin{subfigure}[t]{0.48\linewidth}\centering
    \includegraphics[width=\linewidth]{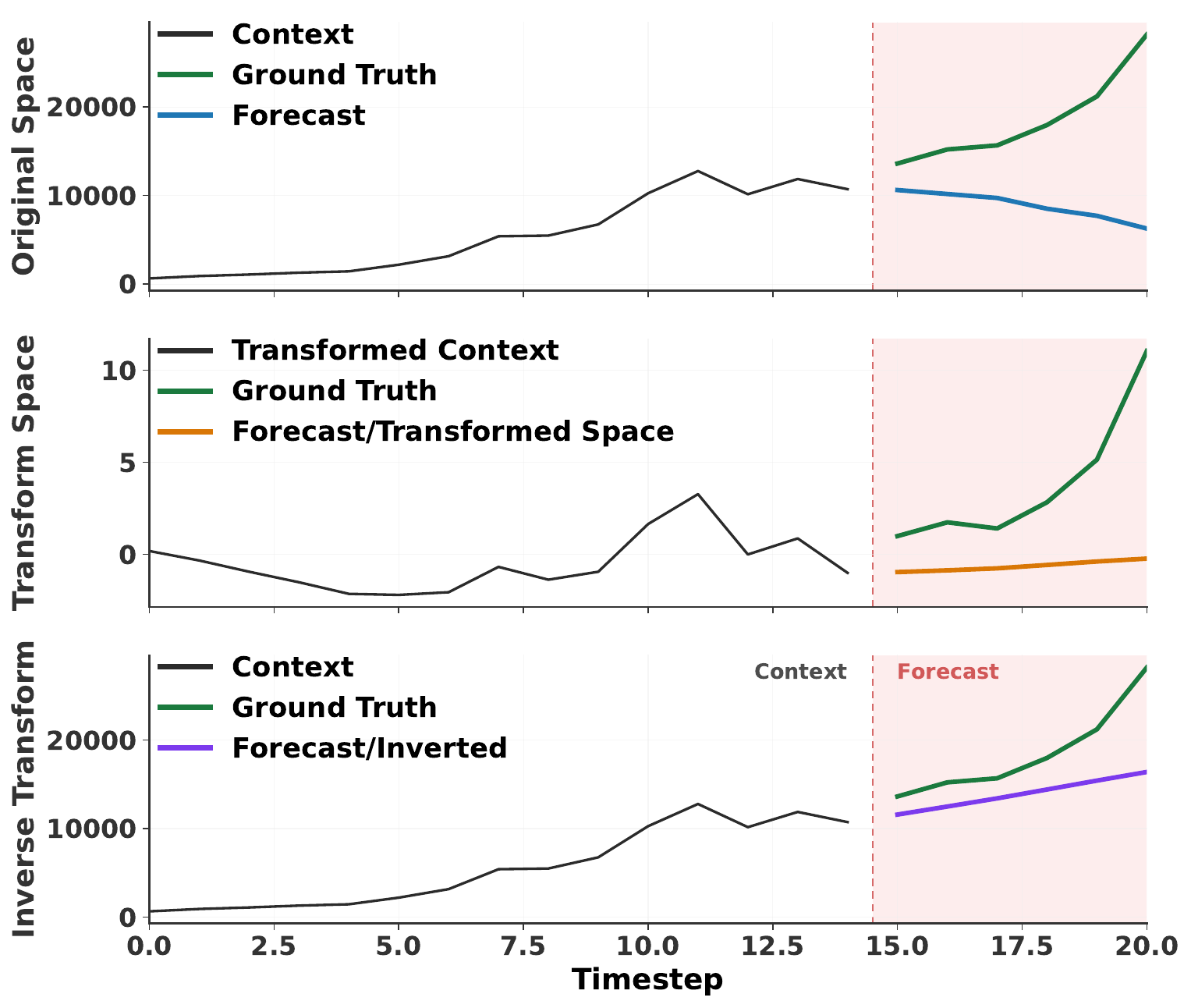}
    \caption{MASE $7.79 \rightarrow 3.73$, wQL $0.420 \rightarrow 0.197$, MAE $9804 \rightarrow 4699$.}
    \label{fig:perseries_m4_21637}
  \end{subfigure}\hfill
  \begin{subfigure}[t]{0.48\linewidth}\centering
    \includegraphics[width=\linewidth]{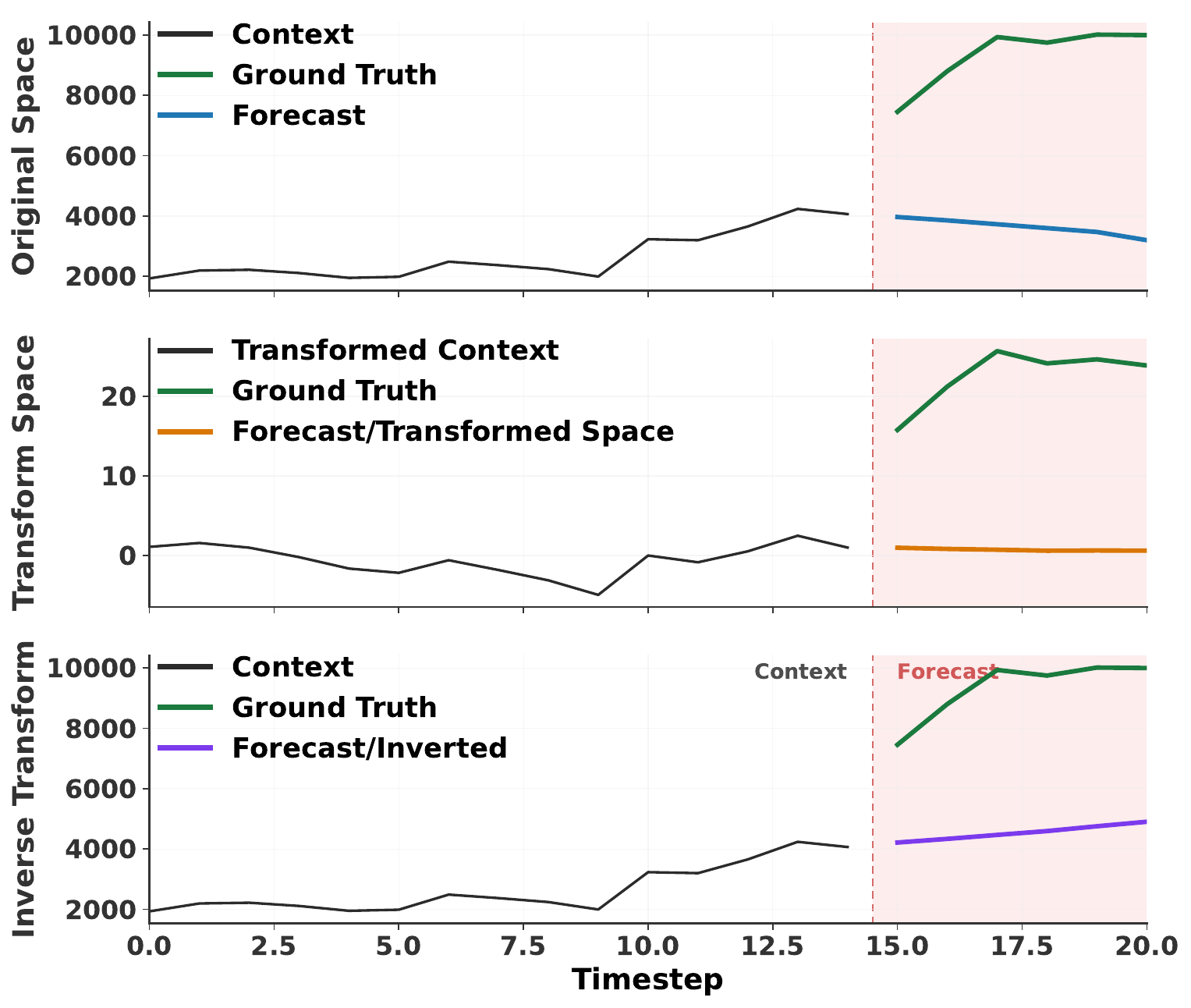}
    \caption{MASE $19.59 \rightarrow 16.46$, wQL $0.561 \rightarrow 0.490$, MAE $5690 \rightarrow 4784$.}
    \label{fig:perseries_m4_21860}
  \end{subfigure}
  \caption{\small{Forecast trajectories on four $\mathtt{m4\_yearly}$ test instances. The yearly horizon is six steps; the panels read as a per-series scale and trend correction rather than the periodic-shape rewriting visible on $\mathtt{covid\_deaths}$ and $\mathtt{solar\_H}$. The forecast-horizon zoom inset is omitted because the six-step horizon is short enough to read directly from the panels.}}
  \label{fig:perseries_m4_2x2}
\end{figure}

The M4-yearly horizon is short and the input windows are short as well. The transformed window typically rescales the trajectory's growth so that Chronos-2 in the transformed space sits closer to the ground truth's slope, and the inverse step then amplifies that corrected slope back to the original units as seen in Figure \ref{fig:perseries_m4_2x2}. We do not see the periodic transform-space shape that appears on $\mathtt{covid\_deaths}$ and $\mathtt{solar\_H}$; the effect here is closer to a learned per-series scale and trend correction. The same caveat as on $\mathtt{covid\_deaths}$ applies: the M4-yearly aggregates contain a substantial financial and macroeconomic component, and multi-year drift in those series is often driven by exogenous events (policy regimes, market shocks, demographic transitions) that the input window does not encode. Fig.~\ref{fig:perseries_m4_21860} is an example where the transform tightens the slope but the held-out trajectory continues a regime change that no input-only mapping could anticipate, leaving a residual MASE above $16$.

\subsection{$\mathtt{solar\_H}$}
\label{app:perseries_solar_H}

\begin{figure}[!t]
  \centering
  \begin{subfigure}[t]{0.48\linewidth}\centering
    \includegraphics[width=\linewidth]{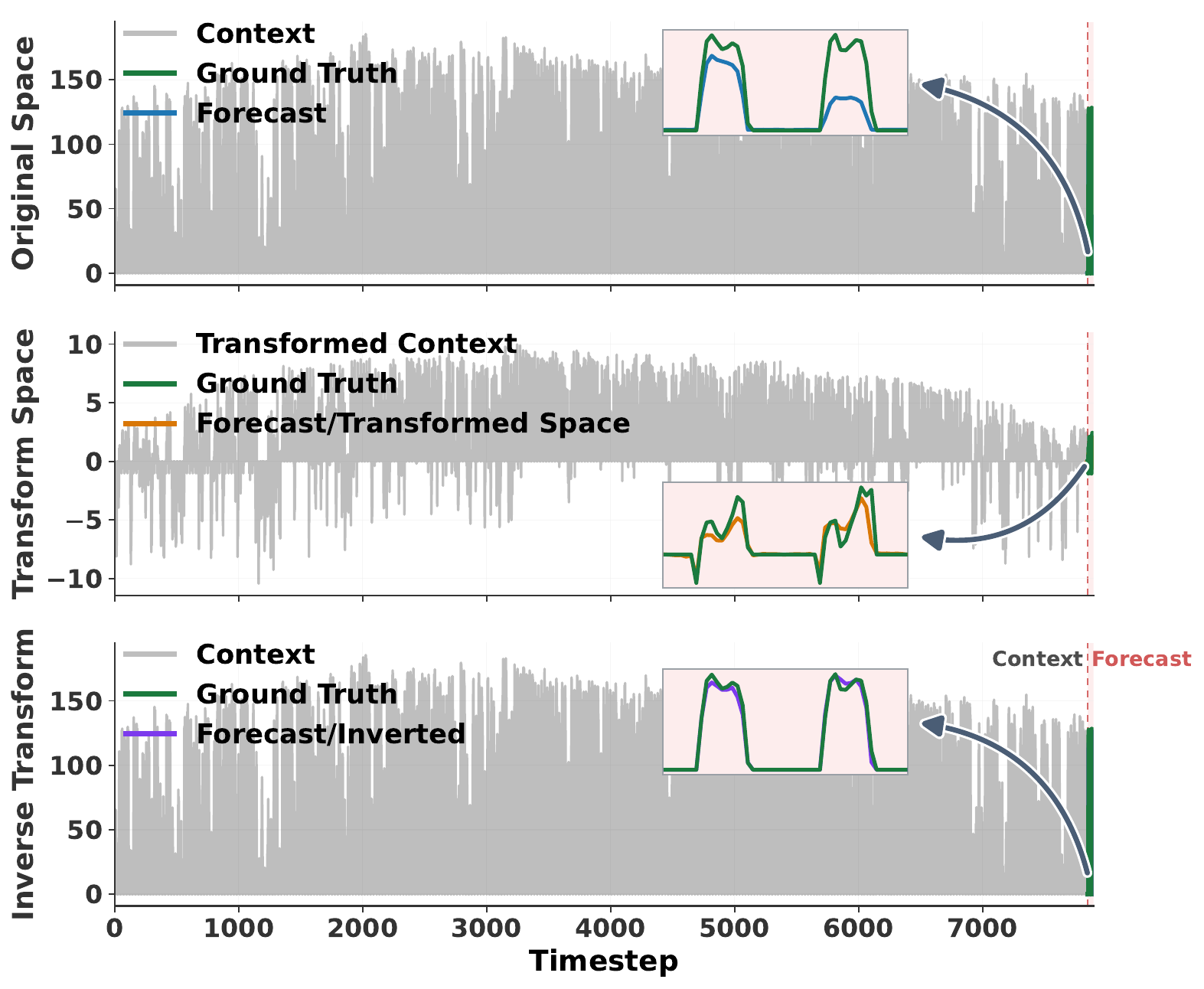}
    \caption{MASE $1.20 \rightarrow 0.16$, wQL $0.333 \rightarrow 0.084$, MAE $19.69 \rightarrow 2.67$.}
    \label{fig:perseries_solar_418}
  \end{subfigure}\hfill
  \begin{subfigure}[t]{0.48\linewidth}\centering
    \includegraphics[width=\linewidth]{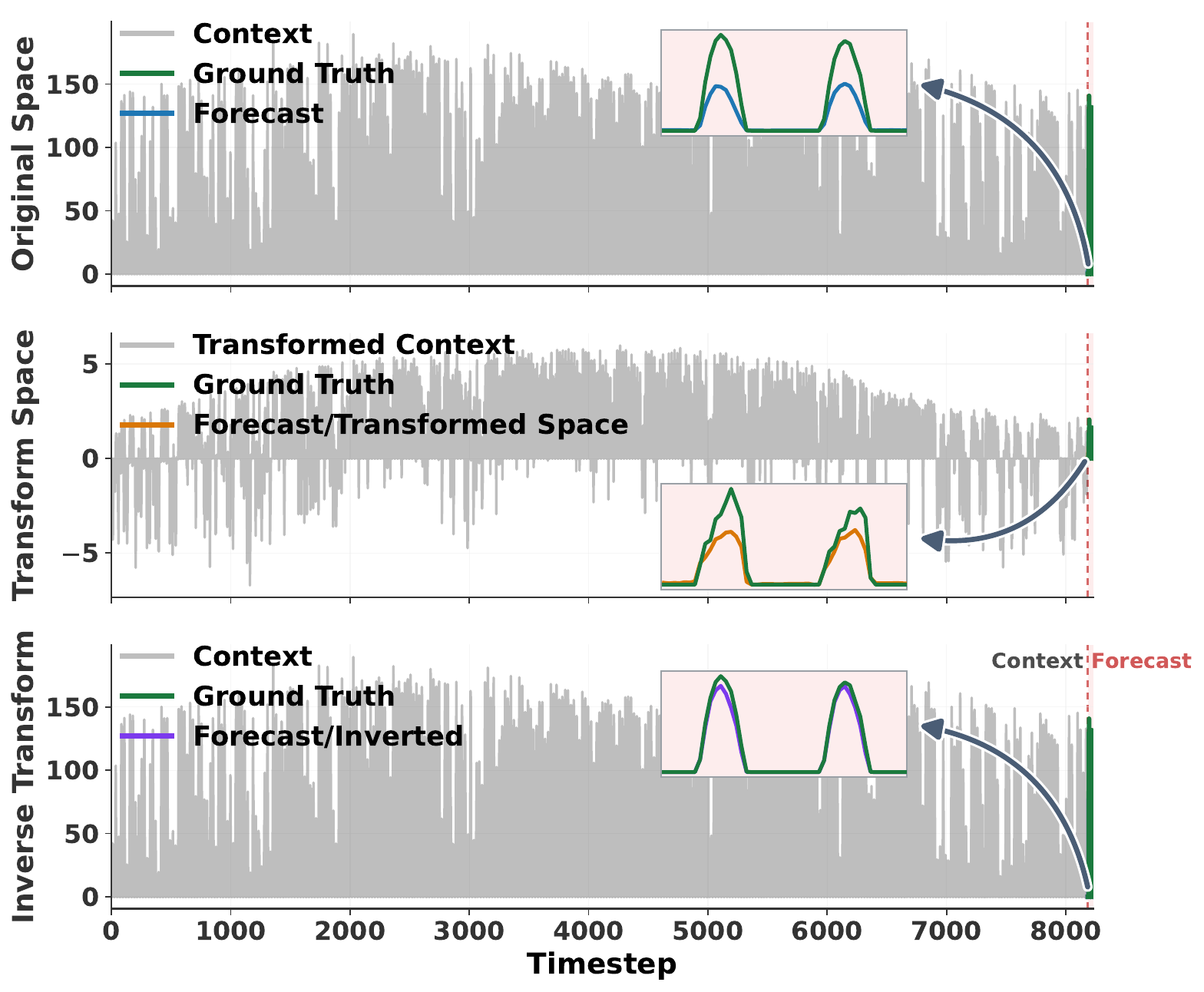}
    \caption{MASE $1.46 \rightarrow 0.28$, wQL $0.363 \rightarrow 0.097$, MAE $18.55 \rightarrow 3.53$.}
    \label{fig:perseries_solar_558}
  \end{subfigure}\\[4pt]
  \begin{subfigure}[t]{0.48\linewidth}\centering
    \includegraphics[width=\linewidth]{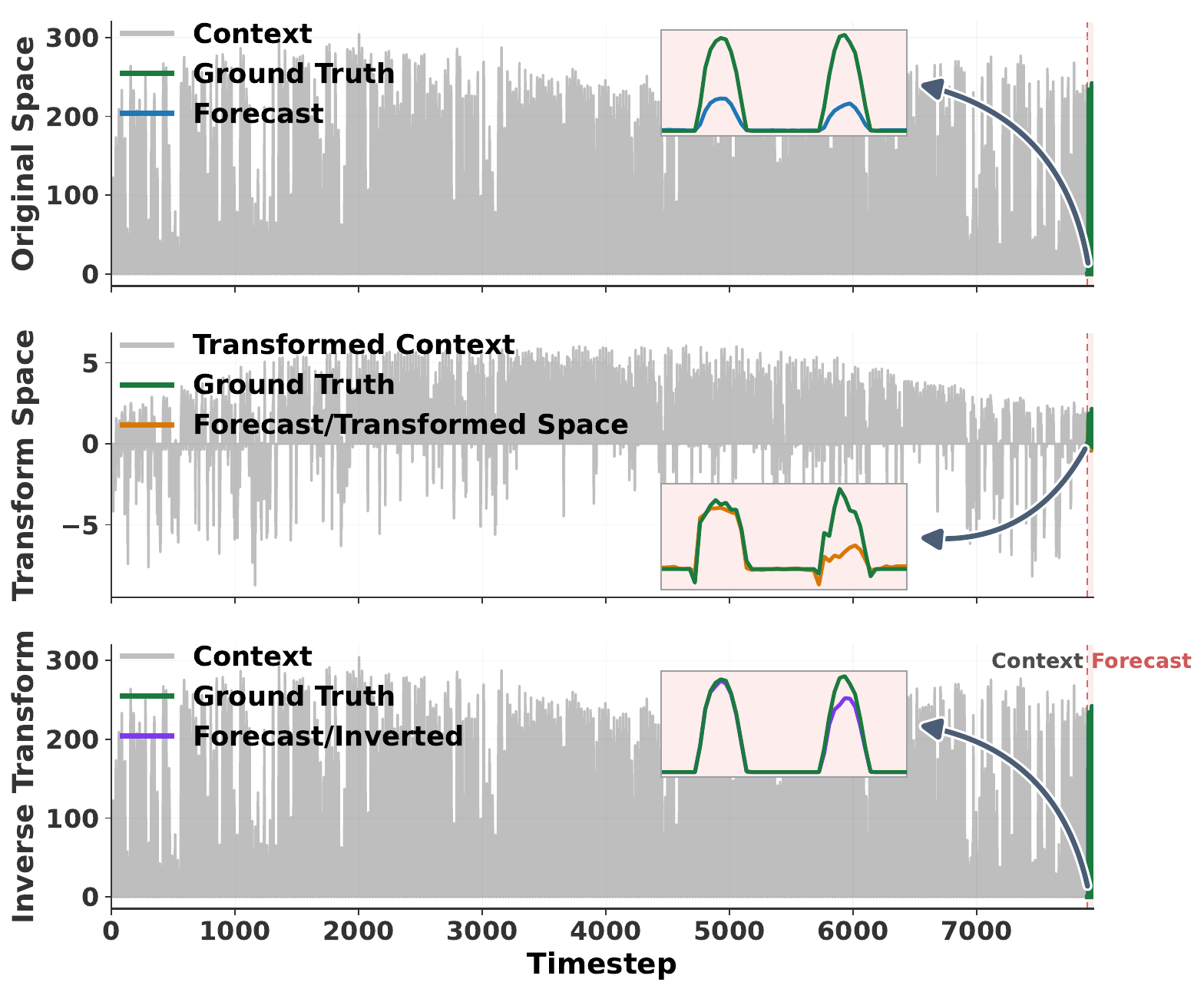}
    \caption{MASE $2.11 \rightarrow 0.32$, wQL $0.473 \rightarrow 0.104$, MAE $45.06 \rightarrow 6.75$.}
    \label{fig:perseries_solar_1977}
  \end{subfigure}\hfill
  \begin{subfigure}[t]{0.48\linewidth}\centering
    \includegraphics[width=\linewidth]{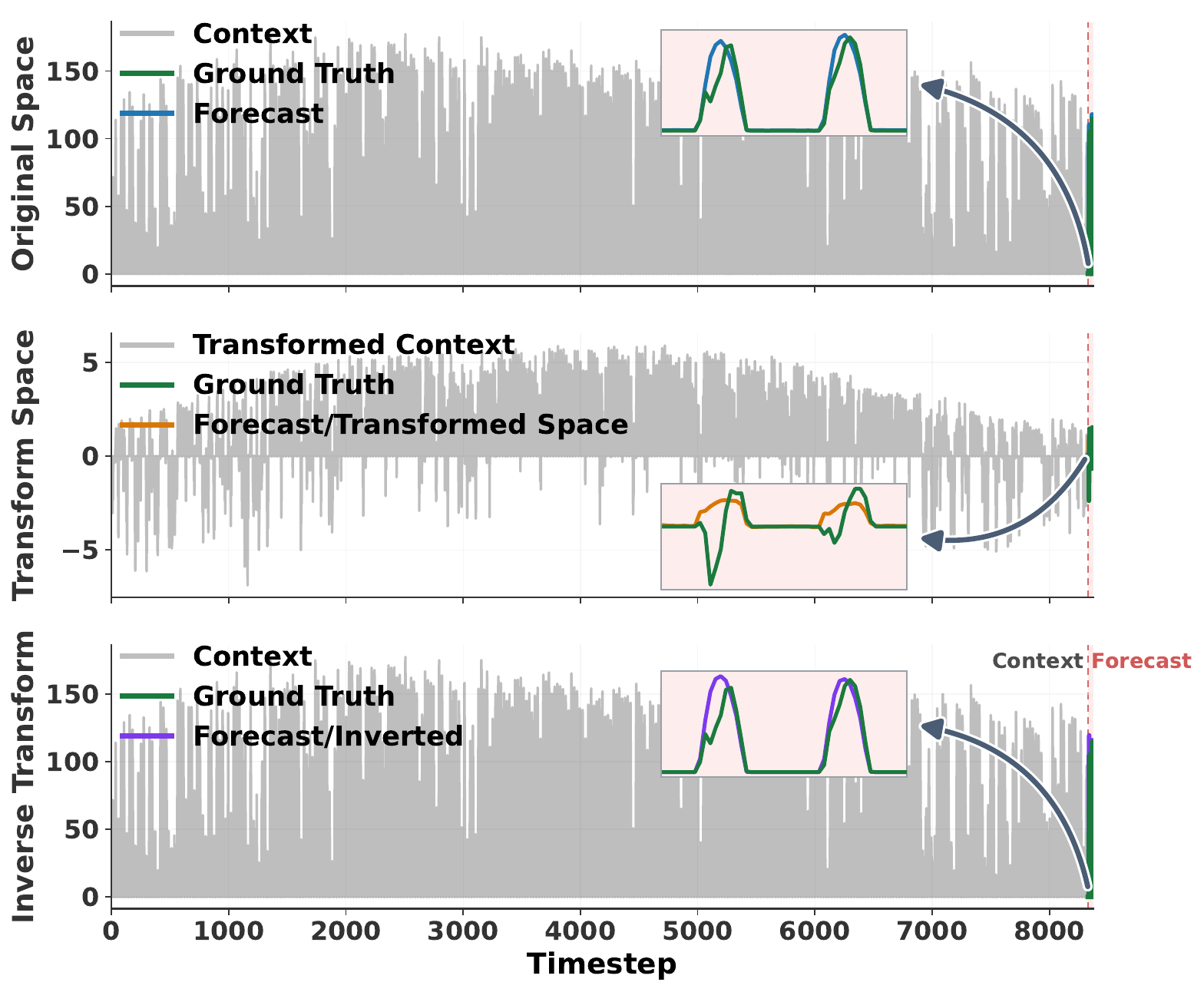}
    \caption{MASE $0.55 \rightarrow 0.60$, wQL $0.195 \rightarrow 0.219$, MAE $6.96 \rightarrow 7.53$.}
    \label{fig:perseries_solar_200}
  \end{subfigure}
  \caption{\small{Forecast trajectories on four $\mathtt{solar\_H}$ test instances. Hourly solar irradiance is strongly diurnal; on panels (a)-(c) the transform folds the daily cycle into a smoother oscillation that Chronos-2 extrapolates more cleanly. Panel (d) is included as a series where the transform leaves Chronos-2 slightly worse on every metric. The history line is rendered with reduced opacity so the forecast traces and the forecast-horizon inset stay legible against the long, densely diurnal context.}}
  \label{fig:perseries_solar_2x2}
\end{figure}

The hourly solar series are dominated by their diurnal cycle. In the transformed space as seen in Figure \ref{fig:perseries_solar_2x2} the daily oscillation is regularized into a smoother periodic shape, and Chronos-2 follows the periodic continuation more closely than it does on the raw window; the inverse-transformed forecast then sits within a tight band around the held-out trajectory. Fig.~\ref{fig:perseries_solar_200} shows the opposite case: a short, near-flat segment where the transform shifts Chronos-2 off the true level by a small amount and every metric regresses. Solar irradiance is largely driven by deterministic astronomy plus weather, so the transform's leverage here is geometric rather than informational, and the failure case on Fig.~\ref{fig:perseries_solar_200} is a small one.

\subsection{$\mathtt{restaurant}$}
\label{app:perseries_restaurant}

\begin{figure}[!t]
  \centering
  \begin{subfigure}[t]{0.48\linewidth}\centering
    \includegraphics[width=\linewidth]{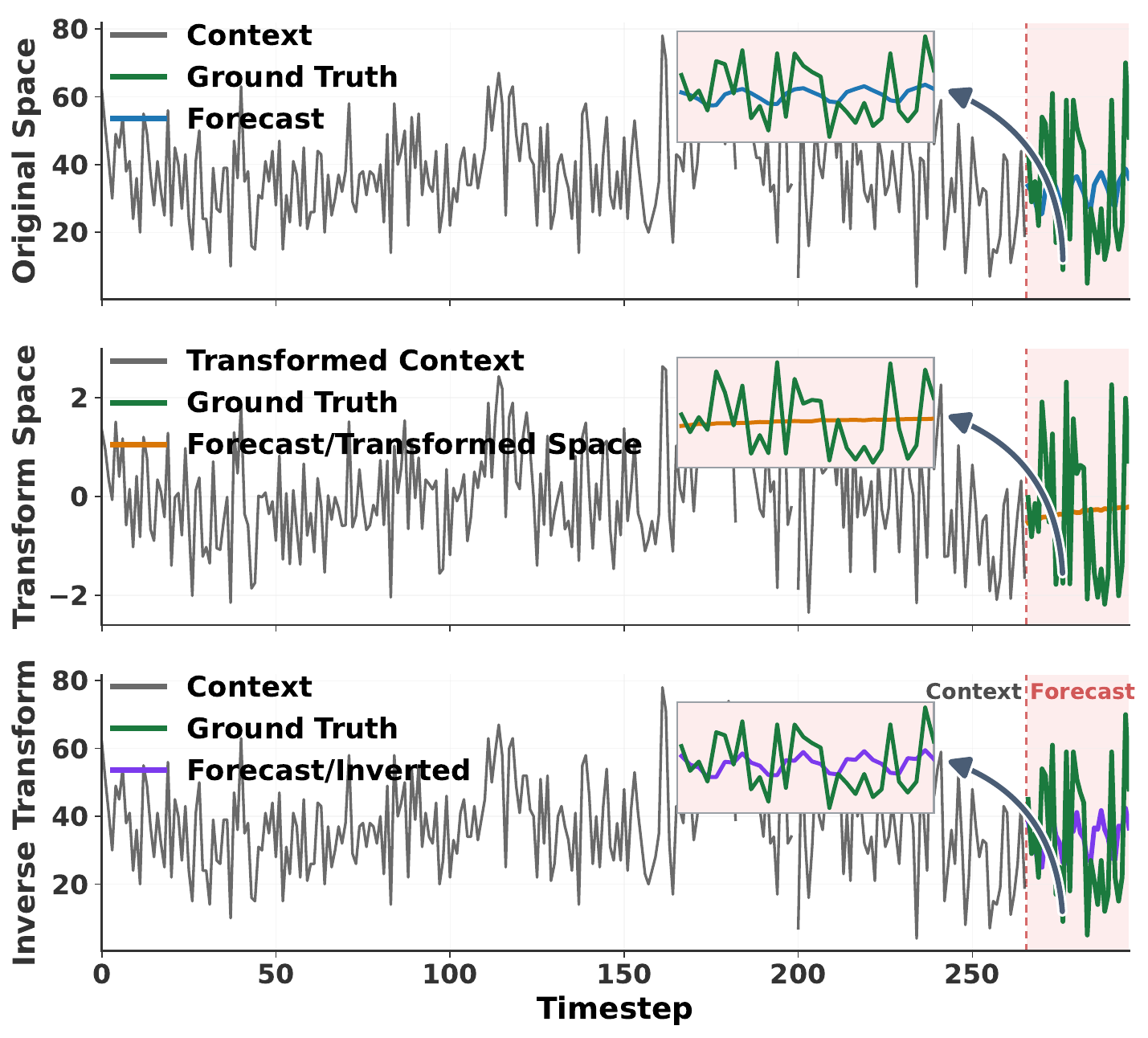}
    \caption{MASE $1.09 \rightarrow 1.08$, wQL $0.348 \rightarrow 0.346$, MAE $15.62 \rightarrow 15.40$.}
    \label{fig:perseries_restaurant_55}
  \end{subfigure}\hfill
  \begin{subfigure}[t]{0.48\linewidth}\centering
    \includegraphics[width=\linewidth]{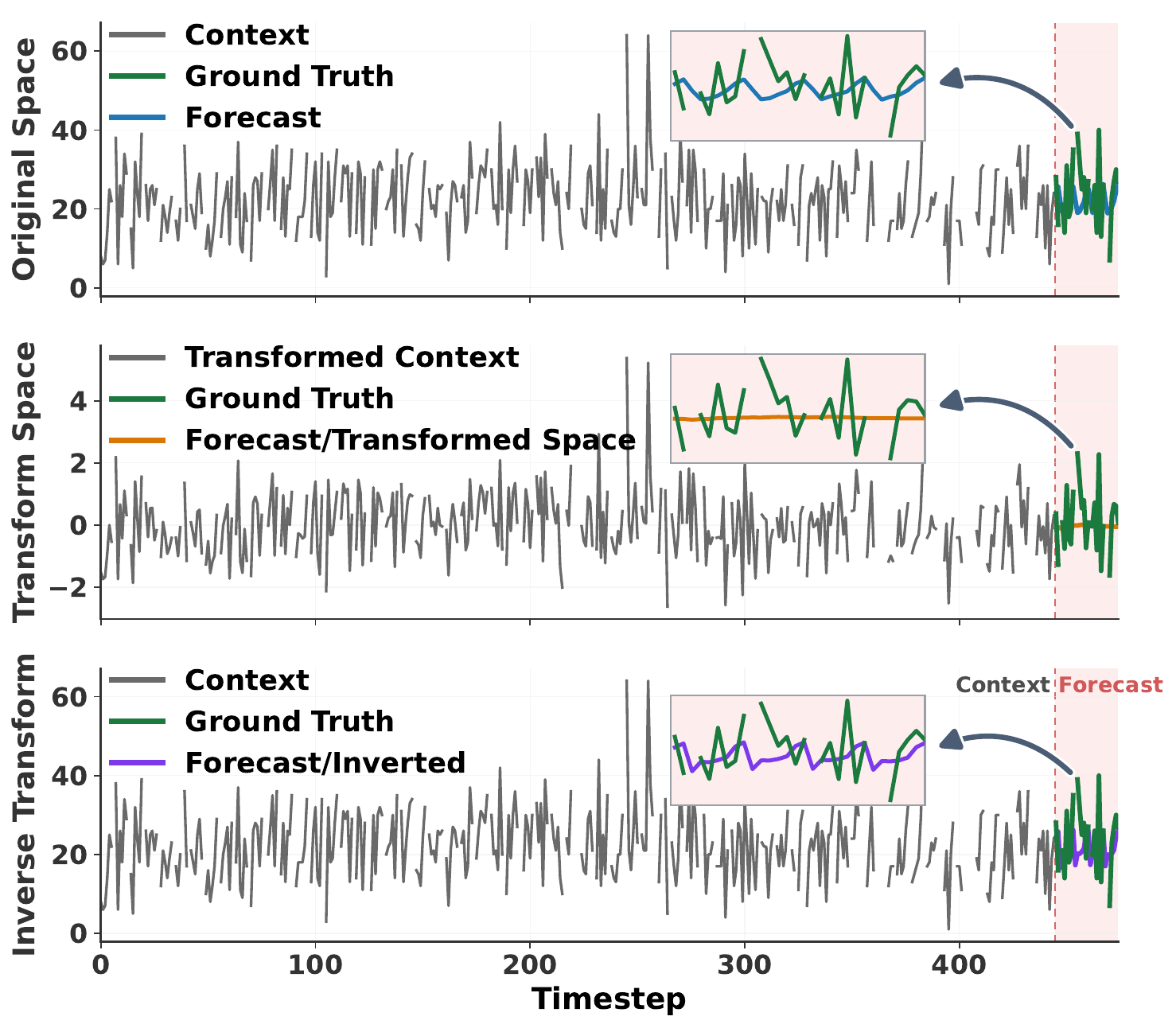}
    \caption{MASE $0.71 \rightarrow 0.70$, wQL $0.214 \rightarrow 0.214$, MAE $6.79 \rightarrow 6.68$.}
    \label{fig:perseries_restaurant_196}
  \end{subfigure}\\[4pt]
  \begin{subfigure}[t]{0.48\linewidth}\centering
    \includegraphics[width=\linewidth]{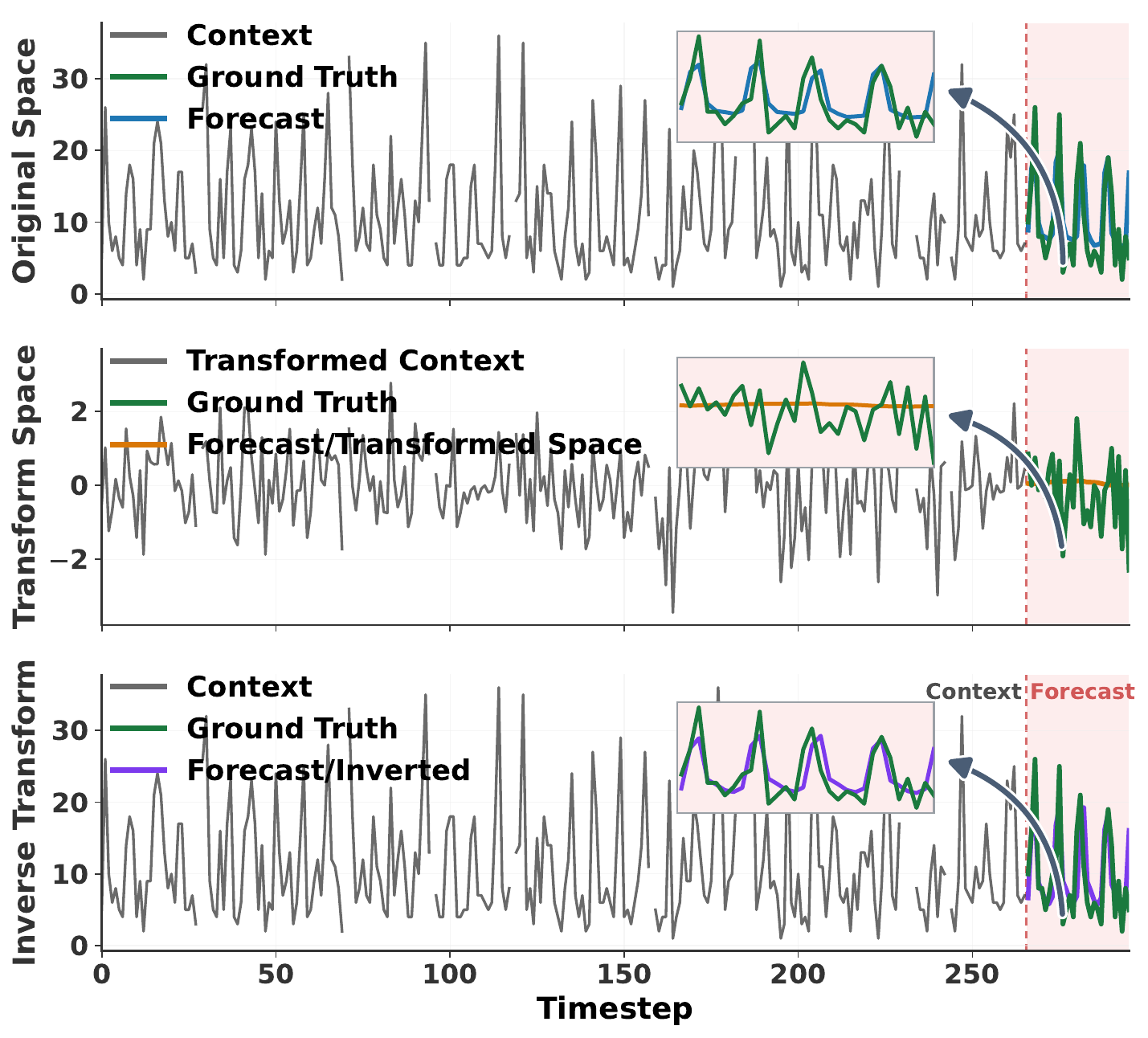}
    \caption{MASE $0.50 \rightarrow 0.49$, wQL $0.292 \rightarrow 0.280$, MAE $3.56 \rightarrow 3.43$.}
    \label{fig:perseries_restaurant_247}
  \end{subfigure}\hfill
  \begin{subfigure}[t]{0.48\linewidth}\centering
    \includegraphics[width=\linewidth]{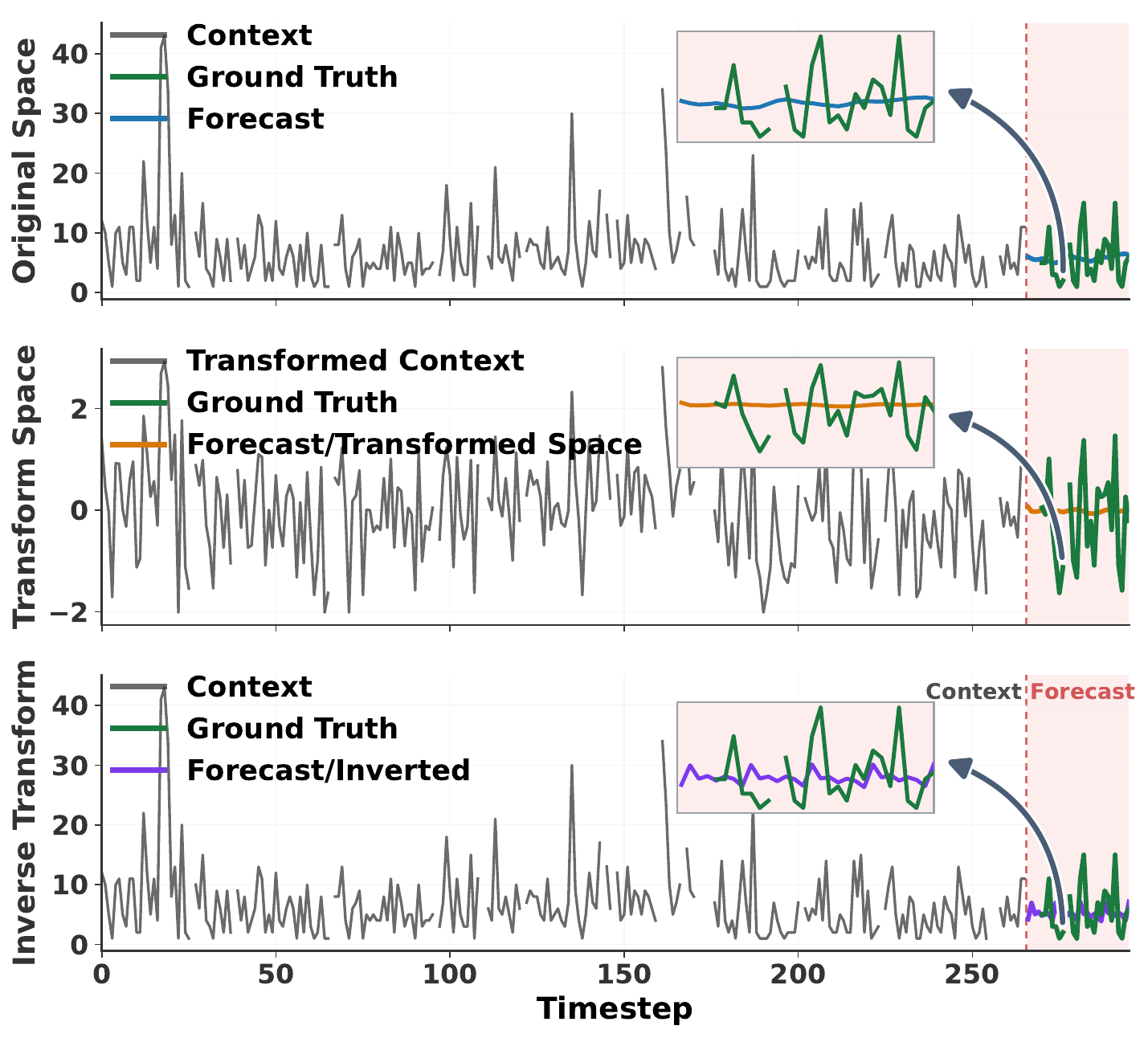}
    \caption{MASE $0.63 \rightarrow 0.61$, wQL $0.495 \rightarrow 0.464$, MAE $3.29 \rightarrow 3.17$.}
    \label{fig:perseries_restaurant_266}
  \end{subfigure}
  \caption{\small{Forecast trajectories on four $\mathtt{restaurant}$ test instances. The dataset-aggregate test MASE shifts by roughly $0.6\%$; the per-series traces match this near-identity regime, with the transformed window resembling the raw window and the inverse-transformed median essentially indistinguishable from the raw median.}}
  \label{fig:perseries_restaurant_2x2}
\end{figure}

The transformed input window resembles the raw input on each of these picks, and the top and middle panel forecasts overlap closely as seen in Figure \ref{fig:perseries_restaurant_2x2}. The inverse-transformed median is essentially indistinguishable from the raw median, and the per-series gains in MASE, wQL, and MAE are small. The corpus is built from daily visitor counts at Japanese restaurants drawn from two reservation platforms, and the official held-out window spans the Golden Week holiday; days on which a restaurant was closed are excluded from scoring. The drivers of variation are therefore primarily exogenous: the holiday calendar, day-of-week effects modulated by closure decisions, and platform-level demand. None of these are encoded in the input history at the level of detail needed to anticipate the held-out trajectory, and an input-only transform cannot manufacture the missing signal. The aggregate near-identity behaviour reported in the main paper is consistent with this: when the relevant covariates live outside the history window, the optimizer settles close to the identity rather than overfit a spurious mapping.

\subsection{$\mathtt{bitbrains\_rnd/H}$}
\label{app:perseries_bitbrains}

\begin{figure}[!t]
  \centering
  \begin{subfigure}[t]{0.48\linewidth}\centering
    \includegraphics[width=\linewidth]{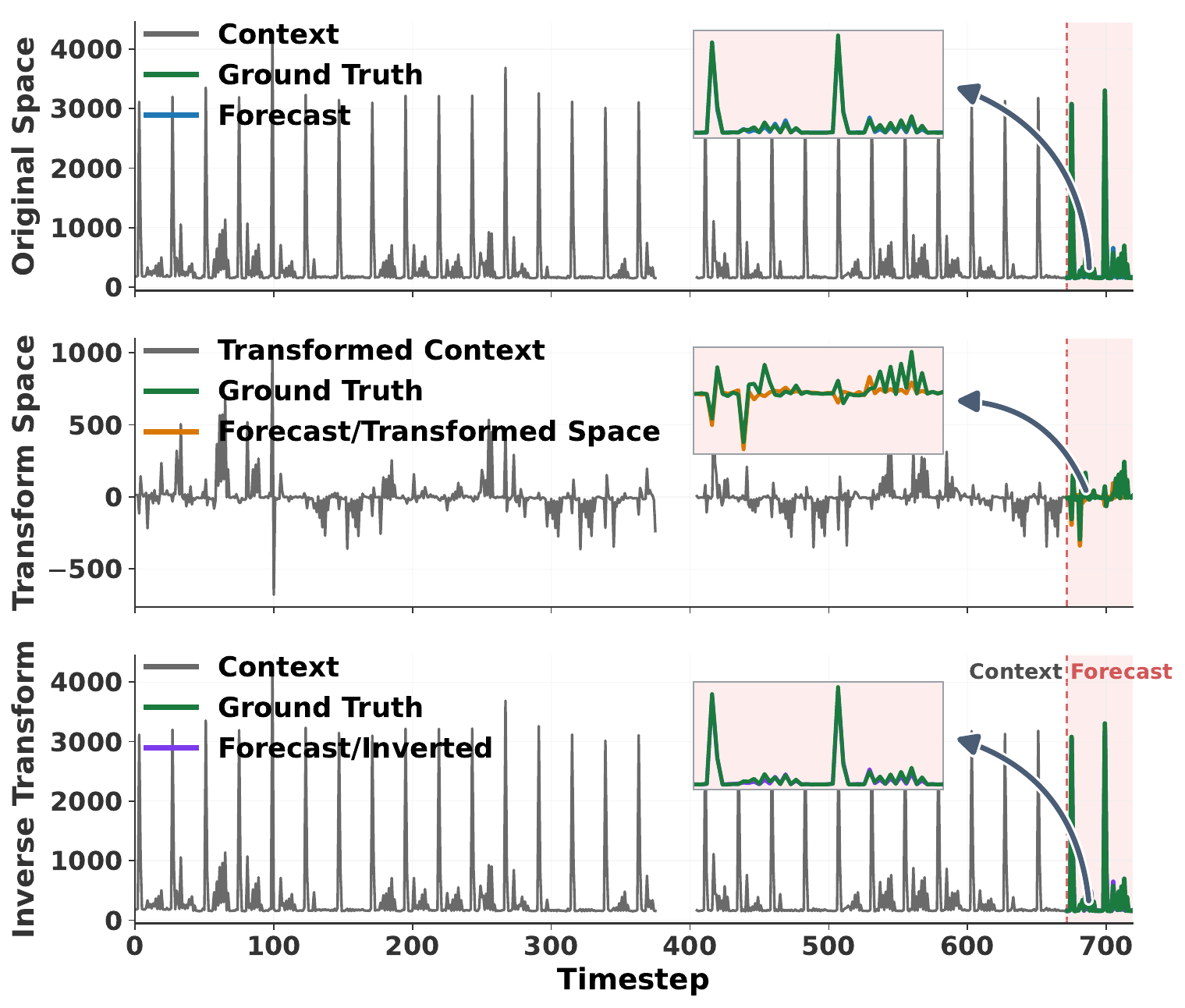}
    \caption{MASE $0.63 \rightarrow 0.55$, wQL $0.080 \rightarrow 0.076$, MAE $41.94 \rightarrow 36.91$.}
    \label{fig:perseries_bitbrains_77}
  \end{subfigure}\hfill
  \begin{subfigure}[t]{0.48\linewidth}\centering
    \includegraphics[width=\linewidth]{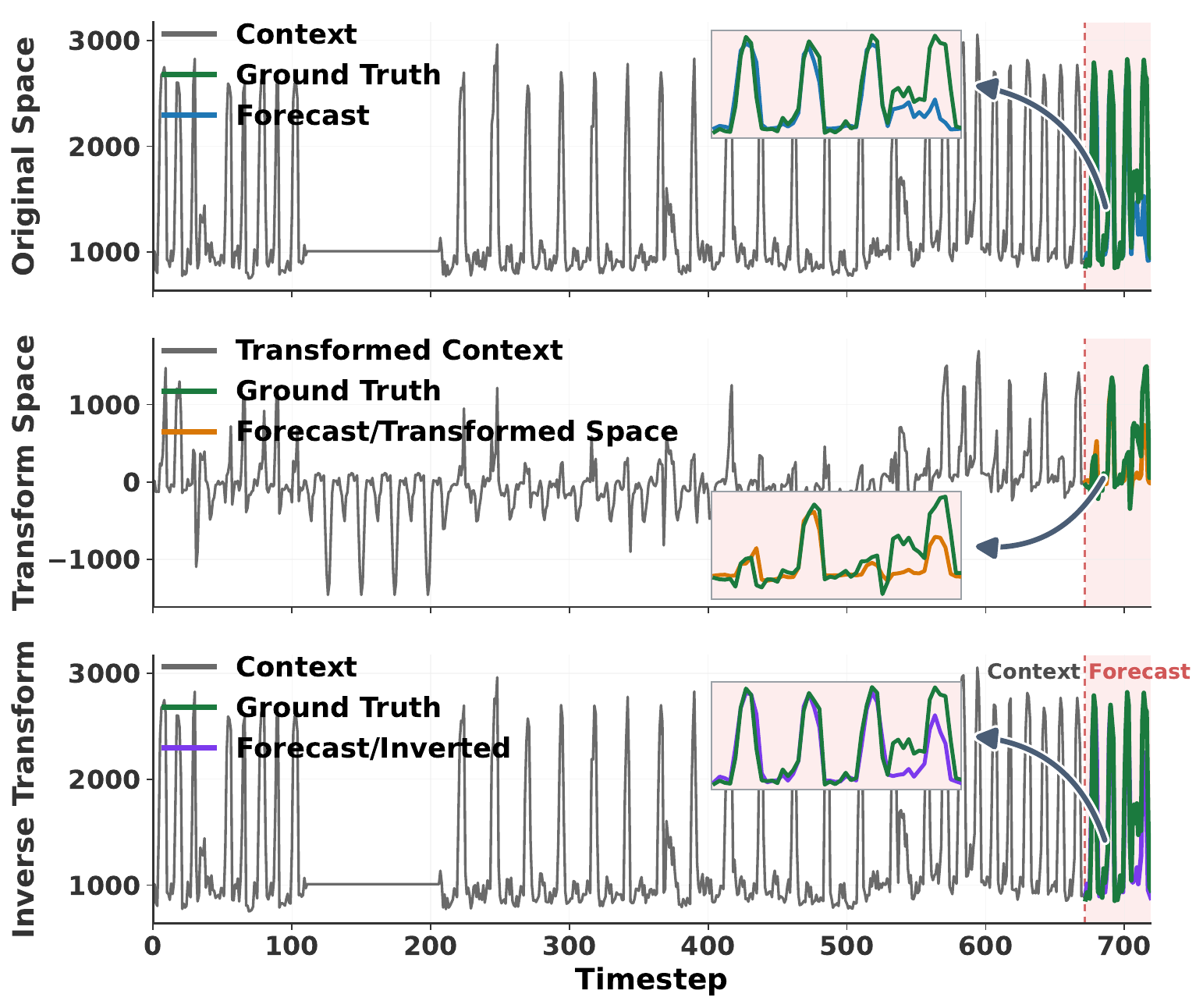}
    \caption{MASE $1.37 \rightarrow 1.18$, wQL $0.122 \rightarrow 0.111$, MAE $269.2 \rightarrow 232.0$.}
    \label{fig:perseries_bitbrains_215}
  \end{subfigure}\\[4pt]
  \begin{subfigure}[t]{0.48\linewidth}\centering
    \includegraphics[width=\linewidth]{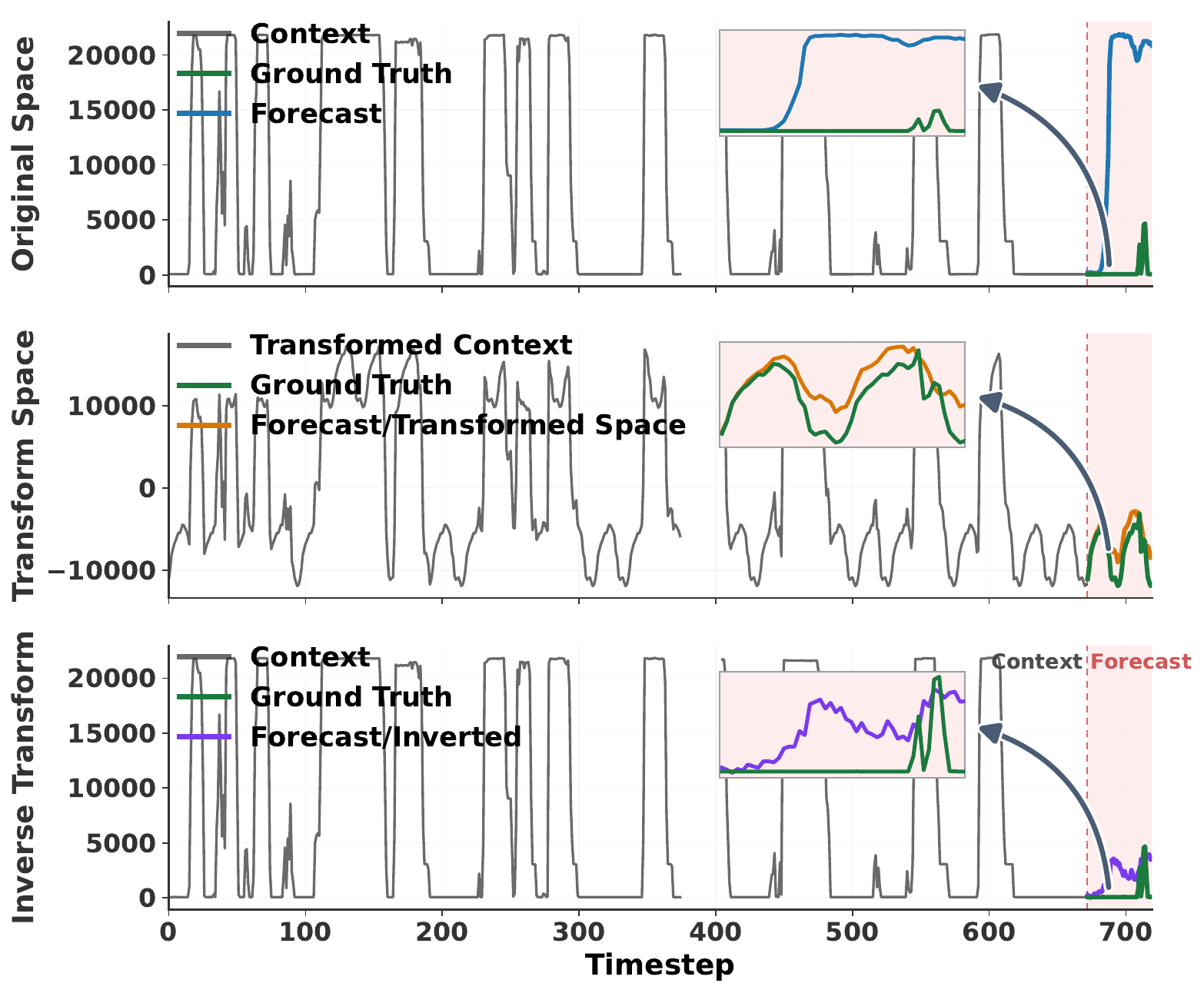}
    \caption{MASE $1.47 \rightarrow 0.18$, wQL $24.23 \rightarrow 6.59$, MAE $14296 \rightarrow 1758$.}
    \label{fig:perseries_bitbrains_267}
  \end{subfigure}\hfill
  \begin{subfigure}[t]{0.48\linewidth}\centering
    \includegraphics[width=\linewidth]{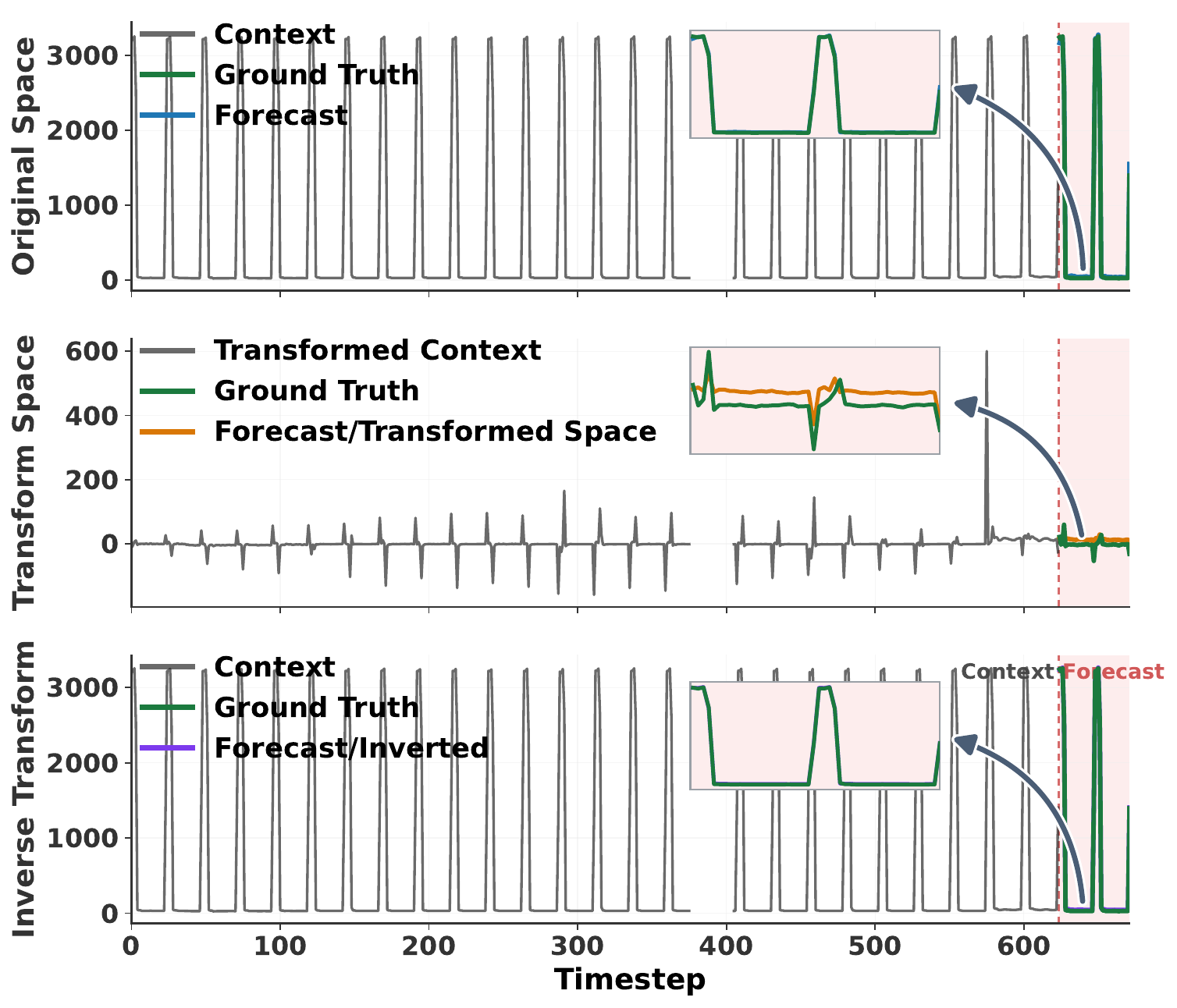}
    \caption{MASE $3.13 \rightarrow 2.00$, wQL $0.031 \rightarrow 0.023$, MAE $24.58 \rightarrow 15.70$.}
    \label{fig:perseries_bitbrains_458}
  \end{subfigure}
  \caption{\small{Forecast trajectories on four $\mathtt{bitbrains\_rnd/H}$ test instances; all four are channel $0$ of the bivariate input. The dataset-aggregate test MASE shifts by about $0.1\%$, but the per-series effect varies sharply: panel (c) shows an order-of-magnitude reduction in MASE on a single series, while panels (a), (b), and (d) show only modest gains and the transformed window retains the spiky character of the raw input.}}
  \label{fig:perseries_bitbrains_2x2}
\end{figure}

The corpus is a workload trace from a virtualized datacenter hosting business-critical applications. The channels record requested and used resource demand on a heterogeneous shared cluster. Demand on such a cluster is driven mostly by external user activity and operational events (deployments, batch windows, traffic incidents), which makes the dynamics highly non-Markovian with respect to the channel-only input window we feed Chronos-2. The per-series picture is correspondingly heavy-tailed: Fig.~\ref{fig:perseries_bitbrains_267} is a series where the transform compresses a high-magnitude spike into a smoother envelope and the inverse-transformed forecast tracks the held-out trajectory inside that envelope, dropping per-series MASE by roughly an order of magnitude. The other three picks retain the raw window's spiky character through the transform, and the inverse-transformed median sits close to the raw median. The aggregate near-identity behaviour reported in the main paper is therefore a mean over a heavy-tailed distribution of per-series effects, not a uniform absence of effect, and we do not extend the periodic transform-space observation from $\mathtt{covid\_deaths}$ and $\mathtt{solar\_H}$ to this dataset.

\clearpage

\clearpage

\section{Evolution-tree visualization of the search process}
\label{app:efetime_evotree}

This section visualizes the OpenEvolve search itself across the EFE-Time evaluation suite. Each evolution run is rendered as a single panel; the three random seeds for a given dataset are laid out horizontally so that within-dataset variability is visible at a glance. The intent is qualitative.

Every run is initialized from an identity-transform seed program at iteration $0$ and proceeds for $100$ iterations. At each iteration, the controller samples a parent program from the in-memory MAP-Elites database (three islands with periodic migration), prompts an LLM with the parent code together with a small set of inspiration and diversity exemplars, and asks for a complete rewrite. The new candidate is evaluated on the per-dataset evolution pool by running Chronos-2 zero-shot on the transformed history, inverse-transforming the forecast, and scoring the result against the identity baseline.

In every panel, each program is plotted at its $(\text{combined score}, \text{iteration})$ coordinates, coloured by the island the program was placed in. Faint grey segments are parent-to-child edges read from the per-program $\mathtt{parent\_id}$ pointer in the run's saved checkpoints. The violet trace highlights the lineage of the run's final selection from root to leaf, and the violet star marks that program. The dashed grey line is the running best-so-far. The y-axis grows downward so the seed program sits at the top of every panel. A few panels show the violet trace truncated, or only the final star with no incoming line. This reflects the bounded MAP-Elites store rather than the search itself: as new candidates arrive, weaker programs are evicted from their feature cells, and a saved checkpoint persists only the active population at the moment of saving. A program that lives transiently in the database and then evicts before the next checkpoint write is never serialized to disk, so the parent JSON needed to draw an edge is missing. The lineage line therefore stops at the last ancestor that remained in the population.

\begin{figure}[!t]
  \centering
  \includegraphics[width=\linewidth]{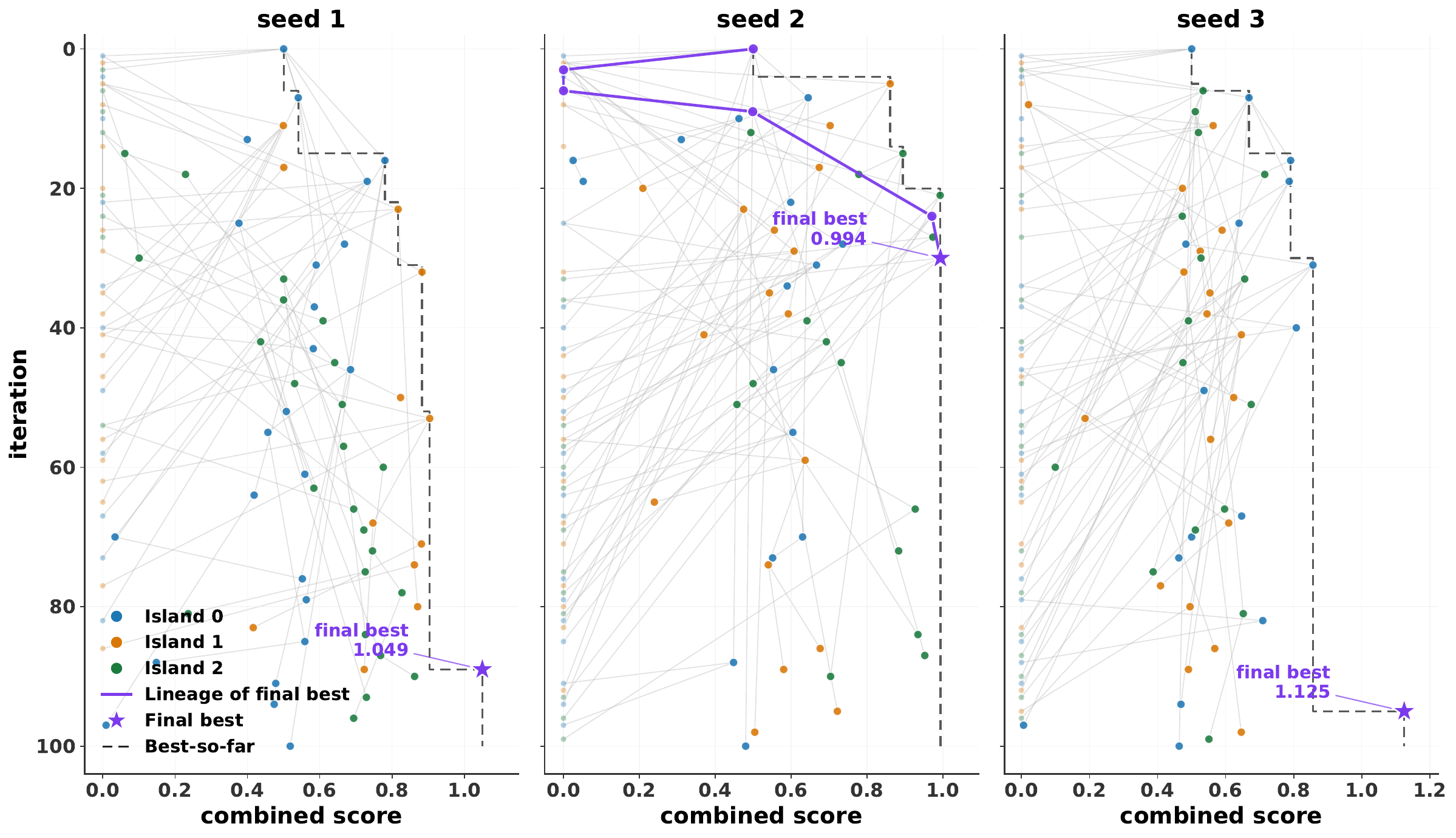}
  \caption{\small{Evolution trees for the three $\mathtt{covid\_deaths}$ runs (seeds $1$, $2$, $3$ from left to right). All three seeds locate programs that cross the $10\%$-improvement line.}}
  \label{fig:evotree_covid}
\end{figure}

\begin{figure}[!t]
  \centering
  \includegraphics[width=\linewidth]{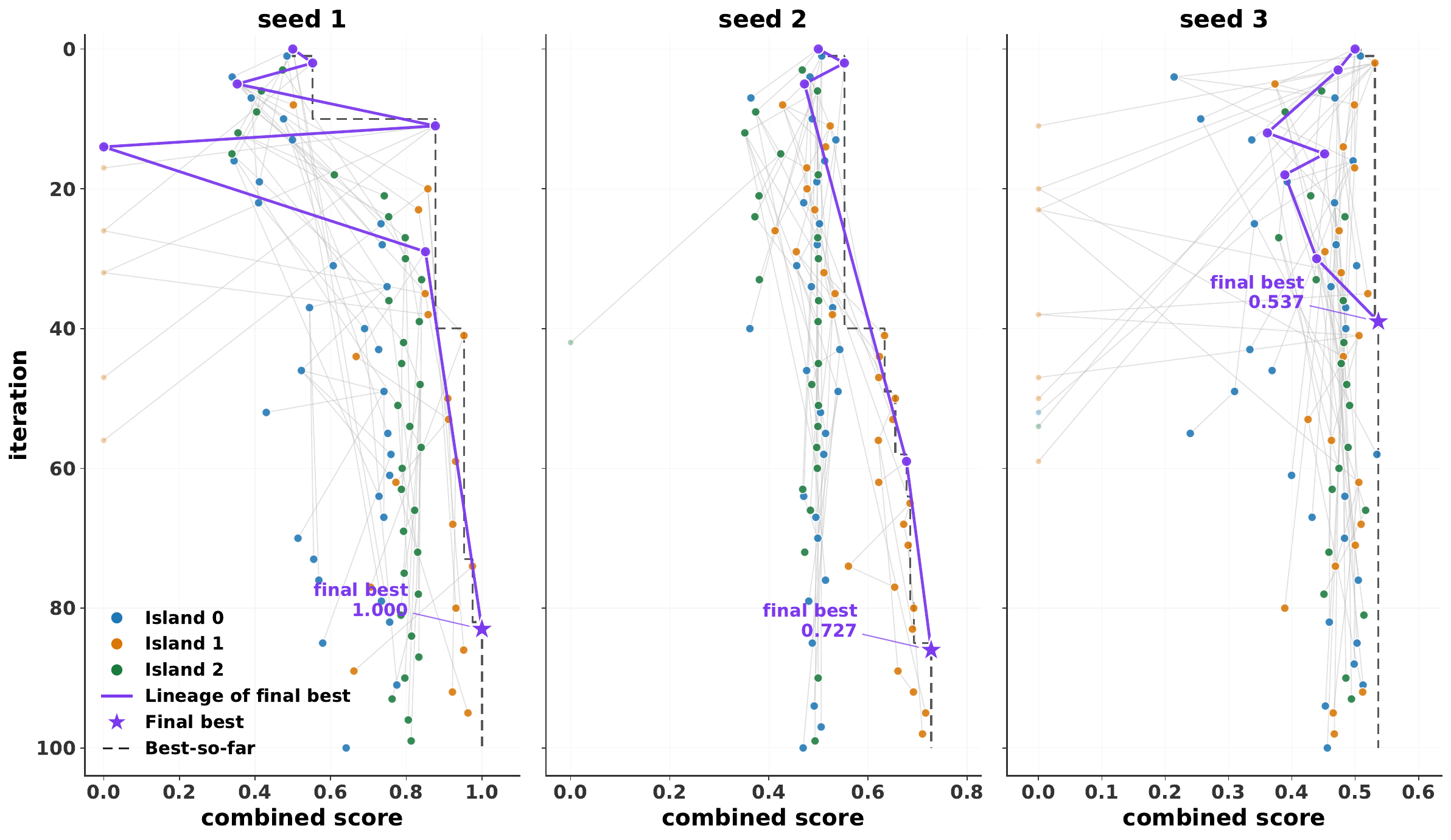}
  \caption{\small{Evolution trees for the three $\mathtt{m4\_yearly}$ runs. The seed-to-seed score variability is visibly wider than on $\mathtt{covid\_deaths}$.}}
  \label{fig:evotree_m4}
\end{figure}

\begin{figure}[!t]
  \centering
  \includegraphics[width=\linewidth]{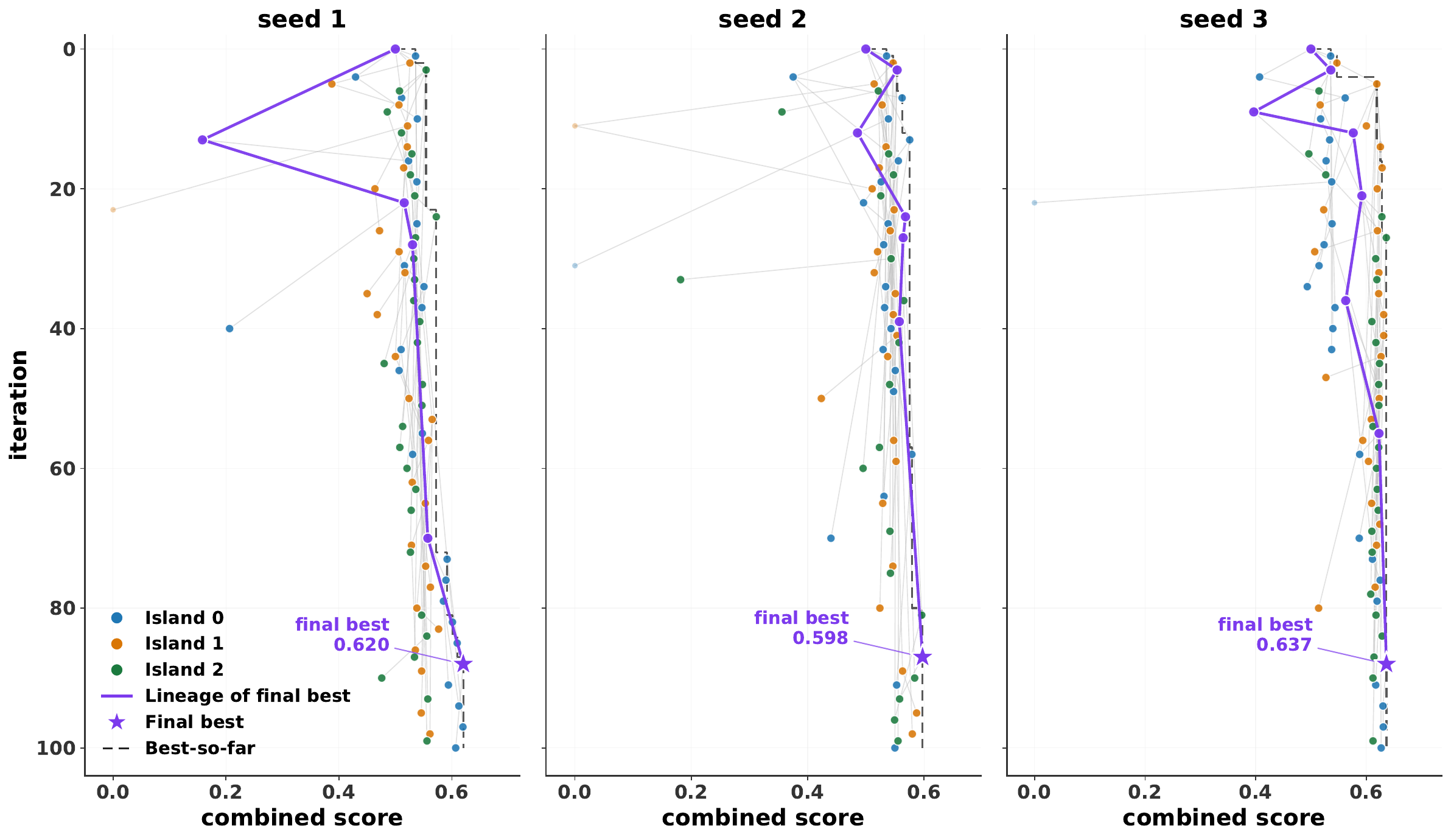}
  \caption{\small{Evolution trees for the three $\mathtt{solar\_H}$ runs. The three panels are visually similar.}}
  \label{fig:evotree_solar}
\end{figure}

\begin{figure}[!t]
  \centering
  \includegraphics[width=\linewidth]{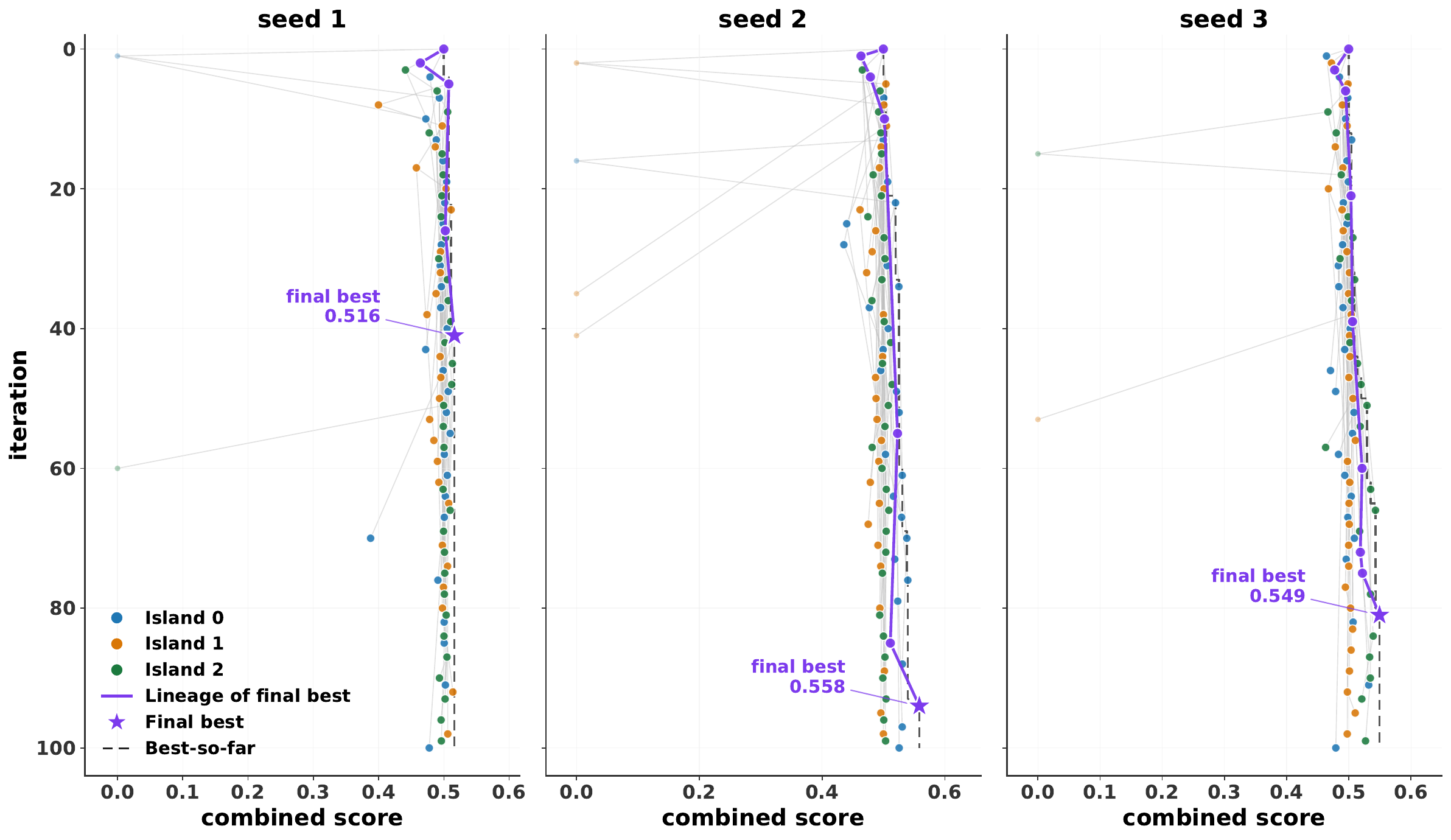}
  \caption{\small{Evolution trees for the three $\mathtt{restaurant}$ runs. The optimizer-selected scores sit only marginally above the identity baseline.}}
  \label{fig:evotree_restaurant}
\end{figure}

\begin{figure}[!t]
  \centering
  \includegraphics[width=\linewidth]{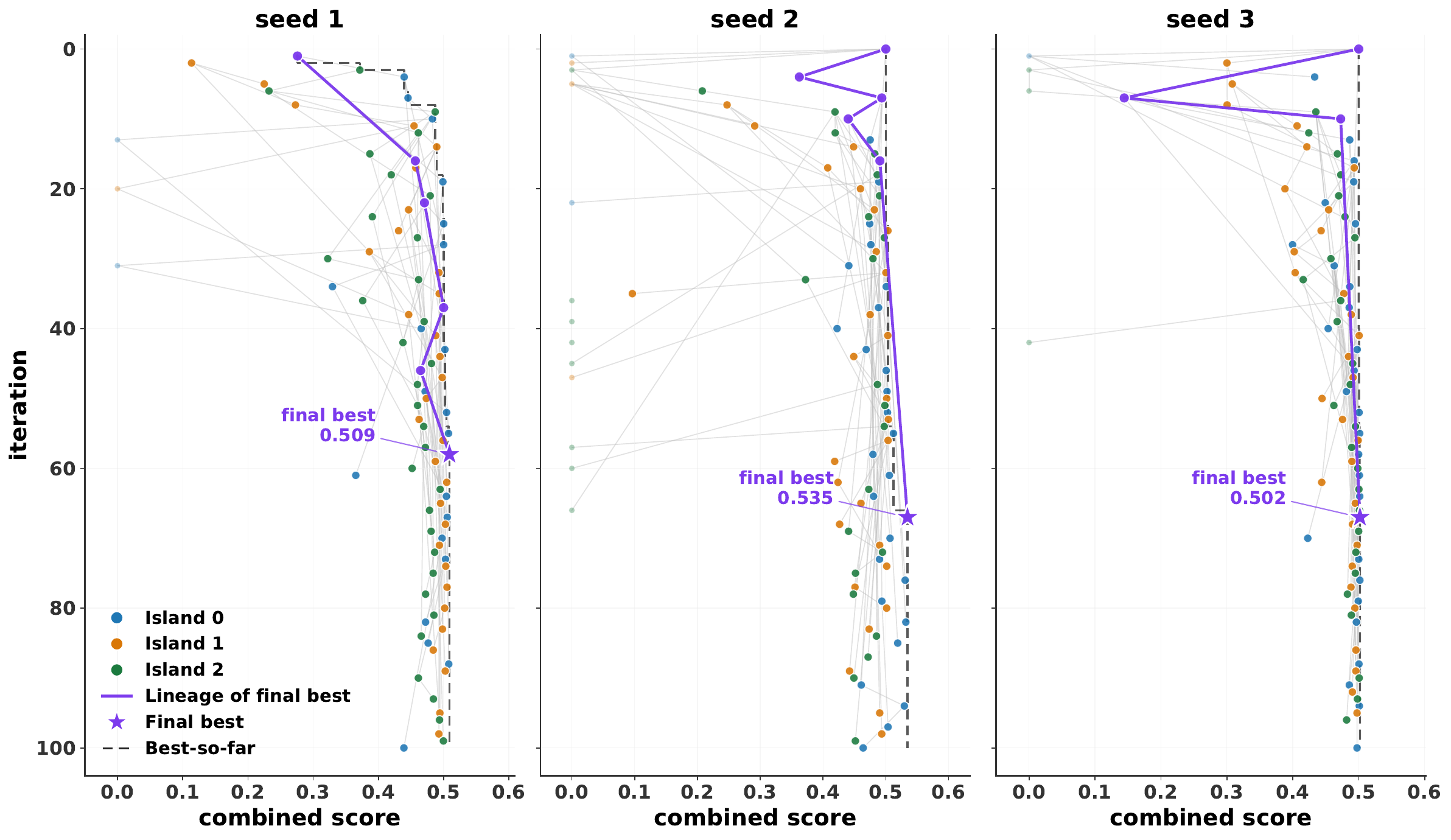}
  \caption{\small{Evolution trees for the three $\mathtt{bitbrains\_rnd/H}$ runs. As on $\mathtt{restaurant}$, the populations crowd a vertical band around the identity baseline.}}
  \label{fig:evotree_bitbrains}
\end{figure}

The five panels reproduce, at the search-trajectory level, the qualitative pattern visible at the per-series level. On $\mathtt{covid\_deaths}$ (Fig.~\ref{fig:evotree_covid}) all three seeds locate strongly-improving programs that cross the $10\%$-improvement line; the seed-$2$ panel shows a long-lived ancestor whose lineage walks back to the seed, whereas the other two panels render only the final star because the winners' direct parents were evicted before the next checkpoint write. On $\mathtt{m4\_yearly}$ (Fig.~\ref{fig:evotree_m4}) the seed-to-seed variance is visibly wider, and only one seed crosses the $10\%$-improvement line; the slowest run flattens its best-so-far frontier in the first half of the budget and does not escape the corresponding plateau, which we read as the search getting stuck in a basin reachable by the LLM's rewrites rather than as evidence that no better program exists. On $\mathtt{solar\_H}$ (Fig.~\ref{fig:evotree_solar}) the three runs converge to similar combined scores in a narrow band; the populations crowd a tight vertical strip and the best-so-far frontier moves in small increments, consistent with the per-series picture from Sec.~\ref{app:perseries_solar_H}, where the diurnal regularity of hourly solar irradiance gives Chronos-2 a strong baseline that bounds the headroom for any single transform. On $\mathtt{restaurant}$ (Fig.~\ref{fig:evotree_restaurant}) and $\mathtt{bitbrains\_rnd/H}$ (Fig.~\ref{fig:evotree_bitbrains}) the optimizer-selected scores sit just above the identity baseline; the populations crowd a near-vertical band around the identity score, the best-so-far frontiers inch rather than jump, and on $\mathtt{restaurant}$ one seed locks in its final selection in the early iterations and does not move thereafter. Read together with Sec.~\ref{app:perseries_restaurant} and Sec.~\ref{app:perseries_bitbrains}, this matches a search that does not find a transform with a meaningful effect on these corpora: the population drifts around identity across all three seeds, ruling out single-run failure as the explanation.

\clearpage

\clearpage

\section{Prompt example and reasoning traces}
\label{app:efetime_prompt_example}

This section demonstrates a working example of EFE-Time on $\mathtt{covid\_deaths}$. We also provide the EFE-Tab system prompt and its example final program on  $\mathtt{churn}$ and $\mathtt{in\_vehicle\_coupon\_recommendation}$.

\subsection{Time-series example: $\mathtt{covid\_deaths}$}
\label{app:prompt_example_covid}

We pick run~$2$ of $\mathtt{covid\_deaths}$ and feature the step from iteration~$6$ to iteration~$9$ in its lineage (middle panel of Fig.~\ref{fig:evotree_covid}, Sec.~\ref{app:efetime_evotree}; Table~\ref{tab:prompt_example_covid_lineage}). The interest of this case is that the search recovers from two consecutive failed mutations (iterations~$3$ and~$6$, both with combined score $0$) and lands on a robust affine normalization that becomes the structural core of every later program in the lineage. We show the rendered prompt sent to the LLM, the LLM's reasoning, and the resulting program. 

\begin{table}[!ht]
  \centering
  \small
    \caption{\small{Six-node lineage of the displayed final program for $\mathtt{covid\_deaths}$ run~$2$. The iteration-$9$ candidate (the worked example in this section) takes the search from a failure regime back to identity-class performance, after which two further candidates bring the score above the $10\%$-improvement line.}}
  \begin{tabular}{rccl}
    \toprule
    Iter. & Combined score & Code lines & Description \\
    \midrule
    $0$  & $0.5002$ & $85$  & identity seed \\
    $3$  & $0.0000$ & $88$  & first mutation, high harm rate \\
    $6$  & $0.0000$ & $79$  & second failure \\
    $9$  & $0.4988$ & $86$  & robust affine normalization (worked example below) \\
    $24$ & $0.9711$ & $108$ & branching and winsorisation added \\
    $30$ & $0.9936$ & $108$ & final displayed program \\
    \bottomrule
  \end{tabular}
  \label{tab:prompt_example_covid_lineage}
\end{table}

The system prompt establishes the transform-program API, lists transform families, and is held constant across iterations of a run.

The user prompt carries the per-iteration content: a JSON evaluation summary of the parent program, the dataset context for $\mathtt{covid\_deaths}$, and the evolution history of previously evaluated programs. To keep the section readable, embedded Python code blocks for previously evaluated programs are elided in place; each elision marker reports the line count of the elided body. Section headers, the evaluation-summary JSON, the dataset-context block, and the response template at the end are kept verbatim.

The LLM's response opens with a diagnostic paragraph identifying why the parent failed, states a strategy, and emits the candidate program. We treat the prose preamble as the model's reasoning trace and present the assistant message as a single transcript.

The takeaway from this step is structural: the recent-window affine normalization introduced at iteration~$9$ is the ancestor of every later program in the lineage. Subsequent iterations refine it (branching for short and constant series, a winsorisation step, a numerical guard) without revisiting the core. For completeness, we list the displayed final program at iteration~$30$ also below.

\clearpage

\begin{promptbox}[System prompt for EFE-Time]
\begin{Verbatim}
You are evolving a target transformation program for time series forecasting.

The program defines a TransformProgram class with three methods:
- fit(y_hist, meta) -> state dict of fitted parameters
- transform(y, state) -> transformed values
- inverse_transform(z, state) -> original-scale values

The downstream forecasting model (Chronos-2) is fixed.
Your job is to find a transform that makes the series easier to forecast.

HOW IT WORKS:
At each forecast origin, the evaluator:
1. Passes the historical target window to fit() along with metadata
2. Transforms the history with transform()
3. Feeds the transformed history to Chronos-2
4. Inverse-transforms the forecast back to original scale
5. Scores MASE against the identity-transform baseline

META DICT AVAILABLE IN fit():
The meta parameter is a dict with these keys:
- prediction_length: forecast horizon (int)
- seasonal_period: e.g. 12 for monthly (int)
- length: total history length
- n_valid: non-NaN count
- mean, std, median: basic location/scale stats
- min_val, max_val, range: value range
- positive_frac: fraction of values > 0
- zero_frac: fraction of values == 0
- skewness: skew of the series
- cv: coefficient of variation (std/|mean|)
- trend_strength: R^2 of linear fit (0 = no trend, 1 = perfect trend)
- recent_mean, recent_std: stats of the last seasonal_period observations

Use these to decide WHICH transform family to use and HOW to parameterize it.
For example:
- High positive_frac + high skewness -> shifted log1p
- Mixed positive/negative + high skewness -> asinh
- Low skewness + low cv -> identity might be best
- Large range + high cv -> scaled transform

TRANSFORM FAMILIES TO EXPLORE:
- identity: no transform (the baseline you must beat)
- shifted log1p: shift = -min + eps, z = log1p(y + shift)
- signed log: z = sign(y) * log1p(|y|)
- asinh: z = asinh(y / scale), with scale fitted from data
- scaled asinh: z = asinh((y - center) / scale)
- bounded power: Box-Cox-like with clamped lambda
- piecewise: different transforms for different value ranges
- affine normalization: (y - center) / scale, e.g. mean/std, median/IQR,
  median/MAD -- robust variants often help when outliers are present
- recent-window normalization: use stats from the last seasonal_period
  observations instead of the full history (useful for non-stationary series)
- detrending: z[i] = y[i] - (slope * i + intercept), fit slope/intercept
  via linear regression. Inverse for the FORECAST must use
  state["hist_len"] + np.arange(len(z)) for the time index, NOT
  np.arange(len(z)) -- see the CRITICAL hist_len section below.
- seasonal phase centering: subtract the mean of each phase i % seasonal_period
  from y. Inverse forecast must use (state["hist_len"] + np.arange(len(z))) % s
  for the phase index, NOT np.arange(len(z)) % s.
- composites: combine any of the above (e.g. detrend -> asinh, or seasonal
  centering -> affine). Each step must be exactly invertible in reverse order.

WHAT TO VARY:
- Family choice based on meta dict
- Shift formula (min-based, percentile-based, etc.)
- Scale formula (std-based, IQR-based, MAD-based)
- Power parameter bounds
- Recent-window vs full-window statistics
- Branching logic (if/else based on meta values)

CRITICAL -- hist_len AND THE FORECAST WORKFLOW:
fit() MUST store "hist_len": len(y_hist) in the returned state dict.
This is critical because transform() and inverse_transform() are called
on DIFFERENT-LENGTH arrays:
- transform() receives the full history (length T)
- inverse_transform() receives the forecast (length prediction_length)
For ANY position-dependent operation (detrending, time-varying scaling),
you MUST use state["hist_len"] to compute the correct time offset for
the forecast positions, NOT len(z). Example for detrending:
  transform:  z[i] = y[i] - slope * i
  inverse:    y[i] = z[i] + slope * (state["hist_len"] + i)
If you use np.arange(len(z)) in inverse_transform, it will produce
indices 0..11 instead of the correct 72..83, DESTROYING the forecast.

RULES -- YOUR PROGRAM MUST:
- fit parameters using ONLY the history passed to fit()
- ALWAYS include "hist_len": len(y_hist) in the state dict
- NOT use future data
- NOT mutate input arrays (copy first)
- Preserve length, order, and NaN positions
- Return finite values (no Inf); NaN only where input had NaN
- Provide an exact inverse: |inverse(transform(y)) - y| < 1e-4
- Keep inverse stable slightly beyond observed transformed range
- inverse_transform MUST work on arrays of any length (not just hist_len)
- Be deterministic given the same input
- Return a dict from fit() (the state)

COMMON MISTAKES TO AVOID:
- Using np.arange(len(z)) in inverse_transform for position-dependent ops
  (MUST use state["hist_len"] + np.arange(len(z)) for forecast positions)
- Dividing by zero (always add eps to denominators)
- log of negative numbers (check and shift first)
- Inverse that doesn't match forward (verify the math)
- Using np.log instead of np.log1p (log(0) = -inf)
- Forgetting to copy arrays before modifying
- State values that are NaN or Inf

FEEDBACK YOU WILL RECEIVE:
The evaluator provides an "evaluation_summary" artifact with:
- candidate_mase / baseline_mase: your MASE vs identity baseline
- mase_ratio: < 1.0 means you're better than baseline
- help_rate: fraction of series where you're better
Use this to understand WHERE your transform helps and hurts.

STRATEGY:
1. Start by understanding the meta dict values
2. Choose an appropriate transform family
3. Fit parameters carefully from the history using meta dict statistics
4. Verify your inverse math is correct
5. Add branching if a single transform family cannot cover all series
\end{Verbatim}
\end{promptbox}

\begin{promptbox}[Openevolve user prompt for Covid-Deaths (parent code blocks omitted)]
\begin{Verbatim}
# Current Program Information
- Fitness: 0.0000
- Feature coordinates: 
- Focus areas: - Fitness unchanged at 0.0000
- No feature coordinates
- Consider simplifying - code length exceeds 500 characters

## Last Execution Output

### evaluation_summary
```
{
  "dataset": "covid_deaths",
  "prediction_length": 30,
  "seasonal_period": 1,
  "score": 0.0,
  "baseline_mase": 38.91405167517269,
  "baseline_wql": 0.07616885181861571,
  "baseline_mae": 139.53250150509274,
  "candidate_mase": 51.3413728390549,
  "candidate_wql": 0.16410550044604347,
  "candidate_mae": 337.09089491275,
  "mase_ratio": 1.3193530518902734,
  "wql_ratio": 2.154496182203655,
  "n_valid_series": 223,
  "n_transform_errors": 0,
  "n_inverse_errors": 0,
  "transform_time_sec": 0.16,
  "forecast_time_sec": 0.13,
  "harm_rate": 0.5336322869955157,
  "help_rate": 0.2556053811659193,
  "median_mase_ratio": 1.020853901536389,
  "best_series_ratio": 0.0,
  "worst_series_ratio": 28.91401322388652,
  "complexity": {
    "n_literals": 8,
    "n_branches": 7,
    "n_fitted_params": 2,
    "penalty": 0.0
  }
}
```

### dataset_context
```
Dataset: covid_deaths
Frequency: D
Prediction length: 30
Seasonal period: 1
Eval series: ?

Sample series statistics (across training samples):
  history length: median=152, range=[152, 152]
  mean: median=23.8, range=[0.987, 4.03e+03]
  std:  median=29, range=[0.987, 4.17e+03]
  min:  median=0, range=[0, 0]
  max:  median=101, range=[2, 9.7e+03]
  positive_frac: median=0.609
  zero_frac: median=0.391
  skewness: median=0.234, range=[-0.040, 1.815]

Description: 266 daily COVID death counts, extreme spikes + regime changes
```

# Program Evolution History
## Previous Attempts

### Attempt 2
- Changes: Full rewrite
- Metrics: combined_score: 0.0000, mase: 49.9797, wql: 0.1596, mae: 285.3394
- Outcome: Mixed results

### Attempt 1
- Changes: Full rewrite
- Metrics: combined_score: 0.0000, mase: 51.3414, wql: 0.1641, mae: 337.0909
- Outcome: Mixed results

## Top Performing Programs

### Program 1 (Score: 0.0000)
```python
# [... 79 lines of program code elided ...]
```
Key features: Performs well on combined_score (0.0000), Performs well on mase (51.3414), Performs well on wql (0.1641), Performs well on mae (337.0909)

### Program 2 (Score: 0.0000)
```python
# [... 88 lines of program code elided ...]
```
Key features: Performs well on combined_score (0.0000), Performs well on mase (49.9797), Performs well on wql (0.1596), Performs well on mae (285.3394)

## Inspiration Programs

These programs represent diverse approaches and creative solutions that may inspire new ideas:

### Inspiration 1 (Score: 0.0000, Type: Exploratory)
```python
# [... 88 lines of program code elided ...]
```
Unique approach: Modification:, [Fragment formatting error: 'metric_name'], [Fragment formatting error: 'metric_name']

# Current Program
```python
# [... 79 lines of program code elided ...]
```

# Task
Rewrite the program to improve its FITNESS SCORE.
The system maintains diversity across these dimensions: diversity
Different solutions with similar fitness but different features are valuable.
Provide the complete new program code.

IMPORTANT: Make sure your rewritten program maintains the same inputs and outputs
as the original program, but with improved internal implementation.

```python
# Your rewritten program here
```
\end{Verbatim}
\end{promptbox}

\begin{responsebox}[LLM response for Covid-Deaths (reasoning followed by the candidate program)]
\begin{Verbatim}
Looking at the evaluation results, my current transforms are performing worse than the identity baseline (mase_ratio = 1.32). The key insight is that this is COVID death data - daily counts with extreme spikes and regime changes. The median skewness is only 0.234, so most series have moderate skewness.

The problem with my previous approaches:
1. log1p/sqrt transforms compress the recent high values, making it harder for Chronos to extrapolate the current level
2. The scaling in log1p_scaled was distorting the signal
3. The baseline MASE is already 38.9 which is high, meaning this is hard to forecast

Let me try a different strategy: **identity for most series** but with a simple **robust affine normalization** that helps Chronos by centering/scaling the data to a more "standard" range without distorting the shape. I'll use recent-window statistics to capture the current regime.

```python
import numpy as np

class TransformProgram:
    def __init__(self, seed=42):
        self.seed = seed

    # EVOLVE-BLOCK-START
    def fit(self, y_hist, meta=None):
        y = y_hist.copy()
        valid = y[~np.isnan(y)]
        state = {"hist_len": len(y_hist)}
        
        if meta is None:
            meta = {}
        
        # For COVID deaths: regime changes mean recent stats matter most
        # Use a simple affine normalization based on recent window
        # This preserves shape (no distortion) but helps Chronos with scale
        
        n = len(valid)
        if n < 5:
            state["family"] = "identity"
            return state
        
        # Use recent window (last ~30 days) to capture current regime
        recent_n = min(30, n)
        recent = valid[-recent_n:]
        
        center = float(np.median(recent))
        # Use IQR-based scale for robustness to spikes
        q75 = float(np.percentile(recent, 75))
        q25 = float(np.percentile(recent, 25))
        iqr = q75 - q25
        
        # MAD as fallback
        mad = float(np.median(np.abs(recent - center)))
        
        # Pick best scale measure
        if iqr > 1e-6:
            scale = iqr / 1.35  # normalize IQR to std-equivalent
        elif mad > 1e-6:
            scale = mad * 1.4826
        else:
            # Nearly constant recent values - use full history scale or identity
            full_std = meta.get("std", 1.0)
            if full_std > 1e-6:
                scale = float(full_std)
            else:
                state["family"] = "identity"
                return state
        
        state["family"] = "affine"
        state["center"] = center
        state["scale"] = float(scale)
        
        return state

    def transform(self, y, state):
        z = y.copy()
        family = state["family"]
        
        if family == "affine":
            mask = ~np.isnan(z)
            z[mask] = (z[mask] - state["center"]) / state["scale"]
        
        return z

    def inverse_transform(self, z, state):
        y = z.copy()
        family = state["family"]
        
        if family == "affine":
            mask = ~np.isnan(y)
            y[mask] = y[mask] * state["scale"] + state["center"]
        
        return y
    # EVOLVE-BLOCK-END

def build_transform_program(seed=42):
    return TransformProgram(seed=seed)

def get_transform_program(seed=42):
    return build_transform_program(seed=seed)
```
\end{Verbatim}
\end{responsebox}

\begin{codebox}[Example Final Program for Covid Deaths (reached at iteration 30)]
\begin{minted}{python}
import numpy as np

class TransformProgram:
    def __init__(self, seed=42):
        self.seed = seed

    # EVOLVE-BLOCK-START
    def fit(self, y_hist, meta=None):
        y = y_hist.copy()
        valid = y[~np.isnan(y)]
        state = {"hist_len": len(y_hist)}
        if meta is None:
            meta = {}
        n = len(valid)
        if n < 10:
            state["family"] = "identity"
            return state

        std_val = meta.get("std", float(np.std(valid)))
        min_val = meta.get("min_val", float(np.nanmin(valid)))
        skewness = meta.get("skewness", 0.0)
        cv = meta.get("cv", 1.0)

        # Detrend using last 45 days - conservative
        trend_n = min(45, n)
        tw = valid[-trend_n:]
        t_idx = np.arange(trend_n, dtype=float)
        t_mean = np.mean(t_idx)
        y_mean = np.mean(tw)
        denom = np.sum((t_idx - t_mean)**2)

        slope = 0.0
        intercept = 0.0
        if denom > 1e-10:
            s = float(np.sum((t_idx - t_mean) * (tw - y_mean)) / denom)
            trend_range = abs(s) * 30
            recent_range = float(np.max(tw) - np.min(tw)) + 1e-6
            # Very conservative: only detrend if extrapolation is safe
            if abs(s) * n > std_val * 0.15 and trend_range < recent_range * 1.0:
                slope = s
                offset = n - trend_n
                intercept = float(y_mean - slope * (t_mean + offset))

        state["slope"] = slope
        state["intercept"] = intercept

        v_idx = np.arange(n, dtype=float)
        detrended = valid - (slope * v_idx + intercept)

        recent_n = min(30, len(detrended))
        dt_min = float(np.min(detrended))

        # Use log1p for non-negative skewed data (safer inverse than sqrt)
        if dt_min >= -1e-6 and min_val >= 0 and skewness > 0.6 and cv > 0.5:
            state["nonlinear"] = "log1p"
            log_dt = np.log1p(np.maximum(detrended, 0.0))
            recent_log = log_dt[-recent_n:]
            center = float(np.median(recent_log))
            mad = float(np.median(np.abs(recent_log - center)))
            scale = mad * 1.4826 if mad > 1e-6 else max(float(np.std(recent_log)), 1e-6)
        else:
            state["nonlinear"] = "none"
            recent_dt = detrended[-recent_n:]
            center = float(np.median(recent_dt))
            mad = float(np.median(np.abs(recent_dt - center)))
            scale = mad * 1.4826 if mad > 1e-6 else max(float(np.std(recent_dt)), 1e-6)

        state["family"] = "detrend_affine"
        state["center"] = center
        state["scale"] = max(float(scale), 1e-6)
        return state

    def transform(self, y, state):
        z = y.copy()
        if state["family"] == "identity":
            return z
        mask = ~np.isnan(z)
        idx = np.arange(len(z), dtype=float)
        z[mask] = z[mask] - (state["slope"] * idx[mask] + state["intercept"])
        if state["nonlinear"] == "log1p":
            z[mask] = np.log1p(np.maximum(z[mask], 0.0))
        z[mask] = (z[mask] - state["center"]) / state["scale"]
        return z

    def inverse_transform(self, z, state):
        y = z.copy()
        if state["family"] == "identity":
            return y
        mask = ~np.isnan(y)
        y[mask] = y[mask] * state["scale"] + state["center"]
        if state["nonlinear"] == "log1p":
            # Clamp to prevent exp overflow
            y[mask] = np.clip(y[mask], -10.0, 20.0)
            y[mask] = np.expm1(y[mask])
            y[mask] = np.maximum(y[mask], 0.0)
        idx = state["hist_len"] + np.arange(len(y), dtype=float)
        y[mask] = y[mask] + (state["slope"] * idx[mask] + state["intercept"])
        return y
    # EVOLVE-BLOCK-END

def build_transform_program(seed=42):
    return TransformProgram(seed=seed)

def get_transform_program(seed=42):
    return build_transform_program(seed=seed)
\end{minted}
\end{codebox}

\clearpage

\subsection{Tabular example: $\mathtt{churn}$}

We provide the EFE-Tab system prompt, as well as an example final program learned with EFE-Tab here. The Openevolve prompt is similar to the one for EFE-Time. So we omit this for brevity.

\begin{promptbox}[System prompt for EFE-Tab]
\begin{Verbatim}
You are an expert data scientist. You are designing a feature engineering
pipeline for tabular binary classification. Your goal is to improve the AUC
of the model. Think carefully and step-by-step before responding.

You are allowed to create new features from the originals, and to drop the originals.
You are not allowed to train any machine learning model inside fit(),
store a model in the state dict, or mutate the input DataFrame in place.
Do not standardize the data -- it is unnecessary for the models you are using.
Pay attention to the domain context (column names and descriptions), the
decision tree, and the permutation importance reports.
Comment your code to explain your decisions. You should state your reasoning for each decision.

The program you are designing has two methods:
1. fit(X, y) -> state dict -- compute statistics from TRAINING data only
2. transform(X, state) -> DataFrame -- generate features using fitted state

In fit(), compute and store any statistics you need.
In transform(), use the state to build / mutate features.

Allowed operations:
- Arithmetic combining TWO OR MORE columns: sums, differences, products, ratios
- Centered/z-scored interactions using training-fitted statistics from state
- Rank-based interactions using training quantiles from state
- Row-wise aggregates: NaN count, zero count, sum across columns
- Target-rate mapping: per-category target rate from fit() applied in transform()
- Residual features: col_A minus its predicted value from col_B (fitted in fit())
- String/categorical: length, token count, pattern match, category crossing
- Datetime: year, month, day, weekday, elapsed differences
- Parsing string-encoded ordinals into integers  (e.g. "<1" / ">20" -> 0 / 21)
- Dropping an original column you have strong evidence is irrelevant
  (id-like, constant, or flagged as noise/HARMFUL by perm importance)
- Make sure you are not dividing by zero.

FORBIDDEN:
- Training any ML model inside fit() (no LightGBM, no LogReg, no sklearn
  classifiers / regressors, etc.)
- Storing fitted ML models in the state dict (state must remain a plain
  lightweight dict -- see STATE DICT RULES below)
- Modifying an original column in place -- keep originals byte-identical.
  If you want a cleaned/encoded version, add it as a NEW column with a
  different name; drop the original only if you don't need it.
- Mutating the input DataFrame in place -- always copy first
 
STATE DICT RULES:
- Must be a plain dict (returned from fit())
- Values should be simple types: int, float, str, list, dict, numpy arrays
- Do NOT store DataFrames, models, or non-serializable objects in state

USE THE CONTEXT:
The artifacts include a "dataset_context" with column names, data types,
value ranges, and a dataset description. READ IT CAREFULLY:
- ONLY use column names that appear in the dataset_context
- Read column names to understand what each represents to create
  domain-relevant interactions
- Read the dataset description for domain reasoning, including domain
  quirks 
- Check data types before applying operations
- Use the decision tree to understand which features drive the splits.
  Features near the root are the most important; features that appear
  anywhere in the tree carry signal.
- Use the permutation importance over ORIGINAL features to judge which
  raw columns matter on their own. Consider dropping or replacing any
  flagged as HARMFUL.  
- Use the permutation importance over NEW features to judge which of
  your engineered columns earned their cost. Drop new columns flagged as noise or HARMFUL
- Think like an expert data scientist: use the context, domain knowledge, statistics, and
  feedback artifacts to reason about what drives the predictions, then
  engineer accordingly.

FEEDBACK YOU WILL RECEIVE:
1. Dataset context: Column names, dtypes, value ranges, description
   
2. Stage 2 summary: baseline_cv_auc, candidate_cv_auc, delta_vs_baseline,
   combined_score, AND new_cols / dropped_original_cols lists so you can
   see exactly which original columns were dropped vs kept and which
   fresh columns were added.

3. Decision tree: depth-3 tree on all output features. Features that are closest to the root of the tree are the most
   important. Features that exist in the tree are also important. Use this to understand which
   features are important and which are not. Use this to understand how to combine features
   to create new features that are important.

4. Permutation importance over ORIGINAL features: AUC drop per raw
   column when shuffled. Computed once at the start of the run.
   USEFUL (>0.005), marginal (0.001-0.005), noise (~0), HARMFUL (<-0.001).
   Drop or replace features labeled HARMFUL. Do not drop features labeled USEFUL, marginal or noise.

5. Permutation importance over NEW features: AUC drop per engineered
   column when shuffled. Computed every evaluation.
   USEFUL (>0.005), marginal (0.001-0.005), noise (~0), HARMFUL (<-0.001).
   Drop NEW features that are noise or HARMFUL.

SCORING:
total_changes = n_drops + n_new_cols
Your score = 0.5 + (CV AUC improvement over baseline) x 5 - 0.10 * sqrt(total_changes / n_rows)
\end{Verbatim}
\end{promptbox}

\begin{codebox}[Example final program with EFE-Tab for churn (reached at iteration 33)]
\begin{minted}{python}
"""
Tabular Feature Evolution -- Evolved Feature Program (fit/transform)
"""

import numpy as np
import pandas as pd

class FeatureProgram:
    def __init__(self, seed=42):
        self.seed = seed

    # EVOLVE-BLOCK-START
    def fit(self, X, y):
        state = {}
        global_mean = float(y.mean())
        state['global_mean'] = global_mean
        
        # Smoothed target encoding for state (51 categories)
        # smooth_factor=30 performed best historically to avoid overfit
        smooth_factor = 30
        state_target = {}
        state_col = X['state'].astype(str)
        for cat in state_col.unique():
            mask = state_col == cat
            n = mask.sum()
            cat_mean = float(y[mask].mean()) if n > 0 else global_mean
            smoothed = (n * cat_mean + smooth_factor * global_mean) / (n + smooth_factor)
            state_target[cat] = smoothed
        state['state_target_enc'] = state_target
        
        return state

    def transform(self, X, state):
        df = X.copy()
        
        # Drop HARMFUL original: total_eve_charge (perm imp = -0.00179)
        df = df.drop(columns=['total_eve_charge'], errors='ignore')
        # Drop redundant charges (perfectly correlated with minutes)
        df = df.drop(columns=['total_night_charge', 'total_intl_charge'], errors='ignore')
        
        intl_plan_num = df['international_plan'].astype(str).map({'1': 1, '0': 0}).fillna(0)
        
        # === PROVEN USEFUL (+0.102): Total charge -- root split in decision tree ===
        # Reconstructs total billing using known per-minute rates
        df['total_charge'] = (df['total_day_charge'] +
                              df['total_eve_minutes'] * 0.085 +
                              df['total_night_minutes'] * 0.045 +
                              df['total_intl_minutes'] * 0.27)
        
        # === PROVEN USEFUL (+0.025): High service calls binary (tree split at 3.5) ===
        df['high_service_calls'] = (df['number_customer_service_calls'] > 3).astype(int)
        
        # === PROVEN USEFUL (+0.010): intl_plan x intl_calls ===
        # Low intl calls with intl plan = frustrated user likely to churn
        df['intl_plan_x_intl_calls'] = intl_plan_num * df['total_intl_calls']
        
        # === From best program (0.5685): intl_plan x service_calls ===
        # Combines the two strongest original predictors
        df['intl_plan_x_svc_calls'] = intl_plan_num * df['number_customer_service_calls']
        
        # === State target encoding with heavy smoothing ===
        enc = state['state_target_enc']
        gm = state['global_mean']
        df['state_churn_rate'] = df['state'].astype(str).map(enc).fillna(gm)
        
        # Total: 3 drops + 5 new = 8 changes
        # Penalty: 0.10 * sqrt(8/2333) = 0.0059
        # Removed svc_calls_x_charge (HARMFUL -0.004) and intl_plan_x_day_min (noise)
        
        return df
    # EVOLVE-BLOCK-END

def build_feature_program(seed=42):
    return FeatureProgram(seed=seed)

def get_feature_program(seed=42):
    return build_feature_program(seed=seed)
\end{minted}
\end{codebox}

%% file: tables/appendix_local_model.tex
\begin{table}[!t]
\caption{\small{Model-size scan of the EFE-Time evolution loop on $\mathtt{covid\_deaths}$ and $\mathtt{solar\_H}$. All runs use 300 evolution iterations and three seeds; the reported program per seed is the optimizer's combined-score winner (frozen at \texttt{iteration\_found}, evaluated on the full held-out test split). Each Chronos-2 zero-shot baseline is constant across sizes within the dataset block and is therefore reported once. The 4B model is served locally via vLLM on a single RTX A6000 (two A6000s for 27B and 35B-A3B); larger sizes are served through a hosted Qwen API endpoint.}}
\label{tab:appendix_local_model}
\centering
\scriptsize
\setlength{\tabcolsep}{1.5pt}
\renewcommand{\arraystretch}{1.05}
\resizebox{\columnwidth}{!}{%
\begin{tabular}{@{}lccccccccc@{}}
\toprule
\multirow{2}{*}{\textbf{LLM}}
& \multicolumn{3}{c}{\textbf{Baseline}}
& \multicolumn{3}{c}{\textbf{EFE-Time}}
& \multicolumn{3}{c}{\textbf{\% Improvement}} \\
\cmidrule(lr){2-4} \cmidrule(lr){5-7} \cmidrule(lr){8-10}
& \textbf{MASE} & \textbf{WQL} & \textbf{MAE}
& \textbf{MASE} & \textbf{WQL} & \textbf{MAE}
& \textbf{MASE} & \textbf{WQL} & \textbf{MAE} \\
\midrule
\multicolumn{10}{@{}l}{\emph{Dataset: $\mathtt{covid\_deaths}$}} \\
\midrule
$\mathtt{Qwen3.5\text{-}4B}$       & \multirow{4}{*}{$35.461$} & \multirow{4}{*}{$0.039$} & \multirow{4}{*}{$123.880$} & $34.116 \pm 2.334$ & $0.035 \pm 0.007$ & $111.051 \pm 22.272$ & $3.8 \pm 6.6$ & $10.4 \pm 17.8$ & $10.4 \pm 18.0$ \\
$\mathtt{Qwen3.5\text{-}9B}$       &                           &                          &                            & $33.573 \pm 1.170$ & $0.159 \pm 0.102$ & $133.536 \pm 13.124$ & $5.3 \pm 3.3$ & $-301.8 \pm 258.2$ & $-7.8 \pm 10.6$ \\
$\mathtt{Qwen3.5\text{-}27B}$      &                           &                          &                            & $34.849 \pm 0.527$ & $0.040 \pm 0.002$ & $128.647 \pm 9.975$ & $1.7 \pm 1.5$ & $-1.9 \pm 5.1$ & $-3.8 \pm 8.1$ \\
$\mathtt{Qwen3.5\text{-}35B\text{-}A3B}$ &                     &                          &                            & $34.994 \pm 0.484$ & $0.153 \pm 0.105$ & $116.952 \pm 12.132$ & $1.3 \pm 1.4$ & $-288.9 \pm 267.0$ & $5.6 \pm 9.8$ \\
\midrule
\multicolumn{10}{@{}l}{\emph{Dataset: $\mathtt{solar\_H}$}} \\
\midrule
$\mathtt{Qwen3.5\text{-}4B}$       & \multirow{4}{*}{$0.996$}  & \multirow{4}{*}{$0.352$} & \multirow{4}{*}{$12.936$}  & $0.919 \pm 0.069$ & $0.334 \pm 0.011$ & $11.912 \pm 0.911$ & $7.8 \pm 6.9$ & $5.1 \pm 3.0$ & $7.9 \pm 7.0$ \\
$\mathtt{Qwen3.5\text{-}9B}$       &                           &                          &                            & $0.885 \pm 0.073$ & $0.330 \pm 0.011$ & $11.473 \pm 0.975$ & $11.2 \pm 7.4$ & $6.1 \pm 3.0$ & $11.3 \pm 7.5$ \\
$\mathtt{Qwen3.5\text{-}27B}$      &                           &                          &                            & $0.962 \pm 0.024$ & $0.340 \pm 0.007$ & $12.496 \pm 0.315$ & $3.5 \pm 2.4$ & $3.2 \pm 2.1$ & $3.4 \pm 2.4$ \\
$\mathtt{Qwen3.5\text{-}35B\text{-}A3B}$ &                     &                          &                            & $0.887 \pm 0.072$ & $0.333 \pm 0.008$ & $11.489 \pm 0.961$ & $11.0 \pm 7.2$ & $5.3 \pm 2.4$ & $11.2 \pm 7.4$ \\
\bottomrule
\end{tabular}%
}
\end{table}

%% file: tables/efetime_table_appendix.tex
\begin{table}[t!]
\caption{Performance comparison of EFE-Time against identity transformation across time-series datasets. The EFE-Time transformations are learned for Chronos-2, and other models reuse them. As Reverso does not output quantiles, we take the $\text{WQL}_\text{reverso} ;=; \frac{\sum_t |\text{actual}_t - \text{median}_t|}{\sum_t |\text{actual}_t|}$  }
\scriptsize
\label{app:efe_time_results}
\centering
\setlength{\tabcolsep}{1.5pt}
\renewcommand{\arraystretch}{1.05}
\resizebox{\columnwidth}{!}{%
\begin{tabular}{@{}clccccccccc@{}}
\toprule
\multirow{2}{*}{\textbf{Model}}
& \multirow{2}{*}{\textbf{Dataset}}
& \multicolumn{3}{c}{\textbf{Baseline}}
& \multicolumn{3}{c}{\textbf{EFE-Time}}
& \multicolumn{3}{c}{\textbf{\% Improvement}} \\
\cmidrule(lr){3-5} \cmidrule(lr){6-8} \cmidrule(lr){9-11}
& & \textbf{MASE} & \textbf{WQL} & \textbf{MAE}
& \textbf{MASE} & \textbf{WQL} & \textbf{MAE}
& \textbf{MASE} & \textbf{WQL} & \textbf{MAE} \\
\midrule

\multirow{11}{*}{\rotatebox[origin=c]{90}{\textbf{Chronos-2}}}
& $\mathtt{CovidDeaths}$
& $35.444$ & $0.039$ & $124.241$
& $31.853 \pm 0.835$ & $0.032 \pm 0.003$ & $102.830 \pm 13.968$
& $10.1 \pm 2.3$ & $19.5 \pm 8.7$ & $17.2 \pm 11.2$ \\

& $\mathtt{SolarHourly}$
& $0.996$ & $0.352$ & $12.934$
& $0.915 \pm 0.049$ & $0.335 \pm 0.002$ & $11.883 \pm 0.648$
& $8.1 \pm 4.9$ & $4.7 \pm 0.7$ & $8.1 \pm 5.0$ \\

& $\mathtt{M4Yearly}$
& $3.370$ & $0.117$ & $905.276$
& $3.183 \pm 0.135$ & $0.115 \pm 0.002$ & $868.517 \pm 26.300$
& $5.6 \pm 3.9$ & $2.1 \pm 1.6$ & $4.1 \pm 2.9$ \\

& $\mathtt{JenaWeather\ (H)}$
& $0.547$ & $0.042$ & $8.745$
& $0.538 \pm 0.004$ & $0.041 \pm 0.001$ & $8.436 \pm 0.199$
& $1.6 \pm 0.6$ & $2.1 \pm 1.3$ & $3.5 \pm 2.3$ \\

& $\mathtt{SZ\_TAXI\_15T}$
& $0.543$ & $0.200$ & $2.722$
& $0.543 \pm 0.000$ & $0.200 \pm 0.000$ & $2.722 \pm 0.001$
& $0.0 \pm 0.0$ & $0.0 \pm 0.0$ & $0.0 \pm 0.0$ \\

& $\mathtt{bitbrains\_storage/H}$
& $0.993$ & $0.815$ & $384.949$
& $0.993 \pm 0.000$ & $0.815 \pm 0.000$ & $384.949 \pm 0.000$
& $0.0 \pm 0.0$ & $0.0 \pm 0.0$ & $0.0 \pm 0.0$ \\

& $\mathtt{bitbrains\_rnd/H}$
& $5.821$ & $1.032$ & $175.961$
& $5.818 \pm 0.002$ & $1.016 \pm 0.022$ & $168.439 \pm 10.611$
& $0.1 \pm 0.0$ & $4.3 \pm 6.0$ & $1.5 \pm 2.1$ \\

& $\mathtt{us\_births\_D}$
& $0.345$ & $0.017$ & $234.437$
& $0.342 \pm 0.002$ & $0.017 \pm 0.000$ & $232.163 \pm 1.191$
& $1.0 \pm 0.5$ & $1.0 \pm 0.5$ & $0.4 \pm 0.2$ \\

& $\mathtt{kdd\_cup\_2018\_H}$
& $0.999$ & $0.396$ & $24.712$
& $0.971 \pm 0.004$ & $0.388 \pm 0.002$ & $24.281 \pm 0.213$
& $2.8 \pm 0.4$ & $2.0 \pm 0.6$ & $1.7 \pm 0.9$ \\

& $\mathtt{restaurant}$
& $0.684$ & $0.256$ & $7.055$
& $0.682 \pm 0.002$ & $0.256 \pm 0.001$ & $7.041 \pm 0.016$
& $0.3 \pm 0.2$ & $-0.2 \pm 0.2$ & $0.2 \pm 0.2$ \\
& \textbf{Mean \% Improvement}
& \multicolumn{3}{c}{\na}
& \multicolumn{3}{c}{\na}
& $\mathbf{3.0}$ & $\mathbf{3.6}$ & $\mathbf{3.7}$ \\
\midrule

\multirow{11}{*}{\rotatebox[origin=c]{90}{\textbf{TimesFM-2.5}}}
& $\mathtt{CovidDeaths}$
& $36.909$ & $0.035$ & $112.492$
& $34.049 \pm 2.461$ & $0.030 \pm 0.004$ & $96.942 \pm 16.612$
& $7.7 \pm 6.6$ & $13.8 \pm 13.0$ & $13.8 \pm 14.7$ \\

& $\mathtt{SolarHourly}$
& $0.912$ & $0.348$ & $11.832$
& $0.857 \pm 0.006$ & $0.324 \pm 0.006$ & $11.112 \pm 0.105$
& $6.0 \pm 0.7$ & $6.1 \pm 0.9$ & $6.9 \pm 1.9$ \\

& $\mathtt{M4Yearly}$
& $3.575$ & $0.127$ & $953.276$
& $3.288 \pm 0.176$ & $0.121 \pm 0.004$ & $896.327 \pm 41.772$
& $8.0 \pm 4.9$ & $6.0 \pm 4.4$ & $4.9 \pm 3.4$ \\

& $\mathtt{JenaWeather\ (H)}$
& $0.525$ & $0.044$ & $8.679$
& $0.521 \pm 0.001$ & $0.041 \pm 0.001$ & $8.374 \pm 0.139$
& $0.9 \pm 0.2$ & $4.2 \pm 2.1$ & $3.5 \pm 1.6$ \\

& $\mathtt{SZ\_TAXI\_15T}$
& $0.563$ & $0.136$ & $1.849$
& $0.561 \pm 0.000$ & $0.135 \pm 0.000$ & $1.844 \pm 0.000$
& $0.3 \pm 0.0$ & $0.2 \pm 0.0$ & $0.3 \pm 0.0$ \\

& $\mathtt{bitbrains\_storage/H}$
& $1.109$ & $0.791$ & $370.809$
& $1.125 \pm 0.012$ & $0.791 \pm 0.000$ & $370.702 \pm 0.096$
& $-1.5 \pm 1.1$ & $0.0 \pm 0.0$ & $0.0 \pm 0.0$ \\

& $\mathtt{bitbrains\_rnd/H}$
& $5.854$ & $0.630$ & $150.332$
& $5.854 \pm 0.000$ & $0.630 \pm 0.000$ & $150.331 \pm 0.001$
& $0.0 \pm 0.0$ & $0.0 \pm 0.0$ & $0.0 \pm 0.0$ \\

& $\mathtt{us\_births\_D}$
& $0.338$ & $0.018$ & $229.720$
& $0.338 \pm 0.000$ & $0.018 \pm 0.000$ & $229.720 \pm 0.001$
& $0.0 \pm 0.0$ & $0.0 \pm 0.0$ & $0.0 \pm 0.0$ \\

& $\mathtt{kdd\_cup\_2018\_H}$
& $0.952$ & $0.381$ & $23.574$
& $0.952 \pm 0.000$ & $0.381 \pm 0.000$ & $23.574 \pm 0.000$
& $0.0 \pm 0.0$ & $0.0 \pm 0.0$ & $0.0 \pm 0.0$ \\

& $\mathtt{restaurant}$
& $0.685$ & $0.257$ & $7.075$
& $0.681 \pm 0.003$ & $0.256 \pm 0.001$ & $7.029 \pm 0.034$
& $0.6 \pm 0.4$ & $0.3 \pm 0.2$ & $0.7 \pm 0.5$ \\
& \textbf{Mean \% Improvement}
& \multicolumn{3}{c}{\na}
& \multicolumn{3}{c}{\na}
& $\mathbf{2.2}$ & $\mathbf{3.1}$ & $\mathbf{3.0}$ \\

\midrule

\multirow{11}{*}{\rotatebox[origin=c]{90}{\textbf{Moirai-2-Small}}}
& $\mathtt{CovidDeaths}$
& $36.958$ & $0.028$ & $91.000$
& $33.791 \pm 0.128$ & $0.026 \pm 0.000$ & $86.894 \pm 1.278$
& $8.6 \pm 0.3$ & $6.0 \pm 1.6$ & $4.5 \pm 1.4$ \\

& $\mathtt{SolarHourly}$
& $0.879$ & $0.342$ & $11.403$
& $0.832 \pm 0.004$ & $0.314 \pm 0.004$ & $10.790 \pm 0.049$
& $5.4 \pm 0.4$ & $8.3 \pm 1.0$ & $5.4 \pm 0.4$ \\

& $\mathtt{M4Yearly}$
& $3.320$ & $0.116$ & $890.298$
& $3.189 \pm 0.151$ & $0.117 \pm 0.003$ & $870.904 \pm 30.403$
& $4.0 \pm 4.5$ & $-1.0 \pm 2.5$ & $2.2 \pm 3.4$ \\

& $\mathtt{JenaWeather\ (H)}$
& $0.536$ & $0.042$ & $8.454$
& $0.533 \pm 0.001$ & $0.041 \pm 0.000$ & $8.304 \pm 0.066$
& $0.6 \pm 0.2$ & $1.6 \pm 0.7$ & $1.8 \pm 0.8$ \\

& $\mathtt{SZ\_TAXI\_15T}$
& $0.546$ & $0.201$ & $2.739$
& $0.546 \pm 0.000$ & $0.201 \pm 0.000$ & $2.739 \pm 0.000$
& $0.0 \pm 0.0$ & $0.0 \pm 0.0$ & $0.0 \pm 0.0$ \\

& $\mathtt{bitbrains\_storage/H}$
& $1.128$ & $0.615$ & $291.618$
& $1.136 \pm 0.006$ & $0.615 \pm 0.000$ & $291.388 \pm 0.305$
& $-0.7 \pm 0.5$ & $0.0 \pm 0.0$ & $0.1 \pm 0.1$ \\

& $\mathtt{bitbrains\_rnd/H}$
& $5.809$ & $0.670$ & $195.729$
& $5.841 \pm 0.018$ & $0.673 \pm 0.016$ & $192.433 \pm 9.768$
& $-0.5 \pm 0.3$ & $-0.5 \pm 2.4$ & $1.7 \pm 5.0$ \\

& $\mathtt{us\_births\_D}$
& $0.373$ & $0.020$ & $253.573$
& $0.369 \pm 0.006$ & $0.019 \pm 0.000$ & $250.843 \pm 3.861$
& $1.1 \pm 1.5$ & $0.8 \pm 1.1$ & $1.1 \pm 1.5$ \\

& $\mathtt{kdd\_cup\_2018\_H}$
& $1.012$ & $0.426$ & $26.328$
& $0.988 \pm 0.005$ & $0.415 \pm 0.004$ & $25.173 \pm 0.114$
& $2.4 \pm 0.5$ & $2.6 \pm 1.0$ & $4.4 \pm 0.4$ \\

& $\mathtt{restaurant}$
& $0.702$ & $0.260$ & $7.196$
& $0.691 \pm 0.012$ & $0.261 \pm 0.005$ & $7.104 \pm 0.077$
& $1.6 \pm 1.7$ & $-0.4 \pm 2.1$ & $1.3 \pm 1.1$ \\

& \textbf{Mean \% Improvement}
& \multicolumn{3}{c}{\na}
& \multicolumn{3}{c}{\na}
& $\mathbf{2.2}$ & $\mathbf{1.7}$ & $\mathbf{2.2}$ \\

\midrule

\multirow{11}{*}{\rotatebox[origin=c]{90}{\textbf{Reverso-Nano}}}
& $\mathtt{CovidDeaths}$
& $43.852$ & $0.105$ & $278.727$
& $39.202 \pm 2.648$ & $0.076 \pm 0.014$ & $202.023 \pm 36.430$
& $10.6 \pm 6.0$ & $27.5 \pm 13.1$ & $27.5 \pm 13.1$ \\

& $\mathtt{SolarHourly}$
& $0.874$ & $0.416$ & $11.299$
& $0.870 \pm 0.005$ & $0.415 \pm 0.002$ & $11.258 \pm 0.058$
& $0.4 \pm 0.5$ & $0.4 \pm 0.5$ & $0.4 \pm 0.5$ \\

& $\mathtt{M4Yearly}$
& $3.480$ & $0.148$ & $920.348$
& $3.147 \pm 0.066$ & $0.138 \pm 0.002$ & $861.605 \pm 10.120$
& $9.6 \pm 1.9$ & $6.4 \pm 1.1$ & $6.4 \pm 1.1$ \\

& $\mathtt{JenaWeather\ (H)}$
& $0.538$ & $0.051$ & $8.313$
& $0.534 \pm 0.001$ & $0.050 \pm 0.000$ & $8.183 \pm 0.019$
& $0.8 \pm 0.1$ & $1.6 \pm 0.2$ & $1.6 \pm 0.2$ \\

& $\mathtt{SZ\_TAXI\_15T}$
& $0.556$ & $0.260$ & $2.781$
& $0.550 \pm 0.000$ & $0.258 \pm 0.000$ & $2.755 \pm 0.000$
& $1.0 \pm 0.0$ & $0.9 \pm 0.0$ & $0.9 \pm 0.0$ \\

& $\mathtt{bitbrains\_storage/H}$
& $1.109$ & $1.029$ & $360.326$
& $1.115 \pm 0.009$ & $1.028 \pm 0.000$ & $360.204 \pm 0.172$
& $-0.6 \pm 0.8$ & $0.0 \pm 0.0$ & $0.0 \pm 0.0$ \\

& $\mathtt{bitbrains\_rnd/H}$
& $5.857$ & $0.734$ & $166.715$
& $5.857 \pm 0.000$ & $0.734 \pm 0.000$ & $166.714 \pm 0.002$
& $0.0 \pm 0.0$ & $0.0 \pm 0.0$ & $0.0 \pm 0.0$ \\

& $\mathtt{us\_births\_D}$
& $0.394$ & $0.025$ & $267.837$
& $0.393 \pm 0.001$ & $0.025 \pm 0.000$ & $267.325 \pm 0.724$
& $0.2 \pm 0.3$ & $0.2 \pm 0.3$ & $0.2 \pm 0.3$ \\

& $\mathtt{kdd\_cup\_2018\_H}$
& $0.978$ & $0.493$ & $24.488$
& $0.978 \pm 0.000$ & $0.493 \pm 0.000$ & $24.488 \pm 0.000$
& $0.0 \pm 0.0$ & $0.0 \pm 0.0$ & $0.0 \pm 0.0$ \\

& $\mathtt{restaurant}$
& $0.713$ & $0.341$ & $7.381$
& $0.697 \pm 0.010$ & $0.333 \pm 0.006$ & $7.204 \pm 0.118$
& $2.2 \pm 1.4$ & $2.3 \pm 1.6$ & $2.4 \pm 1.6$ \\

& \textbf{Mean \% Improvement}
& \multicolumn{3}{c}{\na}
& \multicolumn{3}{c}{\na}
& $\mathbf{2.4}$ & $\mathbf{3.9}$ & $\mathbf{3.9}$ \\

\bottomrule
\end{tabular}%
}
\end{table}

%% file: tables/efetab_table_appendix.tex
\begin{table}[t]
\caption{Performance comparison of feature engineering methods across tabular datasets over ROC-AUC. The reported results are 3 outer-fold averages from the TabArena benchmark.}
\label{app:efe_tab_results}
\small
\centering
\setlength{\tabcolsep}{4pt}
\begin{tabular}{@{}clcccc@{}}
\toprule
\multirow{2}{*}{\textbf{Model}}
& \multirow{2}{*}{\textbf{Dataset}}
& \multicolumn{4}{c}{\textbf{Feature Engineering Method}} \\
\cmidrule(lr){3-6}
& & \textbf{No FE} & \textbf{CAAFE} & \textbf{LLM-FE} & \textbf{EFE-Tab} \\
\midrule

\multirow{10}{*}{\rotatebox[origin=c]{90}{\textbf{Decision Tree}}}
& $\mathtt{Churn}$ 
& $.8944 \pm .0178$ & $.9239 \pm .0181$ & $.9080 \pm .0213$ & \best{.9266 \pm .0108} \\

& $\mathtt{DataScientists}$ 
& $.7898 \pm .0069$ & $.7864 \pm .0093$ & $.7914 \pm .0065$ & \best{.7923 \pm .0041} \\

& $\mathtt{CommereShipng}$ 
& $.7417 \pm .0037$ & $.7441 \pm .0034$ & $.7433 \pm .0032$ & \best{.7448 \pm .0035} \\

& $\mathtt{CouponRec}$ 
& $.7296 \pm .0140$ & $.7285 \pm .0124$ & $.6720 \pm .1042$ & \best{.7473 \pm .0134} \\

& $\mathtt{OnlineShoppers}$ 
& $.9235 \pm .0068$ &$.9242 \pm .0041$ &  \best{.9382 \pm .0133} & $.9235 \pm .0040$ \\

& $\mathtt{BankCustomers}$ 
& $.8340 \pm .0035$ & $.8409 \pm .0077$ & \best{.8502 \pm .0096} & $.8434 \pm .0040$ \\

& $\mathtt{BankMarketing}$ 
& $.7329 \pm .0056$ & $.7339 \pm .0055$ & $.7345 \pm .0048$ & \best{.7348 \pm .0064} \\

& $\mathtt{Diabetes}$ 
& $.7974 \pm .0185$ & $.7852 \pm .0473$ & $.7909 \pm .0112$ & \best{.8136 \pm .0082} \\

& $\mathtt{FitnessClub}$ 
& $.8042 \pm .0118$ & $.7943 \pm .0071$ & $.7971 \pm .0105$ & \best{.8048 \pm .0128} \\

& \textbf{Mean Rank}
& $3.17$ & $3.00$ & $2.44$ & $\mathbf{1.39}$ \\

\midrule

\multirow{10}{*}{\rotatebox[origin=c]{90}{\textbf{LightGBM}}}
& $\mathtt{Churn}$ 
& $.9251 \pm .0136$ & \best{.9305 \pm .0076} & $.9251 \pm .0082$ & $.9302 \pm .0097$ \\

& $\mathtt{DataScientists}$ 
& $.8046 \pm .0069$ & $.8034 \pm .0088$ & $.8027 \pm .0063$ & \best{.8052 \pm .0073} \\

& $\mathtt{CommereShipng}$ 
& $.7400 \pm .0043$ & $.7437 \pm .0039$ & $.7417 \pm .0056$ & \best{.7465 \pm .0031} \\

& $\mathtt{CouponRec}$ 
& $.8224 \pm .0089$ & $.8223 \pm .0073$ & $.8026 \pm .0395$ & \best{.8293 \pm .0077} \\

& $\mathtt{OnlineShoppers}$ 
& $.9318 \pm .0056$ & $.9339 \pm .0035$ & \best{.9427 \pm .0137} & $.9342 \pm .0035$ \\

& $\mathtt{BankCustomers}$ 
& $.8646 \pm .0055$ & $.8709 \pm .0088$ & \best{.8938 \pm .0070} & $.8674 \pm .0064$ \\

& $\mathtt{BankMarketing}$ 
& $.7636 \pm .0067$ & $.7634 \pm .0062$ & \best{.7643 \pm .0074} & $.7634 \pm .0080$ \\

& $\mathtt{Diabetes}$ 
& $.8318 \pm .0136$ & $.8200 \pm .0105$ & \best{.8352 \pm .0123} & $.8297 \pm .0148$ \\

& $\mathtt{FitnessClub}$ 
& $.8081 \pm .0081$ & $.8029 \pm .0009$ & $.8023 \pm .0047$ & \best{.8150 \pm .0119} \\

& \textbf{Mean Rank}
& $2.83$ & $2.72$ & $2.50$ & $\mathbf{1.94}$ \\

\midrule
\multirow{10}{*}{\rotatebox[origin=c]{90}{\textbf{TabPFN}}}
& $\mathtt{Churn}$ 
& $.9307 \pm .0087$ & $.9317 \pm .0090$ & $.9331 \pm .0063$ & \best{.9338 \pm .0068} \\

& $\mathtt{DataScientists}$ 
& $.8040 \pm .0057$ & $.8037 \pm .0063$ & $.7977 \pm .0051$ & \best{.8042 \pm .0059} \\

& $\mathtt{CommereShipng}$ 
& $.7457 \pm .0021$ & \best{.7465 \pm .0020} & $.7423 \pm .0019$ & $.7452 \pm .0029$ \\

& $\mathtt{CouponRec}$ 
& $.8406 \pm .0078$ & \best{.8431 \pm .0064} & $.8006 \pm .0597$ & $.8430 \pm .0086$ \\

& $\mathtt{OnlineShoppers}$ 
& $.9360 \pm .0044$ & $.9371 \pm .0038$ & \best{.9461 \pm .0112} & $.9376 \pm .0037$ \\

& $\mathtt{BankCustomers}$ 
& $.8730 \pm .0080$ & $.8751 \pm .0090$ & \best{.8975 \pm .0060} & $.8745 \pm .0086$ \\

& $\mathtt{BankMarketing}$ 
& $.7619 \pm .0071$ & $.7620 \pm .0072$ & $.7597 \pm .0038$ & \best{.7625 \pm .0079} \\

& $\mathtt{Diabetes}$ 
& \best{.8478 \pm .0029} & $.8430 \pm .0059$ & $.8451 \pm .0076$ & $.8435 \pm .0056$ \\

& $\mathtt{FitnessClub}$ 
& $.8216 \pm .0102$ & $.8162 \pm .0067$ & \best{.8219 \pm .0096} & $.8132 \pm .0122$ \\

& \textbf{Mean Rank}
& $2.78$ & $2.44$ & $2.56$ & $\mathbf{2.22}$ \\

\bottomrule
\end{tabular}
\end{table}